\documentclass[10pt,twocolumn,letterpaper]{article}
\usepackage[letterpaper,margin=0.72in,columnsep=0.28in]{geometry}

\usepackage{iftex}
\ifPDFTeX
  \usepackage[utf8]{inputenc}
  \usepackage[T1]{fontenc}
  \usepackage{tgpagella}
  \usepackage{textcomp}
  \usepackage{amssymb}          % \mathbb, which unicode-math supplies otherwise
\else
  \usepackage{fontspec}
  \usepackage{unicode-math}
  \IfFontExistsTF{texgyrepagella-math.otf}
    {\setmathfont{TeX Gyre Pagella Math}}
    {\setmathfont{Latin Modern Math}}
\fi
\usepackage{newunicodechar}

\usepackage{amsmath}
\usepackage{booktabs}
\usepackage{array}
\usepackage{graphicx}
\usepackage{adjustbox}
\usepackage[above,below]{placeins}
\usepackage{rotating}
\newlength{\rotmaxw}
\newlength{\rotmaxh}
\usepackage[font=small,labelfont=bf,skip=6pt]{caption}
\usepackage{xcolor}
\usepackage{colortbl}
\usepackage[verbose=silent]{microtype}
\usepackage{titlesec}
\usepackage{fancyhdr}
\usepackage{pifont}

\definecolor{accent}{RGB}{38,78,120}
\definecolor{linkblue}{RGB}{20,60,110}
\definecolor{urlblue}{RGB}{0,84,166}
\definecolor{headtint}{RGB}{24,42,64}
\definecolor{wmcell}{RGB}{207,224,240}
\definecolor{lgbcell}{RGB}{250,222,196}
\definecolor{avgcell}{RGB}{214,234,214}

\newif\ifjournal\journalfalse
\newif\ifsupplement\supplementfalse
\ifdefined\journalmode\journaltrue\fi
\ifdefined\supplementmode\journaltrue\supplementtrue\fi
\newcommand{\colfig}{\columnwidth}
\newcommand{\firstver}{\ifjournal development phase\else first version\fi}
\newcommand{\fvpaper}{\ifjournal development phase\else first version of this paper\fi}
\newcommand{\thisrev}{\ifjournal this study\else this revision\fi}
\newcommand{\revn}{\ifjournal study\else revision\fi}
\newcommand{\cpath}[2]{\ifjournal #2\else{\def\UrlBreaks{\do\/\do\_}\def\UrlBigBreaks{\do\/}\nolinkurl{#1}}\fi}  % code paths break only at / and _
\newcommand{\secbarrier}{}
\makeatletter
\expandafter\let\expandafter\origdblfigure\csname figure*\endcsname
\expandafter\let\expandafter\endorigdblfigure\csname endfigure*\endcsname
\makeatother
\newenvironment{topfigure}{\origdblfigure[t]}{\endorigdblfigure}
\makeatletter
\expandafter\let\expandafter\origdbltable\csname table*\endcsname
\expandafter\let\expandafter\endorigdbltable\csname endtable*\endcsname
\makeatother
\newenvironment{toptable}{\origdbltable[t]}{\endorigdbltable}
\newcommand{\seclabel}[1]{\label{#1}}
\ifsupplement
  \usepackage{float}
  \makeatletter
  \let\suporigfigure\figure\let\endsuporigfigure\endfigure
  \let\suporigtable\table\let\endsuporigtable\endtable
  \renewenvironment{figure}[1][]{\suporigfigure[H]}{\endsuporigfigure}
  \renewenvironment{table}[1][]{\suporigtable[H]}{\endsuporigtable}
  \makeatother
  \usepackage{stfloats}
  \renewenvironment{figure*}[1][]{\origdblfigure[!tb]}{\endorigdblfigure}
  \renewenvironment{table*}[1][]{\origdbltable[!tb]}{\endorigdbltable}
  \renewenvironment{topfigure}{\origdblfigure[!tb]}{\endorigdblfigure}
  
\fi
\usepackage{xr-hyper}
\usepackage[colorlinks=true,
            linkcolor=linkblue,
            citecolor=linkblue,
            urlcolor=urlblue,
            filecolor=urlblue,
            breaklinks=true]{hyperref}
\usepackage{xurl}  % URLs break at any character, so none runs into the margin

\ifPDFTeX
  \newcommand{\umap}[3]{\DeclareUnicodeCharacter{#2}{#3}}
\else
  \newcommand{\umap}[3]{\newunicodechar{#1}{#3}}
\fi
\umap{∈}{2208}{\ensuremath{\in}}
\umap{∉}{2209}{\ensuremath{\notin}}
\umap{∖}{2216}{\ensuremath{\setminus}}
\umap{∪}{222A}{\ensuremath{\cup}}
\umap{∩}{2229}{\ensuremath{\cap}}
\umap{⊆}{2286}{\ensuremath{\subseteq}}
\umap{⊂}{2282}{\ensuremath{\subset}}
\umap{ℝ}{211D}{\ensuremath{\mathbb{R}}}
\umap{∅}{2205}{\ensuremath{\emptyset}}
\umap{×}{00D7}{\ensuremath{\times}}
\umap{≤}{2264}{\ensuremath{\leq}}
\umap{≥}{2265}{\ensuremath{\geq}}
\umap{≈}{2248}{\ensuremath{\approx}}
\umap{≠}{2260}{\ensuremath{\neq}}
\umap{→}{2192}{\ensuremath{\rightarrow}}
\umap{↔}{2194}{\ensuremath{\leftrightarrow}}
\umap{−}{2212}{\ensuremath{-}}
\umap{±}{00B1}{\ensuremath{\pm}}
\umap{ρ}{03C1}{\ensuremath{\rho}}
\umap{τ}{03C4}{\ensuremath{\tau}}
\umap{Δ}{0394}{\ensuremath{\Delta}}
\umap{α}{03B1}{\ensuremath{\alpha}}
\umap{β}{03B2}{\ensuremath{\beta}}
\umap{²}{00B2}{\ensuremath{^{2}}}
\umap{³}{00B3}{\ensuremath{^{3}}}
\umap{₁}{2081}{\ensuremath{_{1}}}
\umap{₂}{2082}{\ensuremath{_{2}}}
\umap{·}{00B7}{\,\textperiodcentered\,}
\umap{✓}{2713}{\ding{51}}
\umap{✗}{2717}{\ding{55}}
\umap{ }{2009}{\,}   % U+2009 thin space
\umap{ }{00A0}{~}   % U+00A0 no-break space
\umap{‑}{2011}{-}   % U+2011 non-breaking hyphen

\renewcommand{\arraystretch}{1.3}

\newcommand{\secsquare}{%
  \textcolor{accent}{\rule[0.14em]{0.52em}{0.52em}}\hspace{0.42em}}
\titleformat{\section}
  {\normalfont\large\bfseries\color{headtint}}{\secsquare\thesection}{0.45em}{}
\titlespacing*{\section}{0pt}{16pt plus 4pt minus 2pt}{5pt plus 1pt}
\titleformat{\subsection}
  {\normalfont\normalsize\bfseries\color{headtint}}{\thesubsection}{0.45em}{}
\titlespacing*{\subsection}{0pt}{11pt plus 3pt minus 2pt}{3pt}
\titleformat{\paragraph}[runin]
  {\normalfont\bfseries\color{headtint}}{}{0pt}{}[.]
\titlespacing*{\paragraph}{0pt}{6pt plus 2pt}{0.6em}

\newcommand{\shorttitle}{Forecasting retail sales at every grain}
\renewcommand{\headrulewidth}{0.4pt}
\renewcommand{\headrule}{\hbox to\headwidth{\color{accent!45}\leaders\hrule height \headrulewidth\hfill}}
\newcommand{\computeHours}{44.7}
\newcommand{\computeRuns}{94}
\newcommand{\computeGpu}{Tesla T4}

\input{generated/rev_values}

\newcommand{\rqAvgBestCoarseWord}{eight}
\newcommand{\rqAvgBestItemWord}{nine}
\newcommand{\rqAvgBestBlockWord}{five}
\newcommand{\rqCells}{27}
\newcommand{\rqAvgBestCells}{22}

\newcommand{\rqWmWorstCells}{16}
\newcommand{\rqLgbWorstCells}{11}

\newcommand{\rqBlockMfiveWm}{0.598}
\newcommand{\rqBlockMfiveLgb}{0.953}
\newcommand{\rqBlockMfiveAvg}{0.589}

\newcommand{\rqPAvgWmBetter}{109}
\newcommand{\rqPAvgWmWorse}{22}
\newcommand{\rqPAvgWmN}{240}

\newcommand{\rqPAvgWmItemn}{100}
\newcommand{\rqPAvgLgbBetter}{106}
\newcommand{\rqPAvgLgbWorse}{18}

\newcommand{\rqPWmLgbBetter}{57}
\newcommand{\rqPWmLgbWorse}{63}

\newcommand{\rqPWmLgbCoarsebetter}{33}
\newcommand{\rqPWmLgbCoarseworse}{33}

\newcommand{\rqPWmLgbItembetter}{24}
\newcommand{\rqPWmLgbItemworse}{30}

\newcommand{\rqAvgItemWorseMax}{8}
\newcommand{\rqMcsWorldModel}{25}
\newcommand{\rqMcsLightgbm}{27}
\newcommand{\rqMcsAverage}{39}

\newcommand{\rqMcsCells}{48}
\newcommand{\rqFastShareDomLo}{11}
\newcommand{\rqFastShareDomHi}{21}
\newcommand{\rqFastShareRestLo}{37}
\newcommand{\rqFastShareRestHi}{42}
\newcommand{\rqShortShareDomLo}{11}
\newcommand{\rqShortShareDomHi}{21}
\newcommand{\rqShortShareRestMax}{2}
\newcommand{\rqAvgBestFast}{eight}
\newcommand{\rqAvgBestSlow}{eight}
\newcommand{\rqShortWmBeatsLgb}{seven}
\newcommand{\rqShortExceptions}{frozen entrees and Frat}
\newcommand{\rqShortExceptionsVerb}{being the exceptions}
\newcommand{\rqShortFavWm}{0.539}
\newcommand{\rqShortFavLgb}{0.667}
\newcommand{\rqShortMfiveWm}{0.866}
\newcommand{\rqShortMfiveLgb}{1.024}
\newcommand{\rqNineWord}{nine}
\newcommand{\rqSnaWm}{0.893}
\newcommand{\rqSnaLgb}{0.778}
\newcommand{\rqSnaAvg}{0.790}
\newcommand{\rqRuleLevel}{eight}
\newcommand{\rqRuleSna}{0.855}
\newcommand{\rqCostWmItemLo}{0.008}
\newcommand{\rqCostWmItemHi}{0.040}
\newcommand{\rqCostLgbItemLo}{0.058}
\newcommand{\rqCostLgbItemHi}{0.311}
\newcommand{\rqFaWmItemLo}{89}
\newcommand{\rqFaWmItemHi}{104}
\newcommand{\rqFaLgbItemLo}{66}
\newcommand{\rqFaLgbItemHi}{93}
\newcommand{\rqFaAvgItemLo}{78}
\newcommand{\rqFaAvgItemHi}{96}
\newcommand{\rqWmFaItemOff}{11}
\newcommand{\rqKeptSets}{five}
\newcommand{\rqItemWmBeatsAvg}{five}
\newcommand{\rqStoreWmBeatsLgb}{four}
\newcommand{\rqStoreAvgBeatsWm}{one}

\newcommand{\rqStoreLevel}{three}
\newcommand{\rqStoreWmLowest}{three}
\newcommand{\rqLgbFaItemLoFrac}{two thirds}
\newcommand{\rqCarriedWm}{0.447}
\newcommand{\rqOwnWm}{0.428}
\newcommand{\rqCarriedLgb}{0.450}
\newcommand{\rqOwnLgb}{0.448}
\newcommand{\rqCarriedWmIxs}{0.002}
\newcommand{\rqCarriedLgbIxs}{0.001}
\newcommand{\rqPoolGainMax}{0.003}
\newcommand{\rqPoolGainWm}{0.002}
\newcommand{\rqPoolGainWmSet}{analgesics}
\newcommand{\rqPoolGainLgb}{0.003}
\newcommand{\rqPoolGainLgbSet}{frozen entrees}
\newcommand{\rqPoolLossWm}{0.143}
\newcommand{\rqPoolLossWmSet}{soft drinks}
\newcommand{\rqPoolLossLgb}{0.215}
\newcommand{\rqPoolLossLgbSet}{analgesics}

\newcommand{\rqNoidSets}{five}
\newcommand{\rqNoidCostLo}{0.002}
\newcommand{\rqNoidCostHi}{0.061}
\newcommand{\rqNoidFaLo}{85}
\newcommand{\rqNoidFaHi}{105}
\newcommand{\rqWmBeatsNoid}{four}
\newcommand{\rqNoidBeatsWm}{none}

\newcommand{\rqNoidSeenWorse}{five}
\newcommand{\rqWmNoidLevelSets}{M5}

\newcommand{\loopTriesWm}{472}
\newcommand{\loopKeptWm}{10}
\newcommand{\loopImprovedWm}{six}
\newcommand{\loopBarLoWm}{0.004}
\newcommand{\loopBarHiWm}{0.034}
\newcommand{\loopGainWmSet}{Favorita}
\newcommand{\loopGainWmFrom}{0.547}
\newcommand{\loopGainWmTo}{0.468}
\newcommand{\loopHeldWorseWm}{two}
\newcommand{\loopHeldWorstWmSet}{snacks}
\newcommand{\loopHeldWorstWmFrom}{0.568}
\newcommand{\loopHeldWorstWmTo}{0.608}
\newcommand{\loopTriesLgb}{557}
\newcommand{\loopKeptLgb}{13}
\newcommand{\loopImprovedLgb}{nine}
\newcommand{\loopBarLoLgb}{0.002}
\newcommand{\loopBarHiLgb}{0.010}
\newcommand{\loopGainLgbSet}{Favorita}
\newcommand{\loopGainLgbFrom}{0.922}
\newcommand{\loopGainLgbTo}{0.580}
\newcommand{\loopHeldWorseLgb}{five}
\newcommand{\loopHeldWorstLgbSet}{soft drinks}
\newcommand{\loopHeldWorstLgbFrom}{0.435}
\newcommand{\loopHeldWorstLgbTo}{0.507}
\newcommand{\loopTries}{1,029}
\newcommand{\loopKept}{23}
\newcommand{\loopHeldCorrWm}{0.29}
\newcommand{\loopHeldCorrLgb}{0.05}
\newcommand{\loopSeedSdOne}{6.3}
\newcommand{\loopSeedSdFive}{1.8}
\newcommand{\loopOneCheBar}{0.040}

\newcommand{\computeSearchHours}{4,305}
\newcommand{\computeRevRuns}{8,368}
\newcommand{\computeRevHours}{5,153}
\newcommand{\computeVastUsd}{3,699}
\newcommand{\computeModalUsd}{3,035}
\newcommand{\computeTotalUsd}{6,734}
\newcommand{\computeVastGpuHours}{5,934}
\newcommand{\computeVastGpuUsd}{2,766}

\newcommand{\buildSeconds}{32}

\newcommand{\supTfmPlanBestUntrained}{eight}
\newcommand{\supFratTfmPlan}{0.482}
\newcommand{\supFratWm}{0.490}
\newcommand{\supFratLgb}{0.494}
\newcommand{\supFratAvg}{0.469}

\newcommand{\supAnaTfmPlan}{0.723}

\newcommand{\supAnaTstPlan}{13.892}
\newcommand{\supPNoWmBetter}{2}
\newcommand{\supPNoWmWorse}{22}
\newcommand{\supPNoWmN}{48}
\newcommand{\supPNoLgbBetter}{6}
\newcommand{\supPNoLgbWorse}{34}

\newcommand{\supPNoTfmBetter}{5}
\newcommand{\supPNoTfmWorse}{16}

\newcommand{\supPFitBottomBetter}{14}
\newcommand{\supPFitBottomWorse}{7}
\newcommand{\supPFitBottomN}{39}
\newcommand{\supPFitAllBetter}{8}
\newcommand{\supPFitAllWorse}{16}

\newcommand{\supNoWmWorse}{nine}

\newcommand{\supAblationSets}{nine}
\newcommand{\supPertPriceHelpsWm}{eight}
\newcommand{\supPertPlanHelpsWm}{nine}
\newcommand{\supPertPriceHelpsLgb}{eight}
\newcommand{\supPertPlanHelpsLgb}{nine}
\newcommand{\supPertSets}{nine}
\newcommand{\supPertUnitsWmLo}{7}
\newcommand{\supPertUnitsWmHi}{73}
\newcommand{\supPertUnitsLgbLo}{4}
\newcommand{\supPertUnitsLgbHi}{73}
\newcommand{\supPertHistHelps}{nine}

\newcommand{\supMechSets}{nine}
\newcommand{\supTimingLo}{2}
\newcommand{\supTimingHi}{64}
\newcommand{\supPersistTwoLo}{-4}
\newcommand{\supPersistTwoHi}{54}
\newcommand{\supPersistRiseSets}{six}
\newcommand{\supPersistSets}{eight}
\newcommand{\supProbeGainMax}{0.04}
\newcommand{\supProbeGainMin}{0.01}
\newcommand{\supCatPlanAvgLowest}{seven}
\newcommand{\supCatPlanLgbLowest}{two}

\newcommand{\supOrderAvgBelowLgb}{six}
\newcommand{\supOrderAvgBelowWm}{seven}
\newcommand{\supOrderSets}{nine}
\newcommand{\supCatPlanSets}{nine}
\newcommand{\supCovIxsWm}{79}
\newcommand{\supCovNinetyFiveIxsWm}{95}
\newcommand{\supSplIxsWm}{0.241}
\newcommand{\supCovIxsLgb}{79}
\newcommand{\supCovNinetyFiveIxsLgb}{95}
\newcommand{\supSplIxsLgb}{0.235}
\newcommand{\supCovIxsAvg}{79}
\newcommand{\supCovNinetyFiveIxsAvg}{95}
\newcommand{\supSplIxsAvg}{0.230}
\newcommand{\supCovTotalWm}{75}

\newcommand{\supCovTotalAvg}{80}

\newcommand{\supAvgBeatsEts}{seven}
\newcommand{\supReconSets}{nine}

\newcommand{\sfAblNoLgbWorse}{seven}
\newcommand{\sfAblNoTfmWorse}{six}
\newcommand{\sfAblNoWmWorse}{eight}
\newcommand{\sfBrierAvg}{0.190}

\newcommand{\sfBrierBeatsClimNames}{Frat}

\newcommand{\sfBrierClim}{0.138}
\newcommand{\sfBrierLgb}{0.190}
\newcommand{\sfBrierWm}{0.190}
\newcommand{\sfCalWindows}{3}
\newcommand{\sfCalWindowsMax}{5}
\newcommand{\sfCalendarMax}{0.019}
\newcommand{\sfCovEightyMaxIxsAvg}{94}

\newcommand{\sfCovEightyMaxTotalWm}{92}
\newcommand{\sfCovEightyMinIxsAvg}{71}

\newcommand{\sfCovEightyMinTotalWm}{48}
\newcommand{\sfCovFiftyIxsAvg}{55}
\newcommand{\sfCovFiftyIxsLgb}{55}
\newcommand{\sfCovFiftyIxsWm}{54}

\newcommand{\sfCovLevelItemAvg}{71}
\newcommand{\sfCovLevelItemLgb}{70}
\newcommand{\sfCovLevelItemWm}{71}

\newcommand{\sfCovLevelStoreAvg}{66}
\newcommand{\sfCovLevelStoreLgb}{68}
\newcommand{\sfCovLevelStoreWm}{65}

\newcommand{\sfCovTotalWindowMax}{100}
\newcommand{\sfCovTotalWindowMin}{0}
\newcommand{\sfCrpsAvgBestIxs}{eight}
\newcommand{\sfCrpsAvgBestTot}{five}
\newcommand{\sfCrpsLgbBestIxs}{none}
\newcommand{\sfCrpsLgbBestTot}{three}
\newcommand{\sfCrpsTieIxs}{one}
\newcommand{\sfCrpsTieTot}{none}
\newcommand{\sfCrpsWmBestIxs}{none}
\newcommand{\sfCrpsWmBestTot}{one}
\newcommand{\sfDevBetter}{10}
\newcommand{\sfDevChanges}{16}
\newcommand{\sfDevLatentMax}{0.062}
\newcommand{\sfDevNoRef}{3}
\newcommand{\sfDevNotRun}{13}
\newcommand{\sfDevRuns}{60}
\newcommand{\sfDevScored}{44}
\newcommand{\sfFavHistFour}{8.2}
\newcommand{\sfHistSets}{nine}
\newcommand{\sfHistShortWorse}{nine}
\newcommand{\sfJepaKeptNames}{Snacks}

\newcommand{\sfJepaScreenBetter}{2}
\newcommand{\sfJepaScreens}{18}

\newcommand{\sfLossKeptNames}{Snacks, M5, Favorita}
\newcommand{\sfLossKeptSets}{three}
\newcommand{\sfLossScreenBetter}{11}
\newcommand{\sfLossScreens}{36}
\newcommand{\sfLossSets}{nine}
\newcommand{\sfLossTried}{four}
\newcommand{\sfPermAgree}{43}
\newcommand{\sfPermPairs}{46}
\newcommand{\sfProbSets}{nine}
\newcommand{\sfPromoLgbHelps}{seven}
\newcommand{\sfPromoSets}{eight}
\newcommand{\sfPromoWmHelps}{eight}
\newcommand{\sfRelHalfAvg}{60}

\newcommand{\sfRelNinetyAvg}{82}

\newcommand{\sfRelNinetyTotLgb}{88}
\newcommand{\sfRelNinetyTotWm}{86}

\newcommand{\sfRelTenAvg}{27}

\newcommand{\sfRelTenTotLgb}{7}
\newcommand{\sfRelTenTotWm}{11}

\newcommand{\sfWtLgbBottomMean}{0.38}
\newcommand{\sfWtTfmBottomMean}{0.31}
\newcommand{\sfWtTfmTotalMean}{0.24}
\newcommand{\sfWtWmBottomMean}{0.31}

\newcommand{\lhSets}{seven}

\newcommand{\lhLowestAvgDirect}{five}
\newcommand{\lhDirectBeatsBlocksWm}{seven}
\newcommand{\lhDirectSetsWm}{seven}
\newcommand{\lhDirectBeatsBlocksLgb}{four}
\newcommand{\lhDirectSetsLgb}{seven}
\newcommand{\lhPAvgDirBetter}{30}
\newcommand{\lhPAvgDirWorse}{6}
\newcommand{\lhPAvgDirN}{72}
\newcommand{\lhPAvgDirWmBetter}{59}
\newcommand{\lhPAvgDirWmWorse}{0}

\newcommand{\lhLgbBestBeatsWm}{seven}
\newcommand{\lhPWmBetter}{5}
\newcommand{\lhPWmWorse}{38}
\newcommand{\lhPN}{72}
\newcommand{\lhPAvgWmBetter}{58}
\newcommand{\lhPAvgWmWorse}{2}
\newcommand{\lhPAvgLgbBetter}{19}
\newcommand{\lhPAvgLgbWorse}{9}
\newcommand{\lhGrowWmLo}{−0.065}
\newcommand{\lhGrowWmHi}{+0.237}

\newcommand{\lhGrowLgbLo}{−0.080}
\newcommand{\lhGrowLgbHi}{+0.150}

\newcommand{\lhAllLowestAvgDirect}{three}

\newcommand{\lhTimestralFaOneLo}{1.2}
\newcommand{\lhTimestralFaOneHi}{6.2}
\newcommand{\lhTimestralFaLo}{1.4}
\newcommand{\lhTimestralFaHi}{18.7}

\newcommand{\lhPTunedLgbBetter}{5}
\newcommand{\lhPTunedLgbWorse}{11}

\newcommand{\lhPTunedAvgLgbBetter}{16}
\newcommand{\lhPTunedAvgLgbWorse}{2}

\newcommand{\lhPTunedWmLgbBetter}{3}
\newcommand{\lhPTunedWmLgbWorse}{17}
\newcommand{\lhPTunedWmLgbN}{48}
\newcommand{\lhMfiveAvgRollonLgbBetter}{14}

\newcommand{\lhMfiveAvgRollonLgbN}{24}

\newcommand{\lhMfiveAvgRollon}{0.764}
\newcommand{\lhMfiveLgbDirect}{0.839}
\newcommand{\lhMfiveWmRollon}{0.880}
\newcommand{\lhMfiveWmBlocks}{1.016}
\newcommand{\lhRollonCheese}{5.03}
\newcommand{\lhRollonCheeseFa}{1.74}
\newcommand{\lhWmDirAheadN}{two}
\newcommand{\lhWmDirSets}{seven}
\newcommand{\lhWmDirAheadNames}{bottled juice and snacks}
\newcommand{\lhMfiveWmDirEarly}{0.994}
\newcommand{\lhMfiveLgbDirEarly}{0.770}
\newcommand{\lhMfiveWmDirLate}{0.850}
\newcommand{\lhMfiveLgbDirLate}{0.908}
\newcommand{\lhPWmDirBetter}{9}
\newcommand{\lhPWmDirWorse}{24}
\newcommand{\lhPWmDirN}{72}
\newcommand{\lhGapCutLo}{+0.021}
\newcommand{\lhGapCutHi}{+0.135}
\newcommand{\lhGapRegLo}{−0.024}
\newcommand{\lhGapRegHi}{+0.011}

\newcommand{\kAheadFav}{11}

\newcommand{\kAheadMfive}{9}

\newcommand{\kCutWorseDom}{six}

\newcommand{\kDFavOtherPeriods}{$-$0.025}
\newcommand{\kDFavPriceCutPeriods}{$-$0.069}

\newcommand{\kDFratOtherPeriods}{+0.005}
\newcommand{\kDFratPriceCutPeriods}{+0.005}

\newcommand{\kDMfiveOtherPeriods}{$-$0.018}
\newcommand{\kDMfivePriceCutPeriods}{$-$0.010}

\newcommand{\kHeavyWorseDom}{five}
\newcommand{\kKinds}{11}

\newcommand{\kShortBetterAll}{seven}

\newcommand{\kWmBetterNoPromotion}{five}

\newcommand{\kWmBetterShortHistory}{six}
\newcommand{\kWmBetterSlow}{five}

\newcommand{\kWmWorseFast}{six}
\newcommand{\kWmWorseHeavyPromotion}{six}

\newcommand{\kWmWorseLongHistory}{six}
\newcommand{\kWmWorseNoPromotion}{three}

\newcommand{\kWmWorseShortHistory}{one}
\newcommand{\kWmWorseSlow}{four}

\newcommand{\ptThreeBelowAvg}{six}
\newcommand{\ptThreeAboveAvg}{one}
\newcommand{\ptNoWmAtOrBelowAvg}{four}
\newcommand{\ptNoWmAboveAvg}{five}
\newcommand{\ptSets}{nine}

\newcommand{\histAheadLong}{four}
\newcommand{\histAheadShort}{three}
\newcommand{\histAvgBest}{26}
\newcommand{\histCells}{30}
\newcommand{\histDomSets}{six}
\newcommand{\histDomShrinks}{three}

\newcommand{\histFavGapLong}{$-$0.024}
\newcommand{\histFavGapShort}{+0.020}

\newcommand{\histFavYearsLong}{4}
\newcommand{\histFavYearsShort}{1}
\newcommand{\histMfiveGapLong}{+0.019}
\newcommand{\histMfiveGapShort}{$-$0.029}

\newcommand{\histMfiveYearsLong}{4}
\newcommand{\histMfiveYearsShort}{1}
\newcommand{\histSets}{eight}

\newcommand{\hoAvgBeatsLgb}{six}
\newcommand{\hoAvgBeatsWm}{eight}
\newcommand{\hoAvgBest}{three}

\newcommand{\hoLgbBest}{three}
\newcommand{\hoMfiveFinalAvg}{0.565}
\newcommand{\hoMfiveFinalLGB}{0.551}
\newcommand{\hoMfiveFinalWM}{0.643}
\newcommand{\hoOrderSame}{four}
\newcommand{\hoSets}{eight}
\newcommand{\hoWmBest}{two}

\newcommand{\domCatDealRest}{28}
\newcommand{\domCatDealSix}{36}
\newcommand{\domCatDollarShare}{42}
\newcommand{\domCatEarlyGapHardHi}{+0.876}
\newcommand{\domCatEarlyGapHardLo}{+0.117}
\newcommand{\domCatEarlyLgbWins}{14}
\newcommand{\domCatEarlyWmWins}{9}
\newcommand{\domCatN}{28}
\newcommand{\domCatRankBjc}{fourth}
\newcommand{\domCatRankChe}{third}
\newcommand{\domCatRankSdr}{seventh}
\newcommand{\domCatRankSna}{first}
\newcommand{\domCatRest}{22}
\newcommand{\domCatTunedBehind}{three}
\newcommand{\domCatTunedGapHi}{+0.115}
\newcommand{\domCatTunedGapLo}{$-$0.030}

\ifsupplement\else\ifjournal\fi\fi
\begin{document}

% ------------------------------------------------------------ title block -
\ifsupplement\twocolumn[
  \vspace*{6pt}\begin{center}{\LARGE\bfseries\color{headtint} Online supplement\par}\vspace{4pt}
  {\large Forecasting Retail Sales at Every Grain: Trees, Foundation Models,\\ Transformers and World Models Compared on Four Public Data Sets\par}\vspace{6pt}
  \vspace{14pt}\end{center}]
\else
\twocolumn[
  \vspace*{6pt}
  \begin{center}
    {\LARGE\bfseries\color{headtint} Forecasting Retail Sales at Every Grain:\\[2pt]
     Trees, Foundation Models, Transformers and World Models\\[2pt]
     Compared on Four Public Data Sets\par}
    \vspace{10pt}
    \ifjournal\else
    {\large Karthik Iyer$^{1}$ \quad Madan Dabbeeru$^{2}$\par}
    \vspace{2pt}
    {\footnotesize $^{1}$La Trobe University, Bundoora, Australia \quad $^{2}$Eizen, Suwanee, Georgia, United States\par}
    \vspace{3pt}
    {\footnotesize Correspondence:
     \href{mailto:karthik.iyer.au@gmail.com}{karthik.iyer.au@gmail.com}
     \quad ORCID:
     \href{https://orcid.org/0009-0004-0593-1602}{0009-0004-0593-1602}\par}
    \fi
    \vspace{18pt}
    \begin{minipage}{0.86\textwidth}
      \ifjournal\small
\textbf{Abstract.}
Retailers plan orders, promotions, category reviews and supplier terms from one demand at
different grains. We ask whether one item-by-store forecast, summed to each coarser grain, can serve
them all, and which model does it best, comparing gradient-boosted trees (LightGBM), foundation
models, the Temporal Fusion Transformer and a world model that carries each item's state under the
price and promotion plan on four public data sets (M5, Dominick's, Breakfast at the Frat, Favorita),
the world model and LightGBM tuned by one search procedure. The average of the two, summed to every
grain, is the safe default: it has the lowest error in \rqAvgBestCells{} of \rqCells{} grain and data
set cases, wins paired item tests against either model far more often than it loses, and on final
periods held out of every tuned run is never the least accurate, though the most accurate on only
\hoAvgBest{} of \hoSets{}. The world model alone is the choice for existing items not yet trained on, where it
loses at most \rqCostWmItemHi{} in scaled error against up to \rqCostLgbItemHi{} for LightGBM, and it
leads on lines with short history. LightGBM is stronger on heavily promoted lines, fast sellers,
long histories and price-cut weeks on Dominick's. The evidence is inconclusive for slow and
never-promoted lines, for longer training history, for pooling data sets, and a year ahead, where
the average is the most accurate on \lhLowestAvgDirect{} of \lhSets{} data sets and LightGBM on the rest.

\vspace{4pt}
\textbf{Keywords:} retail forecasting; forecast grain; hierarchical evaluation;
forecast combination; promotions; world models; foundation models.
\else\small
\textbf{Abstract.}
A retailer makes decisions from the same demand at different grains and on
different cycles: store orders by item and day, promotions by item and week in
blocks of weeks, category reviews week by week, and supplier ranges and terms
over a year. It wants one forecast that serves all of them without losing value
between plans. This paper asks whether one forecast, built at the finest grain
and summed to every coarser grain, can serve these decisions across every kind
of line, and which kind of model does that best. It compares gradient-boosted
trees (LightGBM), pretrained foundation models (TimesFM, Chronos-Bolt,
Timestral), an attention-based model (the Temporal Fusion Transformer) and a
world model that carries each item's state forward under the price and
promotion plan, on four public data sets (M5, six Dominick's categories,
Breakfast at the Frat and Corporaci\'on Favorita), with the world model and
LightGBM tuned by a common search procedure. Results are read by grain, by the nature of
the line, on seen and unseen data, across retailers and a year ahead, and fall into four parts.
\emph{When to use the average.} On lines and stores the models were trained on, the tuned world
model and LightGBM are close, and their average, summed to every grain, is the safe default: it
has the lowest error in \rqAvgBestCells{} of \rqCells{} grain and data set cases; in paired item tests
it is clearly more accurate than the world model in \rqPAvgWmBetter{} of \rqPAvgWmN{} comparisons
(clearly less in \rqPAvgWmWorse{}) and than LightGBM in \rqPAvgLgbBetter{} (clearly less in
\rqPAvgLgbWorse{}); and on the final periods every final run held out it is never the least
accurate, though the most accurate on only \hoAvgBest{} of \hoSets{} data sets.
\emph{When to use the world model.} On existing items kept out of training (new launches with no sales were not
tested) the world model loses at most
\rqCostWmItemHi{} in scaled error and forecasts within \rqWmFaItemOff\% of actual sales, while
LightGBM loses up to \rqCostLgbItemHi{} and forecasts as little as \rqLgbFaItemLoFrac{} of actual
sales; fitted without the item label, LightGBM loses at most \rqNoidCostHi{} but is still less
accurate on \rqWmBeatsNoid{} of \rqNoidSets{} data sets. On lines with short history the world model
is clearly more accurate than LightGBM on \kWmBetterShortHistory{} of \rqNineWord{} data sets
(less on \kWmWorseShortHistory{}).
\emph{When to use LightGBM.} LightGBM is clearly more accurate than the world model on heavily
promoted lines, fast sellers and lines with long history (on \kWmWorseHeavyPromotion{},
\kWmWorseFast{} and \kWmWorseLongHistory{} of \rqNineWord{} data sets), and in price-cut weeks on
every Dominick's category.
\emph{Where the evidence is inconclusive.} Lines never promoted (\kWmBetterNoPromotion{} data
sets to \kWmWorseNoPromotion{}) and slow lines (\kWmBetterSlow{} to \kWmWorseSlow{}) are close to
even; more training history helped the world model on Favorita and hurt it on M5; settings
carried from M5 to Favorita and training on other data sets improved neither model by more than
\rqPoolGainMax{}; and in a test 52 weeks ahead on Dominick's and M5, added in \thisrev{}, the
average forecasting the whole year at once is the most accurate on \lhLowestAvgDirect{} of
\lhSets{} data sets and LightGBM on the rest, with the world model ahead of LightGBM head to head on
\lhWmDirAheadN{} of \lhWmDirSets{}. The findings hold for the data sets studied and the price and
promotion levers; anyone adopting them should rerun every comparison, including those that lost
here, on their own data.
% Numbers from results/analysis_tuned (nine data sets) and the transfer runs.

\vspace{4pt}
\textbf{Keywords:} retail forecasting; forecast grain; hierarchical evaluation;
forecast combination; promotions; world models; foundation models.
\fi
    \end{minipage}
    \vspace{16pt}
  \end{center}
]
\fi
\thispagestyle{plain}

\ifsupplement% Body of the online supplement: sections numbered S1, S2, ...; tables and figures S1, S2, ...
\renewcommand{\thesection}{S\arabic{section}}
\renewcommand{\thetable}{S\arabic{table}}
\renewcommand{\thefigure}{S\arabic{figure}}

This supplement holds the detail behind the main text: every run of the settings
searches; the supporting analyses rerun on the tuned models, among them the
ensemble ablation, the plan-input occlusion, the carried-state tests and the
calibration of the forecast ranges; what was tried and dropped, with every loss
and pretraining recipe explored; and every constraint the models work within. Every number is generated
from the same result files as the main text.

\section{Losses and setup in detail}
\label{sec:sup-method}

\subsection{The changes to the loss tried}
% The changes to the loss tried: in the main text for arXiv, in the online supplement for the journal.
\paragraph{The changes tried}
Each change below adds to or replaces $L_p$ for one batch $B$ of item, store and
period examples. Table~\ref{tab:losses} scores each against the configuration it
changed in the \firstver{}, where none was kept; the search of
Section~\ref{sec:loop} tried them again on the tuned settings, with the
likelihoods above, and kept the changes listed in Table~\ref{tab:loop}.
\begin{itemize}\setlength{\itemsep}{1pt}
\item \emph{Negative binomial} on units, $-\log\mathrm{NB}(y;\,\mu\ell_i, r)$
with the spread $r$ learnt.
\item \emph{Totals term}: the loss adds
$\lambda_T\,\mathrm{mean}_t\big[(\sum_{b\in B_t}\hat y_b - \sum_{b\in B_t} y_b) /
\max(1, \sum_{b\in B_t} y_b)\big]^2$, the relative gap between summed forecasts
and sales over the examples $B_t$ sharing a date, so that a small bias in the
same direction on many items is penalised ($\lambda_T = 1$).
\item \emph{Volume and dollar weights}: each example's loss times
$\ell_i^{\,\gamma}$, or $(\ell_i\bar\pi_i)^{\gamma}$ with $\bar\pi_i$ the item's
usual price, normalised to mean one in the batch, with $\gamma = 0.9$ and
$p = 1.1$, which makes the loss close to LightGBM's on units or dollars.
\item \emph{Peak weight}: each period's loss times
$1 + \kappa\min(|z_t - 1|, 5)$, normalised to mean one, so that peaks and dips
count for more ($\kappa = 0.5$, with the dollar weight above).
\item \emph{After-promotion weight}: each period with no lever on that follows a
period with one counts $1 + w$ times, normalised to mean one ($w = 2$ or $5$).
\item \emph{Latent term}: from the state now and the plan, the transition
predicts the state the encoder gives once those periods are history, against a
slowly updated copy of the encoder; the loss adds
$\lambda_Z \sum_h \delta^h \lVert s_{t+h}(p) - \bar s_{t+h}\rVert^2$ with
$\delta \le 1$ discounting far periods, and optionally a variance and covariance
penalty on the state~\cite{vicreg}. It was also used alone to pretrain the
encoder, which was then frozen, trained at a lower rate or trained fully on the
forecast loss.
\end{itemize}

\begin{table*}[t]
\centering
\caption{\textbf{No change to the training loss became a default: each was lower in error than the configuration
it changed in at most half its comparisons.} Each row is one change to the Tweedie loss of the world model or lift
model, scored on the development windows against the configuration with the same data, windows and every other
setting. WRMSSE: overall (mean of the hierarchy levels), averaged over the seeds and windows both ran and over the
data sets in the row. Lower: the number of data sets (or category and model pairs) in which the change had the lower
error. The weight after a promotion comes from a separate rerun of the world model in the same job. Rows marked
not run were configured but have no full-data run in the records. Built by \texttt{scripts/loss\_ledger.py}.}
\label{tab:losses}
\footnotesize
\begin{tabular}{p{0.36\textwidth}p{0.17\textwidth}rrl}
\toprule
\textbf{Change to the loss} & \textbf{Tested on} & \textbf{With it} & \textbf{Without} & \textbf{Result} \\
\midrule
Negative binomial likelihood & Dominick's, Frat & 0.537 & 0.541 & Lower in 1 of 2 \\
Totals term & Dominick's, Frat & 0.655 & 0.583 & Lower in 0 of 3 \\
Unit-volume weight and Tweedie power 1.1 & Dominick's & n/a & n/a & Not run on full data \\
Dollar-volume weight and Tweedie power 1.1 & Dominick's & n/a & n/a & Not run on full data \\
Peak weight, dollar weight and Tweedie power 1.1 & Dominick's & 0.853 & 0.856 & Lower in 5 of 12 \\
Weight 3 on the periods after a promotion & Dominick's & 0.834 & 0.832 & Lower in 2 of 6 \\
Weight 6 on the periods after a promotion & Dominick's & 0.810 & 0.832 & Lower in 2 of 6 \\
Latent term, weight 0.1 & Frat & 0.411 & 0.406 & Lower in 0 of 1 \\
Latent term, weight 0.5 & Frat & 0.414 & 0.407 & Lower in 0 of 2 \\
Latent term, weight 2 & Frat & 0.410 & 0.406 & Lower in 0 of 1 \\
Latent term with a spread penalty and discount & Dominick's, Frat & 0.567 & 0.549 & Lower in 0 of 4 \\
Latent pretraining, encoder frozen afterwards & Frat & 0.428 & 0.406 & Lower in 0 of 1 \\
Latent pretraining, encoder full afterwards & Frat & 0.407 & 0.406 & Lower in 0 of 1 \\
Latent pretraining, encoder at a lower rate afterwards & Frat & 0.411 & 0.407 & Lower in 0 of 3 \\
Latent pretraining, then forecast with the latent term & Frat & 0.415 & 0.406 & Lower in 0 of 1 \\
Latent pretraining, then forecast with a totals term & Dominick's, Frat & 0.729 & 0.667 & Lower in 0 of 1 \\
\bottomrule
\end{tabular}
\end{table*}

\subsection{Setup of the supporting analyses}
% Setup of the supporting analyses: in the main text for arXiv, in the online supplement for the journal.
\paragraph{Intervals}
Each model's interval forecasts come from its own errors in the earlier rolling
windows of the same data set, a rolling form of conformal
calibration~\cite{vovk2005}. With $e_{j,h} = (y_{j,t+h} - \hat y_{j,t+h})/\sigma_j$
the scaled errors of the earlier windows at horizon $h$ ($\sigma_j$ the series'
one-step scale), the interval at level $1-\alpha$ is
$\big[\hat y_{j,t+h} + \sigma_j q_{\alpha/2,h},\ \hat y_{j,t+h} + \sigma_j
q_{1-\alpha/2,h}\big]$, with $q_{\cdot,h}$ the empirical quantiles of the pooled
$e_{\cdot,h}$. Where a level has fewer than 50
earlier errors for each horizon, all horizons are pooled into one set of
quantiles, and interval ends are rounded outwards to whole units except on
Favorita; this fix was added on 16 October after the first intervals covered
far less than planned at the totals, and is used throughout. The first
rolling window has no earlier window and is not scored. We report the scaled
pinball loss averaged over the quantiles, weighted as in the M5 uncertainty
competition~\cite{m5uncertainty}, and coverage at 50\%, 80\% and 95\%. Models
that give quantiles of their own (Chronos-Bolt, TFT) are also scored on them,
over every window.

\paragraph{Decisions}
Two simulations turn forecasts into costs. \emph{Store orders}: for each
item-store-period, order the model's interval quantile at a critical ratio $r$
of 0.5, 0.8 or 0.95; each unit over costs 1 and each unit short $r/(1-r)$.
\emph{Category plan}: for category-by-store nodes $n$ with sales $y_{n,t}$,
forecast $\hat y_{n,t}$ and average price $\bar\pi_n$ before the window,
\begin{equation}
C = \frac{\sum_n \bar\pi_n \sum_{t} |y_{n,t} - \hat y_{n,t}|}{\sum_n \bar\pi_n \sum_t y_{n,t}},
\end{equation}
the absolute error in dollars over sales in dollars. Both are reported against seasonal naive. Both are simulations under stated
costs, not business results.

\paragraph{Reconciliation}
Base forecasts at every node come from AutoETS fitted on the history before each
window (seasonal at the aggregate nodes, non-seasonal item by store). They are
reconciled bottom-up, top-down by historical proportions, and by the MinT
family~\cite{mint}. With $S$ the summing matrix (nodes by item-store series) and
$\hat{\mathbf y}$ the base forecasts at every node, the reconciled item-store
forecasts are
\begin{equation}
\tilde{\mathbf b} = (S^\top W^{-1} S)^{-1} S^\top W^{-1}\hat{\mathbf y},
\end{equation}
with $W = I$ (ordinary least squares), $W = \mathrm{diag}(S\mathbf 1)$ (structural
weights), $W$ the diagonal of in-sample error variances, or $W = \lambda
\mathrm{diag}(\hat\Sigma) + (1-\lambda)\hat\Sigma$ with the Sch\"afer--Strimmer
intensity $\lambda$ (MinT-shrink, where the hierarchy has at most 6,000 nodes:
Frat); nodes with no
in-sample error are given the smallest variance of the others. Each model's own
item-by-store forecasts are also combined with ETS above them by weighted least
squares.

\paragraph{Ranking}
Accuracy says how close a forecast is, not whether it puts items in the right
order. For every window we report the Spearman correlation, across item-store
series that sold in the window before, between the forecast change (horizon
total over the previous horizon's total) and the actual change; the share of the
tenth of nodes forecast to grow most that are in the actual top tenth (0.10 by
chance); the Spearman correlation across promoted periods between forecast and
actual units over the series' unpromoted level; and bias, forecast over actual.
These were written down before they ran and have no pass rule.

\paragraph{Dominick's categories}
Table~\ref{tab:domcats} places the six categories used among all \domCatN{}. In the first, untuned run
LightGBM was the most accurate of the three models then built on \domCatEarlyLgbWins{} categories and the world
model on \domCatEarlyWmWins{}. The six sell more per week than the median category, and a median of
\domCatDealSix\% of their units sell on a deal against \domCatDealRest\% for the other \domCatRest{}. Tuning
closed most of the gap on the four chosen as hard: the world model trailed LightGBM by \domCatEarlyGapHardLo{} to
\domCatEarlyGapHardHi{} on them in the first run, and after tuning its gap on the six runs from
\domCatTunedGapLo{} to \domCatTunedGapHi{} (behind on \domCatTunedBehind{}).
\begin{table*}[t]
\centering
\caption{\textbf{Dominick's 28 categories, the six used in bold.} Weeks the Kilts Center flags as reliable. Units and
dollars per week with sales, in thousands; units per item-store-week, median over item-store pairs; on deal: share of
units sold with a deal code; price changed: share of item-store weeks whose price differs from the previous recorded
week. First run: all 28 categories with the first, untuned settings (one seed, development windows), WRMSSE of LightGBM
and the world model and the world model's gap; rows in order of that gap; the six were chosen from this run, before
any tuning. Tuned gap: the same gap for the final tuned models on the rolling windows. Descriptive columns added in
\thisrev{} (scripts/dominicks\_eda.py).}
\label{tab:domcats}
\footnotesize\setlength{\tabcolsep}{4pt}
\begin{tabular}{lrrrrrrrrrr}
\toprule
 & & \multicolumn{2}{c}{Per week (000)} & Units per & On deal & Price & \multicolumn{3}{c}{First run, untuned} & Tuned \\
\cmidrule(lr){3-4}\cmidrule(lr){8-10}
Category & Items & Units & Dollars & item-store-week & (\%) & changed (\%) & LGB & WM & Gap & gap \\
\midrule
\textbf{Snack crackers} & 420 & 86 & 178 & 4.5 & 37 & 25 & 0.587 & 1.463 & +0.876 & +0.115 \\
Bath soap & 579 & 3 & 8 & 1.3 & 15 & 16 & 1.188 & 1.465 & +0.277 &  \\
\textbf{Cheese} & 657 & 303 & 610 & 6.9 & 35 & 28 & 0.742 & 0.996 & +0.254 & $-$0.030 \\
\textbf{Bottled juice} & 506 & 121 & 247 & 6.1 & 31 & 24 & 0.992 & 1.229 & +0.237 & $-$0.006 \\
Fabric softeners & 318 & 44 & 114 & 5.0 & 24 & 17 & 0.650 & 0.832 & +0.182 &  \\
Cigarettes & 793 & 39 & 124 & 1.7 & 0 & 5 & 0.534 & 0.690 & +0.156 &  \\
\textbf{Soft drinks} & 1,608 & 763 & 1,239 & 6.8 & 55 & 48 & 0.747 & 0.864 & +0.117 & +0.026 \\
Toilet tissue & 127 & 224 & 241 & 20.0 & 44 & 30 & 0.394 & 0.465 & +0.071 &  \\
Frozen juices & 175 & 153 & 180 & 10.0 & 41 & 29 & 0.314 & 0.376 & +0.062 &  \\
Toothbrushes & 491 & 15 & 29 & 2.3 & 24 & 23 & 0.647 & 0.699 & +0.052 &  \\
Dish detergent & 287 & 57 & 113 & 6.1 & 30 & 19 & 0.344 & 0.393 & +0.049 &  \\
Toothpastes & 608 & 35 & 77 & 3.0 & 28 & 24 & 0.731 & 0.779 & +0.048 &  \\
Cookies & 1,124 & 160 & 319 & 3.6 & 34 & 20 & 0.487 & 0.522 & +0.035 &  \\
Beer & 787 & 44 & 264 & 2.6 & 46 & 29 & 0.454 & 0.470 & +0.016 &  \\
Soaps & 334 & 49 & 102 & 3.8 & 24 & 20 & 0.517 & 0.516 & $-$0.001 &  \\
Paper towels & 163 & 162 & 165 & 22.3 & 35 & 28 & 0.486 & 0.485 & $-$0.001 &  \\
Front-end candies & 503 & 179 & 96 & 8.5 & 19 & 12 & 0.414 & 0.397 & $-$0.017 &  \\
Crackers & 330 & 47 & 89 & 3.2 & 35 & 25 & 0.709 & 0.687 & $-$0.022 &  \\
Oatmeal & 96 & 33 & 78 & 4.8 & 25 & 18 & 1.084 & 1.060 & $-$0.024 &  \\
Canned soup & 445 & 275 & 233 & 8.2 & 24 & 19 & 0.613 & 0.583 & $-$0.030 &  \\
\textbf{Frozen entrees} & 898 & 201 & 360 & 6.3 & 45 & 28 & 0.417 & 0.382 & $-$0.035 & +0.018 \\
\textbf{Analgesics} & 638 & 20 & 98 & 1.5 & 14 & 13 & 0.702 & 0.666 & $-$0.036 & $-$0.003 \\
Frozen dinners & 266 & 53 & 105 & 5.5 & 46 & 34 & 0.930 & 0.873 & $-$0.057 &  \\
Cereals & 489 & 248 & 716 & 12.2 & 22 & 16 & 0.753 & 0.673 & $-$0.080 &  \\
Laundry detergents & 581 & 77 & 376 & 5.3 & 37 & 19 & 0.768 & 0.687 & $-$0.081 &  \\
Shampoos & 2,930 & 47 & 127 & 1.5 & 29 & 25 & 0.994 & 0.906 & $-$0.088 &  \\
Grooming products & 1,381 & 38 & 110 & 1.9 & 25 & 24 & 0.901 & 0.809 & $-$0.092 &  \\
Canned tuna & 278 & 113 & 136 & 5.3 & 36 & 19 & 1.163 & 0.959 & $-$0.204 &  \\
\bottomrule
\end{tabular}
\end{table*}

\section{Searching the settings in detail}
\label{sec:loopdetail}
% Detail of the searches: in the main text for arXiv, in the online supplement for the journal.

\begin{topfigure}
\centering
\includegraphics[width=\textwidth]{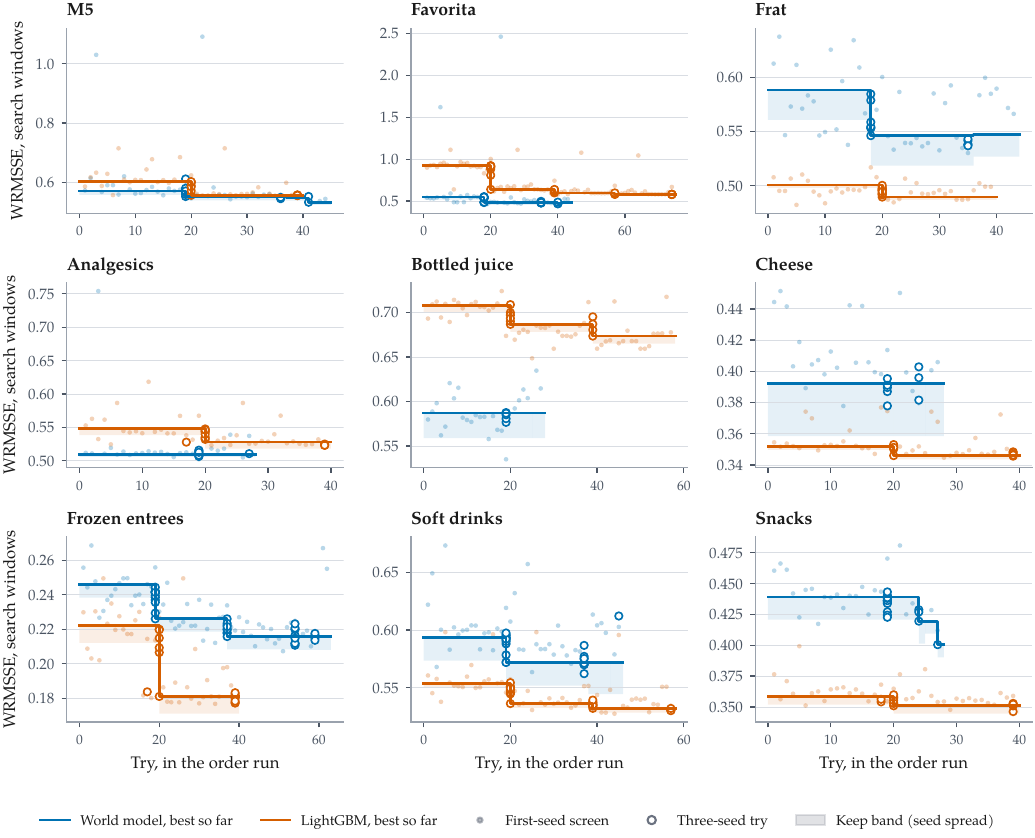}
\caption{Every run of every search, in the order it was made. Faint dots are first-seed screens,
open circles are three-seed confirmations, and the step line is the best settings so far. The band
under the line is the keep bar: a change had to land below it to be kept. The world model's line
runs on into its likelihood and JEPA rounds after its own loop ends.}
\label{fig:looppath}
\end{topfigure}

\subsection{What the searches kept}

The searches made \loopTries{} runs (a run split across machines counts once here; the run-time table counts each piece of a split run as a run), \loopTriesWm{} of the world model and \loopTriesLgb{} of LightGBM, and kept \loopKept{} changes, \loopKeptWm{} for the world model and \loopKeptLgb{} for LightGBM.
LightGBM kept at least one change on every data set and the world model kept nothing on three of
the six Dominick's categories. No single change helped the world model everywhere
(Figure~\ref{fig:loopheat}). Weighting the loss toward high-volume series made the first seed worse on most data sets. The seven-step latent
check, kept on Favorita, was among the changes that moved Favorita most and did little elsewhere.

On the search windows the world model improved on \loopImprovedWm{} of the nine data sets and LightGBM on all
\loopImprovedLgb{} (Figure~\ref{fig:loopslopes}). The largest gain of either model was LightGBM on \loopGainLgbSet{},
from \loopGainLgbFrom{} to \loopGainLgbTo{}. The world model's largest was on \loopGainWmSet{}, from \loopGainWmFrom{} to \loopGainWmTo{}.

\begin{figure}[t]
\centering
\includegraphics[width=\colfig]{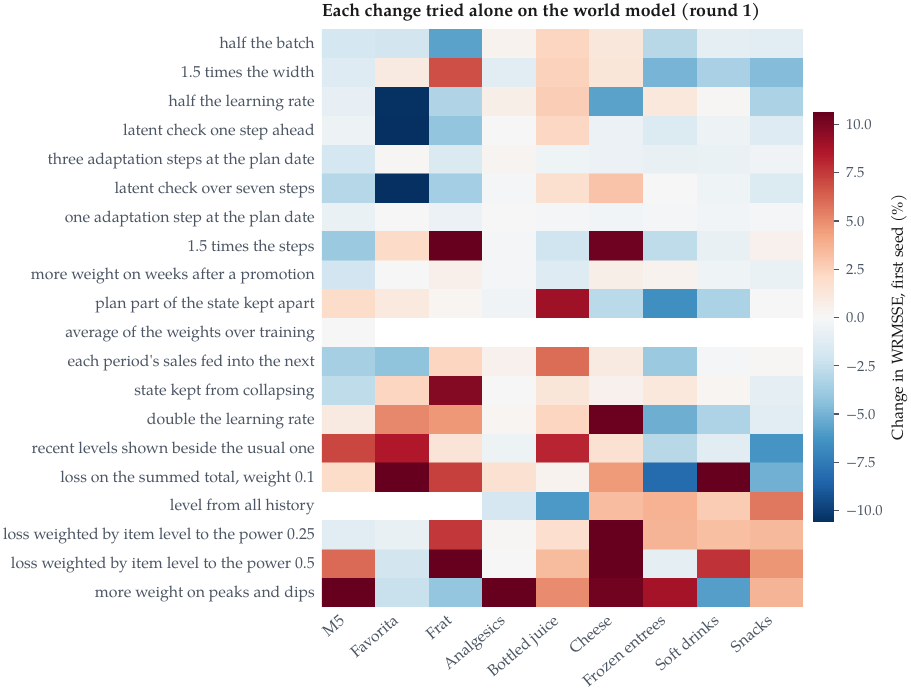}
\caption{Each change tried alone on the world model in the first round, against that data set's
starting settings on the first seed. Blue is better. A blank cell is a change that was not on that
data set's menu.}
\label{fig:loopheat}
\end{figure}

\subsection{Did the held-out window agree?}

The two agreed only in part. Across every three-seed run, the change on the search windows and the change on the
held-out window are correlated at \loopHeldCorrWm{} for the world model and \loopHeldCorrLgb{} for LightGBM
(Figure~\ref{fig:loopheldout}). A change that helps on the search windows is a weak guide to the
next window. The world model's final settings were worse than its starting settings on the
held-out window on \loopHeldWorseWm{} data sets, most on \loopHeldWorstWmSet{} (\loopHeldWorstWmFrom{} to \loopHeldWorstWmTo{}), and
LightGBM's on \loopHeldWorseLgb{}, most on \loopHeldWorstLgbSet{} (\loopHeldWorstLgbFrom{} to \loopHeldWorstLgbTo{}). This is why the paper reports the final settings on windows that the
searches never saw, in Section~\ref{sec:results}, and not the search scores.

\begin{figure}[t]
\centering
\includegraphics[width=\colfig]{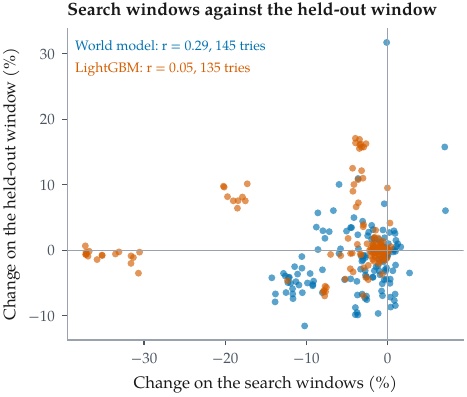}
\caption{Each three-seed run's change on the search windows against its change on the held-out
window, both against the starting settings of its search.}
\label{fig:loopheldout}
\end{figure}

\subsection{Seed noise, one copy against five}
\label{sec:loopnoise}

A single world-model fit is noisy enough to hide most of the changes the loop tries. On the six
Dominick's categories the three seeds of the starting settings differ from their mean by a
standard deviation of \loopSeedSdOne\% for a single fit and \loopSeedSdFive\% for a five-copy average
(Figure~\ref{fig:loopnoise}). The
single-fit search on cheese had a bar of \loopOneCheBar{} and kept nothing; with the average the bars on the
Dominick's categories were between \loopBarLoWm{} and \loopBarHiWm{}. LightGBM's seeds differ much less, and its
bars were between \loopBarLoLgb{} and \loopBarHiLgb{}.

\begin{figure}[t]
\centering
\includegraphics[width=\colfig]{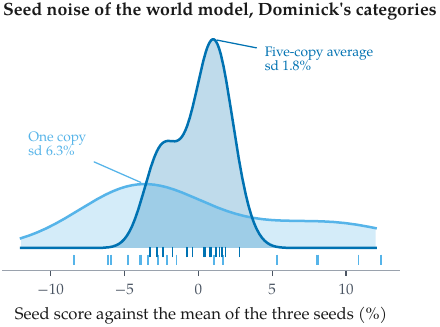}
\caption{The starting settings' three seed scores on each Dominick's category, as a share of their
mean, for a single world-model fit and for the five-copy average. Ticks are the scores.}
\label{fig:loopnoise}
\end{figure}

\begin{figure*}[t]
\centering
\includegraphics[width=\textwidth]{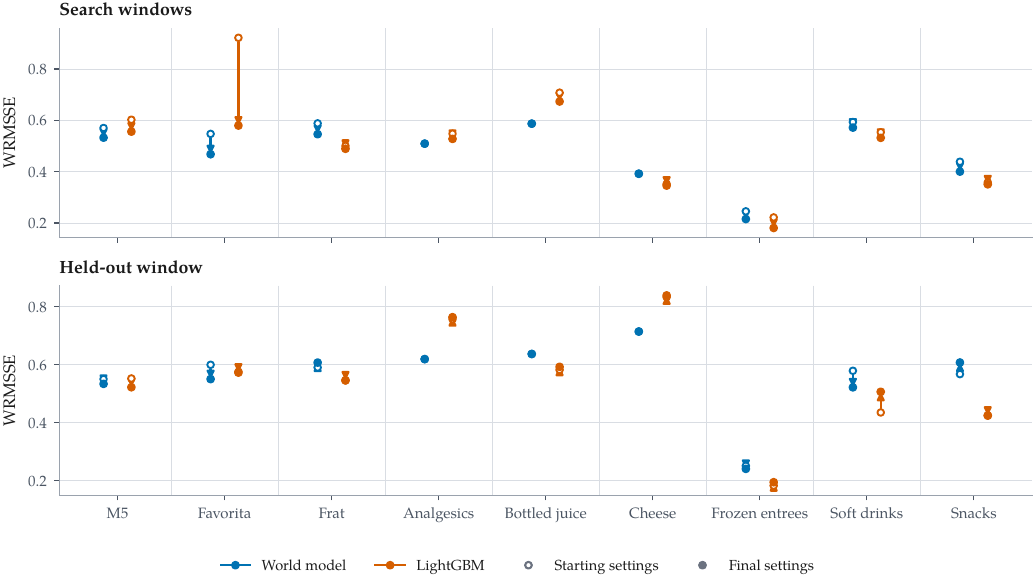}
\caption{Each model's starting and final settings on each data set. Open circles are the starting
settings and filled circles the final settings, on the search windows (top) and the held-out
window (bottom).}
\label{fig:loopslopes}
\end{figure*}

% Designs and models added after the planned 52-week test (Section~\ref{sec:long-horizon}): in the results of the
% arXiv version, a section of the journal supplement.
\ifsupplement\section{Forecasting 52 weeks ahead: designs and models added after the test}
\label{sec:long-horizon-more}
The planned 52-week test, its table and figure are in Section~\ref{sec:long-horizon} of the paper. This section
gives what was added after it.
\fi

\paragraph{Added after the planned test: other designs and other models}
After seeing the planned result we added two designs in which no model takes in
its own forecasts, and ran the other models of the paper over 52 weeks
(Table~\ref{tab:long-horizon-all}). In the first, the world model rolls its
state forward 13 weeks at a time on the plan; in the second, LightGBM fits one
model per block of 13 weeks. Rolling on helps on M5 and fails on Dominick's. On M5
the world model rolled on scores \lhMfiveWmRollon{} against \lhMfiveWmBlocks{} in
blocks, and its average with LightGBM, \lhMfiveAvgRollon, is the most accurate
forecast on M5 and clearly more accurate than LightGBM forecasting directly
(\lhMfiveLgbDirect) in \lhMfiveAvgRollonLgbBetter{} of
\lhMfiveAvgRollonLgbN{} paired comparisons, clearly less in none. On the
Dominick's categories the rolled-on state drifts: on cheese the forecast is
\lhRollonCheeseFa{} times what sold and the WRMSSE \lhRollonCheese. LightGBM by
block is close to LightGBM forecasting directly. Of all the forecasts in Table~\ref{tab:long-horizon-all}, the average of the
two models forecasting directly is the most accurate on \lhAllLowestAvgDirect{}
data sets, and the average with the world model rolled on on M5. On analgesics
and soft drinks the most accurate is LightGBM fitted per block with settings
searched again for 52 weeks, and on cheese the average of that LightGBM with the
world model; both were added after the planned test. TFT, TimesFM, Chronos and the seasonal naive forecast are
behind LightGBM and the average on every data set; TimesFM given the plan is
ahead of the world model alone on analgesics. Timestral given the plan cannot
forecast 52 weeks: its forecast is \lhTimestralFaOneLo{} to \lhTimestralFaOneHi{}
times what sold in the first 13 weeks and \lhTimestralFaLo{} to
\lhTimestralFaHi{} times over the year. On M5 the world model forecasting the whole year at once is ahead of LightGBM in the later
window (\lhMfiveWmDirLate{} against \lhMfiveLgbDirLate) and behind in the earlier (\lhMfiveWmDirEarly{}
against \lhMfiveLgbDirEarly).

\paragraph{Settings searched again for 52 weeks (added after the planned test)}
Each model's settings were searched again for 52 weeks on the six Dominick's
categories, with the paper's search (Section~\ref{sec:loop}) on three 52-week
windows 13 weeks apart. The world model's searches kept one change, on snacks;
LightGBM's kept changes on five categories,
none on cheese. Retuned, the world model is clearly less accurate than LightGBM
in \lhPTunedWmLgbWorse{} of \lhPTunedWmLgbN{} paired comparisons and clearly more
in \lhPTunedWmLgbBetter; the average of the two is clearly more accurate than
LightGBM in \lhPTunedAvgLgbBetter{} and clearly less in \lhPTunedAvgLgbWorse.
Retuning did not help LightGBM on balance: its retuned forecast is clearly more
accurate than the first in \lhPTunedLgbBetter{} comparisons and clearly less in
\lhPTunedLgbWorse.

\begin{table*}[t]
\centering\footnotesize\setlength{\tabcolsep}{3pt}
\caption{Every forecast 52 weeks ahead (364 days on M5): WRMSSE over the 52 weeks, mean over levels and two windows.
Each forecasts all 52 weeks at once except the world model rolled on (its state rolled 13 weeks at a time on the plan,
without taking in its own forecasts) and LightGBM by block (one model per block of 13 weeks). Averages: the world
model with LightGBM directly (Avg; with the world model rolled on, the M5 average) or by block. Tuned: settings searched again for 52 weeks on Dominick's. Bold: lowest
of each row, leaving out Timestral with the plan, which could not forecast 52 weeks. Of these, only WM, LGB and Avg
were in the planned test; the rest were added after it ran.}
\label{tab:long-horizon-all}
\begin{adjustbox}{max width=\textwidth}\begin{tabular}{lcccccccccccccccccc}
\toprule
Data set & \rotatebox{90}{WM} & \rotatebox{90}{WM rolled on} & \rotatebox{90}{LGB} & \rotatebox{90}{LGB by block} & \rotatebox{90}{Avg} & \rotatebox{90}{Avg, WM rolled on} & \rotatebox{90}{Avg by block} & \rotatebox{90}{WM tuned} & \rotatebox{90}{LGB tuned} & \rotatebox{90}{LGB tuned by block} & \rotatebox{90}{Avg tuned} & \rotatebox{90}{Avg tuned by block} & \rotatebox{90}{Seasonal naive} & \rotatebox{90}{Chronos} & \rotatebox{90}{TimesFM} & \rotatebox{90}{TimesFM with plan} & \rotatebox{90}{Timestral with plan} & \rotatebox{90}{TFT}\\
\midrule
Analgesics & 0.964 & 1.062 & 0.867 & 0.846 & 0.868 & 0.892 & 0.888 & 0.964 & 0.853 & \textbf{0.837} & 0.866 & 0.862 & 1.909 & 1.065 & 0.988 & 0.909 & 89.128 & 0.987 \\
Bottled juice & 0.764 & 1.174 & 0.769 & 0.765 & \textbf{0.726} & 0.886 & 0.889 & 0.764 & 0.768 & 0.760 & 0.734 & 0.731 & 1.493 & 0.936 & 0.911 & 0.858 & 33.123 & 0.821 \\
Cheese & 0.773 & 5.027 & 0.707 & 0.680 & 0.677 & 2.703 & 2.696 & 0.773 & 0.707 & 0.680 & 0.677 & \textbf{0.667} & 1.395 & 0.942 & 0.903 & 0.798 & 18.044 & 0.810 \\
Frozen entrees & 0.412 & 0.888 & 0.393 & 0.399 & \textbf{0.387} & 0.581 & 0.586 & 0.412 & 0.395 & 0.403 & 0.388 & 0.391 & 0.803 & 0.628 & 0.594 & 0.547 & 4.591 & 0.498 \\
Soft drinks & 0.851 & 1.024 & 0.754 & 0.754 & 0.767 & 0.824 & 0.825 & 0.851 & 0.751 & \textbf{0.745} & 0.763 & 0.762 & 1.230 & 0.987 & 0.951 & 0.847 & 11.052 & 0.878 \\
Snacks & 0.866 & 0.932 & 0.882 & 0.866 & \textbf{0.843} & 0.865 & 0.859 & 0.866 & 0.933 & 0.908 & 0.858 & 0.846 & 1.747 & 1.040 & 0.958 & 0.896 & 18.321 & 1.102 \\
M5 & 0.922 & 0.880 & 0.839 & 0.849 & 0.796 & \textbf{0.764} & 0.771 & n/a & n/a & n/a & n/a & n/a & 1.064 & 1.843 & 1.708 & 1.813 & 2.972 & 1.204 \\
\bottomrule
\end{tabular}\end{adjustbox}
\end{table*}

% =====================================================================
\section{Every model compared}\seclabel{sec:support}

\label{sec:sup-comparison}
Every analysis in this and the next five sections (the ablation, the occlusion,
the carried-state tests, the calibration and reliability of the forecast ranges,
and the decision costs) was rerun on the tuned models of
Table~\ref{tab:final-settings}, on the same rolling windows as
Section~\ref{sec:results}; none of its numbers comes from the settings of the
\firstver{}.

\textbf{The average of the tuned pair is more accurate than every model not
tuned on the data, on every data set; TimesFM given the plan beats the world
model alone on snacks and Frat, and LightGBM alone on Frat.} Table~\ref{tab:sup-comparison}
adds the pretrained models, the Temporal Fusion Transformer and seasonal naive.
TimesFM given the plan is the most accurate of the models not trained on the
data on \supTfmPlanBestUntrained{} data sets. On Frat it is more accurate than
either tuned model alone (\supFratTfmPlan{} against \supFratWm{} for the world
model and \supFratLgb{} for LightGBM) but not than their average
(\supFratAvg{}); on snacks it is more accurate than the world model alone. Timestral given the plan forecasts far above sales on the
weekly data, where a few extreme upper quantiles dominate the mean of its
output: \supAnaTstPlan{} on analgesics against \supAnaTfmPlan{} for TimesFM given
the plan.

\begin{table*}[t]
\centering\footnotesize\setlength{\tabcolsep}{3pt}
\caption{Every model on the rolling windows: WRMSSE averaged over all levels, mean over windows. WM and LGB: the tuned
world model and LightGBM; Avg: their average; Ens: the three-model ensemble with TimesFM given the plan; TFM+ and
TFM: TimesFM with and without the plan; Chr: Chronos-Bolt; TST+ and TST: Timestral with and without the plan; TFT:
the Temporal Fusion Transformer, with the settings of the \firstver{}; SN: seasonal naive. The comparison models
are scored on cleaned prices on Dominick's; on M5, Frat and Favorita they are the runs of the \firstver{} on the
same windows, since nothing in them was tuned; the models that version left off (the Temporal Fusion Transformer on M5
and Favorita, Chronos-Bolt and TimesFM on Favorita) were run for \thisrev{} with the same settings. Shading marks the lowest: \colorbox{wmcell}{world model}, \colorbox{lgbcell}{LightGBM}, \colorbox{avgcell}{average}.}
\label{tab:sup-comparison}
\begin{adjustbox}{max width=\textwidth}\begin{tabular}{lccccccccccc}
\toprule
Data set & WM & LGB & Avg & Ens & TFM+ & TFM & Chr & TST+ & TST & TFT & SN\\
\midrule
Analgesics & 0.699 & 0.702 & 0.687 & \textbf{0.676} & 0.723 & 0.804 & 0.846 & 13.892 & 1.341 & 0.817 & 1.374 \\
Bottled juice & 0.575 & 0.581 & 0.551 & \textbf{0.551} & 0.673 & 0.791 & 0.802 & 3.889 & 1.012 & 0.741 & 1.279 \\
Cheese & 0.502 & 0.532 & 0.477 & \textbf{0.471} & 0.612 & 0.727 & 0.765 & 3.437 & 0.770 & 0.600 & 1.187 \\
Frozen entrees & 0.357 & 0.338 & \cellcolor{avgcell}\textbf{0.332} & 0.349 & 0.476 & 0.529 & 0.569 & 1.781 & 0.571 & 0.457 & 0.677 \\
Soft drinks & 0.705 & 0.680 & \cellcolor{avgcell}\textbf{0.650} & 0.655 & 0.770 & 0.909 & 0.976 & 2.281 & 0.858 & 0.815 & 1.001 \\
Snacks & 0.893 & \cellcolor{lgbcell}\textbf{0.778} & 0.790 & 0.795 & 0.855 & 1.079 & 2.316 & 8.666 & 3.277 & 0.959 & 1.399 \\
Frat & 0.490 & 0.494 & 0.469 & \textbf{0.454} & 0.482 & 0.691 & 0.722 & 1.002 & 0.685 & 0.571 & 1.038 \\
M5 & 0.709 & 0.697 & \cellcolor{avgcell}\textbf{0.658} & 0.729 & 0.982 & 0.964 & 1.184 & 1.038 & 0.988 & 1.155 & 1.146 \\
Favorita & 0.428 & 0.448 & \cellcolor{avgcell}\textbf{0.403} & 0.419 & 0.501 & 0.518 & 0.595 & 0.515 & 0.548 & 0.901 & 0.662 \\
\bottomrule
\end{tabular}\end{adjustbox}
\end{table*}

\secbarrier
\section{Ablation: what each model adds to the combination}
\label{sec:sup-ablation}

\textbf{Each tuned model makes the combination more accurate, and TimesFM adds a
little at item by store.} Table~\ref{tab:sup-ablation} drops each member of the
three-model ensemble in turn. Without the world model the ensemble is less
accurate item by store on \supNoWmWorse{} of \supAblationSets{} data sets, and
clearly less accurate in \supPNoWmWorse{} of \supPNoWmN{} window comparisons
(clearly more accurate in \supPNoWmBetter{}); without LightGBM, in
\supPNoLgbWorse{} (clearly more accurate in \supPNoLgbBetter{}). Without TimesFM,
which leaves the average of the two tuned models, it is clearly less accurate item
by store in \supPNoTfmWorse{} comparisons and clearly more accurate in
\supPNoTfmBetter{}. Item by store, then, adding TimesFM with the plan to the
average helps; over all levels the average is in the model confidence set more
often (Section~\ref{sec:rq-grain}). A retailer whose main use is store orders can
add it at the cost of a third model to run. Weights fitted on earlier windows to
the item-by-store error beat equal weights in \supPFitBottomBetter{} of
\supPFitBottomN{} comparisons and lost in \supPFitBottomWorse{}; weights fitted to
the error over all levels beat them in \supPFitAllBetter{} and lost in
\supPFitAllWorse{}.

What the average needs is a second model unlike LightGBM, and the world model is
not the only one. Item by store, the three-model ensemble is lower than the
average of the two tuned models on \ptThreeBelowAvg{} of \ptSets{} data sets and
higher on \ptThreeAboveAvg; LightGBM averaged with TimesFM given the plan, with no
world model, is level with or lower than that average on \ptNoWmAtOrBelowAvg{} and
higher on \ptNoWmAboveAvg. The world model's own case rests where no other partner
was tested to stand in for it: items the models were not trained on
(Section~\ref{sec:rq-unseen}) and lines sold at a steady price
(Section~\ref{sec:rq-lines}).

Figure~\ref{fig:sup-ablation}a shows the same comparison data set by data set.
Dropping the world model raises the item-by-store error on \sfAblNoWmWorse{}
data sets, dropping LightGBM on \sfAblNoLgbWorse{} and dropping TimesFM on
\sfAblNoTfmWorse{}; fitted weights lower the error most on frozen entrees,
bottled juice and snacks. Fitted to the item-by-store error, the mean
weights are \sfWtWmBottomMean{} for the world model, \sfWtLgbBottomMean{} for
LightGBM and \sfWtTfmBottomMean{} for TimesFM given the plan
(Figure~\ref{fig:sup-ablation}b); fitted to the total, TimesFM's weight falls to
\sfWtTfmTotalMean{}, and the weights move from data set to data set.

\begin{table}[t]
\centering\footnotesize\setlength{\tabcolsep}{3pt}
\caption{What each member adds to the three-model ensemble: WRMSSE item by store, mean over the rolling windows, of
the ensemble, of the ensemble without each member (without TimesFM it is the average of the two tuned models), and of
the ensemble with weights fitted on earlier windows. Lowest in bold.}
\label{tab:sup-ablation}
\begin{adjustbox}{max width=\columnwidth}\begin{tabular}{lccccc}
\toprule
Data set & Three & No WM & No LGB & No TFM & Fitted\\
\midrule
Analgesics & 0.655 & 0.655 & 0.664 & 0.655 & \textbf{0.651} \\
Bottled juice & 0.584 & 0.598 & 0.589 & 0.599 & \textbf{0.558} \\
Cheese & 0.599 & 0.609 & 0.617 & 0.605 & \textbf{0.598} \\
Frozen entrees & 0.401 & 0.416 & 0.422 & 0.395 & \textbf{0.336} \\
Soft drinks & \textbf{0.547} & 0.550 & 0.567 & 0.561 & 0.568 \\
Snacks & 0.556 & 0.559 & 0.576 & 0.556 & \textbf{0.531} \\
Frat & \textbf{0.544} & 0.547 & 0.555 & 0.550 & 0.549 \\
M5 & \textbf{0.832} & 0.839 & \textbf{0.832} & 0.836 & 0.849 \\
Favorita & 0.687 & 0.702 & 0.685 & 0.689 & \textbf{0.678} \\
\bottomrule
\end{tabular}\end{adjustbox}
\end{table}

\begin{figure*}[t]
\centering
\includegraphics[width=\textwidth]{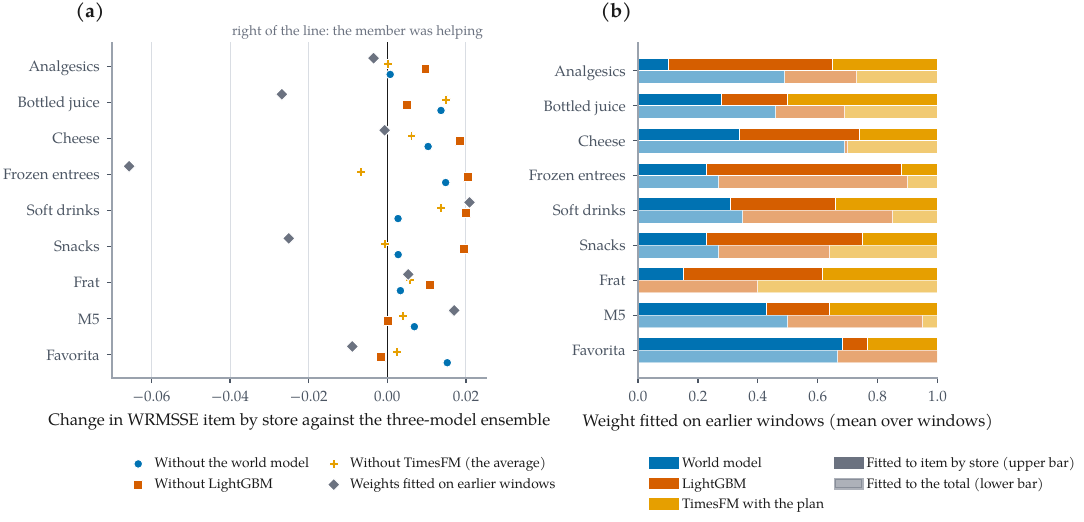}
\caption{What each member adds to the three-model ensemble. (a) Change in WRMSSE
item by store, mean over windows, when one member is dropped or the weights are
fitted on earlier windows, against the ensemble with equal weights; right of the
line, the dropped member was helping. (b) Weights fitted on earlier windows, mean
over windows, when fitted to the item-by-store error (upper bar) and to the total
(lower bar).}
\label{fig:sup-ablation}
\end{figure*}

\secbarrier
\section{Occlusion: which plan inputs each model uses}
\label{sec:sup-plan}

\textbf{Both tuned models forecast from the plan.} Setting the price to the
regular price makes the item-by-store error of the world model worse on
\supPertPriceHelpsWm{} of \supPertSets{} data sets and of LightGBM on
\supPertPriceHelpsLgb{} (Favorita carries no price, so its forecasts do not
change); setting every plan input to nothing planned makes it
worse on \supPertPlanHelpsWm{} and \supPertPlanHelpsLgb{}
(Table~\ref{tab:sup-perturbation}). With the plan removed, the world model's
forecast on the rows where the plan had something on falls by
\supPertUnitsWmLo\% to \supPertUnitsWmHi\%, and LightGBM's by
\supPertUnitsLgbLo\% to \supPertUnitsLgbHi\%: the two read about the same
lift from a promotion. Cutting the world model's history to the last few periods
makes its error worse on \supPertHistHelps{} of \supPertSets{}, so its carried
state draws on more than the latest sales.

Figure~\ref{fig:sup-occlusion} breaks the plan into its parts. On the
Dominick's categories and Frat, price and the promotion flag carry the plan;
display, feature advertising and a price cut on its own add less on Frat, the one
data set that records them. The promotion flag helps the world model on
\sfPromoWmHelps{} of the \sfPromoSets{} data sets that carry it and LightGBM on
\sfPromoLgbHelps{}. Events, SNAP days and paydays change the error by at most
\sfCalendarMax{}. Shuffling an input across the periods of each series, which
keeps how often it is on but breaks its timing, moves the error the same way as
switching it off for \sfPermAgree{} of \sfPermPairs{} pairs of input and data
set. Keeping only the last few periods of history raises the world model's error
on \sfHistShortWorse{} of \sfHistSets{} data sets at the shortest cut
(Figure~\ref{fig:sup-occlusion}c). The cut lengths are 4, 13 and 26 periods on
every data set but M5, which used 28 and 56 days; on Favorita, which is daily,
the shortest keeps four days and the error rises by \sfFavHistFour{}.

\begin{table*}[t]
\centering\footnotesize\setlength{\tabcolsep}{4pt}
\caption{Which plan inputs each tuned model uses. Change in WRMSSE item by store, mean over windows, when the price
is set to the regular price (price off) and when every plan input is set to nothing planned (plan off); the change
in forecast units on the rows where the plan had something on; and, for the world model, the largest change when its
history before the last few periods is removed. A positive change means the input was helping.}
\label{tab:sup-perturbation}
\begin{tabular}{l ccc ccc c}
\toprule
 & \multicolumn{3}{c}{World model} & \multicolumn{3}{c}{LightGBM} & World model\\
\cmidrule(lr){2-4}\cmidrule(lr){5-7}\cmidrule(lr){8-8}
Data set & Price off & Plan off & Units, plan rows & Price off & Plan off & Units, plan rows & History cut\\
\midrule
Analgesics & +0.019 & +0.061 & -47\% & +0.037 & +0.075 & -51\% & +0.029 \\
Bottled juice & +0.065 & +0.095 & -53\% & +0.072 & +0.098 & -55\% & +0.031 \\
Cheese & +0.095 & +0.125 & -52\% & +0.136 & +0.161 & -58\% & +0.076 \\
Frozen entrees & +0.100 & +0.130 & -73\% & +0.129 & +0.162 & -72\% & +0.024 \\
Soft drinks & +0.073 & +0.127 & -61\% & +0.168 & +0.195 & -73\% & +0.019 \\
Snacks & +0.065 & +0.096 & -37\% & +0.089 & +0.116 & -44\% & +0.059 \\
Frat & +0.110 & +0.229 & -46\% & +0.117 & +0.233 & -45\% & +0.081 \\
M5 & +0.001 & +0.023 & -7\% & +0.001 & +0.006 & -4\% & +0.044 \\
Favorita & +0.000 & +0.037 & -33\% & +0.000 & +0.040 & -28\% & +8.158 \\
\bottomrule
\end{tabular}
\end{table*}

\begin{topfigure}
\centering
\includegraphics[width=\textwidth]{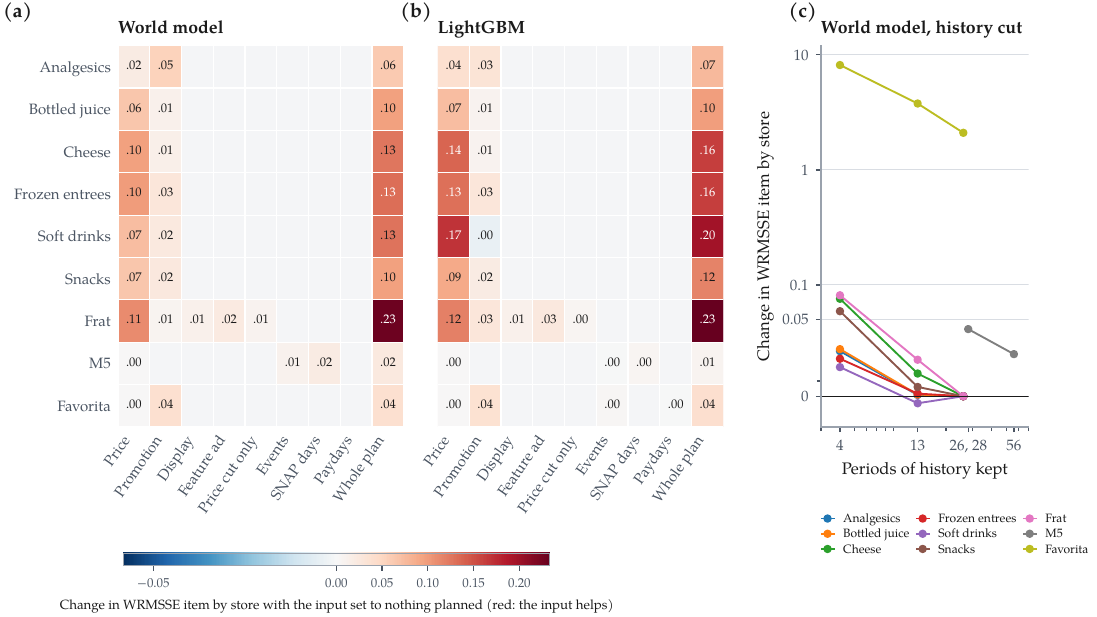}
\caption{Which plan inputs the tuned models use. (a, b) Change in WRMSSE item by
store, mean over windows, when one input is set to nothing planned (the price to
the regular price), for the world model and LightGBM; grey cells are inputs a
data set does not carry. (c) The world model's change in WRMSSE item by store
when its history is cut to the last 4, 13 or 26 periods (M5: 28 or 56 days); the
vertical scale is linear below 0.05 and logarithmic above.}
\label{fig:sup-occlusion}
\end{topfigure}

\secbarrier
\section{Carried-state tests: what the world model's state does}
\label{sec:sup-mechanism}

\textbf{The tuned world model uses the timing of the plan, and carries a
promotion forward as more sales rather than a dip.} Shuffling the plan's timing
within each series makes its item-by-store error worse on all \supMechSets{}
data sets, by \supTimingLo\% to \supTimingHi\% (Table~\ref{tab:sup-mechanism}):
the forecast reads when a promotion runs, not only that one runs. A promotion and
a price cut added in one period raise that period's forecast and still raise the
next on \supPersistRiseSets{} of the \supPersistSets{} data sets where it can be read (the change in the next
period runs from \supPersistTwoLo\% to +\supPersistTwoHi\% of the forecast
without them), where shoppers who stocked up would buy less; the dip
after a promotion is not in the forecast, as in the \firstver{}. The state
rolled through the window predicts sales after it with an $R^2$ only
\supProbeGainMin{} to \supProbeGainMax{} higher than the state before it.

\begin{table}[t]
\centering\footnotesize\setlength{\tabcolsep}{3pt}
\caption{The tuned world model's carried state, mean over windows. Timing: change in WRMSSE item by store when the
plan's timing is shuffled within each series. Carry-over: forecast change in periods 1, 2 and 4 after a promotion and a
20\% price cut added in period 1 to series with nothing planned, as a share of the forecast without them. Probe: $R^2$
of log sales after the window from the state before the window and from the state rolled through it (first of the
five copies).}
\label{tab:sup-mechanism}
\begin{adjustbox}{max width=\columnwidth}\begin{tabular}{lcccccc}
\toprule
Data set & Timing & Period 1 & Period 2 & Period 4 & Probe before & Probe after\\
\midrule
Analgesics & +16\% & +199\% & +13\% & +11\% & 0.80 & 0.81 \\
Bottled juice & +29\% & +188\% & +6\% & +9\% & 0.86 & 0.88 \\
Cheese & +34\% & +109\% & -4\% & +3\% & 0.91 & 0.92 \\
Frozen entrees & +64\% & +748\% & +54\% & +17\% & 0.80 & 0.84 \\
Soft drinks & +46\% & +327\% & +34\% & +10\% & 0.85 & 0.86 \\
Snacks & +24\% & +129\% & +10\% & +9\% & 0.86 & 0.87 \\
Frat & +52\% & +59\% & -2\% & -1\% & 0.86 & 0.89 \\
M5 & +2\% & n/a & n/a & n/a & 0.64 & 0.65 \\
Favorita & +5\% & +48\% & +7\% & +4\% & 0.75 & 0.77 \\
\bottomrule
\end{tabular}\end{adjustbox}
\end{table}

\secbarrier
\section{Calibration and reliability of the forecast ranges}
\label{sec:sup-calibration}

\begin{topfigure}
\centering
\includegraphics[width=\textwidth]{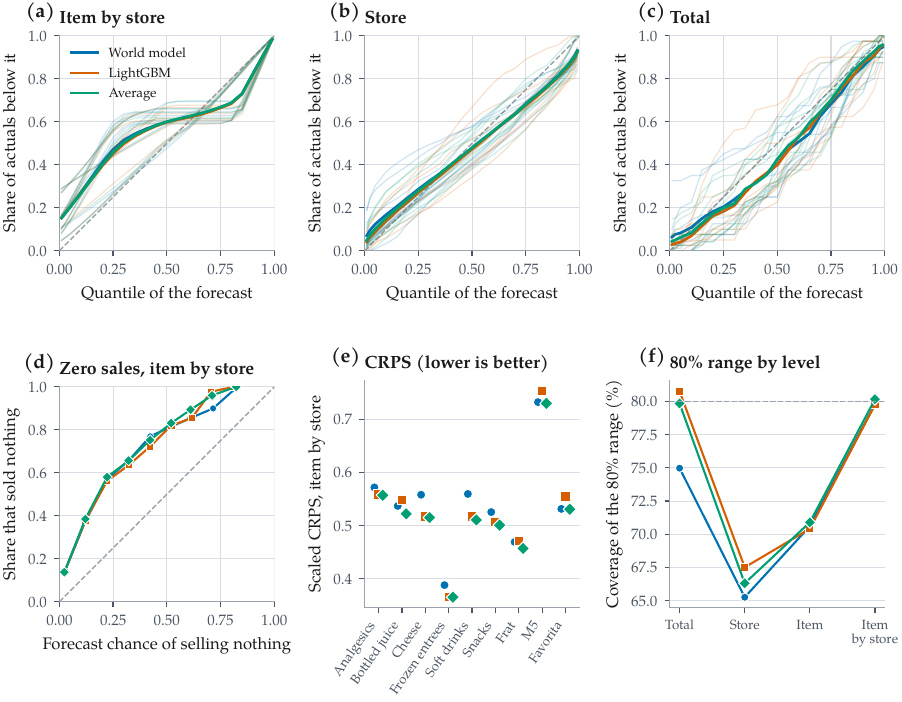}
\caption{Reliability of the whole forecast distribution. (a to c) Share of
actuals below each quantile of the forecast, item by store, at the store and at
the total, for the world model, LightGBM and the average; faint lines are data
sets, solid lines their mean, and the dashed line is perfect calibration. For
whole-unit sales each share is the midpoint of the shares strictly below and at
or below the quantile. (d) Reliability of the forecast chance of selling
nothing, item by store: node-periods grouped by their forecast chance in tenths,
pooled over data sets and windows (groups of fewer than 100 left out). (e)
Scaled CRPS item by store, mean over windows. (f) Coverage of the 80\% range by
level, mean over the nine data sets; the dashed line is the stated 80\%.}
\label{fig:sup-probabilistic}
\end{topfigure}

\textbf{Item by store the ranges hold their stated coverage on average over the
data sets; data set by data set, and at the total for the world model, they do
not.}
Each model's ranges come from its own errors in the earlier windows
(Section~\ref{sec:setup}). Item by store, the 80\% ranges of all three forecasts
cover close to 80\% of sales across data sets and windows (\supCovIxsWm\%,
\supCovIxsLgb\% and \supCovIxsAvg\% for the world model, LightGBM and the
average), and the average has the lowest scaled pinball loss (\supSplIxsAvg{},
against \supSplIxsWm{} and \supSplIxsLgb{}). At the chain total, where each window
adds one error to learn from, the world model's ranges are too narrow
(\supCovTotalWm\% for 80\%) and the average's are close to target
(\supCovTotalAvg\%); the coverage of a single data set swings with so few earlier
windows (Table~\ref{tab:sup-intervals}).

Item by store, the 50\% ranges hold \sfCovFiftyIxsWm\%,
\sfCovFiftyIxsLgb\% and \sfCovFiftyIxsAvg\% of sales for the world model,
LightGBM and the average, and the 95\% ranges \supCovNinetyFiveIxsWm\%,
\supCovNinetyFiveIxsLgb\% and \supCovNinetyFiveIxsAvg\%, so all three lie on the
diagonal on average. Data set by data set the 80\% coverage item by store runs
from \sfCovEightyMinIxsAvg\% to \sfCovEightyMaxIxsAvg\% for the average. At the
total the world model's 80\% coverage runs from \sfCovEightyMinTotalWm\% to
\sfCovEightyMaxTotalWm\% across data sets; a single window at the total holds
one actual against one range, so its coverage is anywhere from
\sfCovTotalWindowMin\% to \sfCovTotalWindowMax\%, and the
\sfCalWindows{} to \sfCalWindowsMax{} scored windows of each data set give a coarse
estimate. By level (Figure~\ref{fig:sup-probabilistic}f), the 80\% ranges are too
narrow at the store (\sfCovLevelStoreWm\%, \sfCovLevelStoreLgb\% and
\sfCovLevelStoreAvg\%) and for the item across stores (\sfCovLevelItemWm\%,
\sfCovLevelItemLgb\% and \sfCovLevelItemAvg\%): the earlier windows' errors at
those levels understate the next window's. A planner who orders from the 80\%
range at item by store gets what it says; one who plans store labour from the
store-level range needs a wider one.

\paragraph{The whole distribution} Coverage checks three ranges;
Figure~\ref{fig:sup-probabilistic} checks every quantile of the forecast
distribution and scores it whole (Table~\ref{tab:sup-probabilistic}). Item by
store, \sfRelTenAvg\% of actuals fall below the average's 10th percentile,
\sfRelHalfAvg\% below its median and \sfRelNinetyAvg\% below its 90th: the
lower quantiles of a slow item are zero and many actuals are zero too, so the
curve is read at the midpoint of the shares strictly below and at or below each
quantile. At the total, \sfRelTenTotWm\% of actuals fall below the world model's
10th percentile and \sfRelNinetyTotWm\% below its 90th; for LightGBM the shares
are \sfRelTenTotLgb\% and \sfRelNinetyTotLgb\%. The ranges at the total lie
close to the diagonal with the upper tail short; item by store the
upper tail is shorter, with \sfRelNinetyAvg\% of actuals below the 90th
percentile.
On the continuous ranked probability score (CRPS), which scores the whole
distribution, the average is lowest item by store on \sfCrpsAvgBestIxs{} of
\sfProbSets{} data sets, LightGBM on \sfCrpsLgbBestIxs{} and the world model on
\sfCrpsWmBestIxs{}, with \sfCrpsTieIxs{} level at three decimals; at the total the
average is lowest on \sfCrpsAvgBestTot{}, LightGBM on \sfCrpsLgbBestTot{} and the
world model on \sfCrpsWmBestTot{}, with \sfCrpsTieTot{} level.

\paragraph{The chance of selling nothing} The ranges do not forecast whether an
item sells at all. Built from errors scaled to each series, they put the chance
of a zero too low at every level: in each group of forecast chance, more
node-periods sold nothing than the chance says
(Figure~\ref{fig:sup-probabilistic}d). Their Brier score is
\sfBrierAvg{} for the average, \sfBrierWm{} for the world model and
\sfBrierLgb{} for LightGBM, against \sfBrierClim{} for each series' own share of
zero-sale periods over the year before; the models beat that share on
\sfBrierBeatsClimNames{} only (Table~\ref{tab:sup-probabilistic}). A retailer
who needs the chance of no sale, to check availability or prune the range,
should forecast it directly: the hurdle likelihood kept on M5 and Favorita
(Section~\ref{sec:sup-explore}) does so inside the world model, but these ranges
use only its point forecast.

\begin{table*}[t]
\centering\footnotesize\setlength{\tabcolsep}{4pt}
\caption{Probabilistic scores of the forecast ranges, mean over windows. Scaled CRPS item by store and at the
total, from each model's point forecast and the quantiles of its scaled errors in earlier windows on a grid of 23
quantiles, weighted by revenue as WRMSSE. Brier score of the forecast chance of selling nothing, item by store,
against each series' share of zero-sale periods over the year before (Past). Lowest in bold.}
\label{tab:sup-probabilistic}
\begin{adjustbox}{max width=\textwidth}\begin{tabular}{l ccc ccc cccc c}
\toprule
 & \multicolumn{3}{c}{CRPS, item by store} & \multicolumn{3}{c}{CRPS, total} & \multicolumn{4}{c}{Brier score, zero sales} & Periods\\
\cmidrule(lr){2-4}\cmidrule(lr){5-7}\cmidrule(lr){8-11}
Data set & WM & LGB & Avg & WM & LGB & Avg & WM & LGB & Avg & Past & with none\\
\midrule
Analgesics & 0.572 & 0.558 & \textbf{0.557} & 0.622 & 0.598 & \textbf{0.591} & 0.395 & 0.388 & 0.393 & \textbf{0.199} & 52\% \\
Bottled juice & 0.537 & 0.548 & \textbf{0.522} & 0.397 & 0.354 & \textbf{0.335} & 0.129 & 0.132 & 0.131 & \textbf{0.124} & 17\% \\
Cheese & 0.558 & 0.517 & \textbf{0.515} & 0.270 & 0.340 & \textbf{0.264} & 0.110 & 0.109 & 0.110 & \textbf{0.090} & 13\% \\
Frozen entrees & 0.388 & 0.365 & \textbf{0.365} & 0.249 & \textbf{0.208} & 0.209 & 0.205 & 0.204 & 0.205 & \textbf{0.165} & 30\% \\
Soft drinks & 0.560 & 0.517 & \textbf{0.511} & 0.630 & 0.626 & \textbf{0.564} & 0.185 & 0.192 & 0.188 & \textbf{0.151} & 30\% \\
Snacks & 0.525 & 0.507 & \textbf{0.501} & 1.430 & \textbf{0.898} & 1.047 & 0.194 & 0.189 & 0.192 & \textbf{0.148} & 26\% \\
Frat & 0.469 & 0.471 & \textbf{0.457} & \textbf{0.248} & 0.344 & 0.275 & 0.023 & 0.024 & \textbf{0.023} & 0.051 & 8\% \\
M5 & 0.733 & 0.753 & \textbf{0.730} & 0.427 & \textbf{0.336} & 0.360 & 0.334 & 0.326 & 0.331 & \textbf{0.195} & 56\% \\
Favorita & 0.532 & 0.555 & \textbf{0.531} & 0.163 & 0.160 & \textbf{0.128} & 0.137 & 0.143 & 0.140 & \textbf{0.117} & 20\% \\
\bottomrule
\end{tabular}\end{adjustbox}
\end{table*}

The reliability of a single run is a separate question: the world model's
scores vary from seed to seed, by a standard deviation of \loopSeedSdOne\% of the score on
the Dominick's categories for a single fit and \loopSeedSdFive\% for the five-copy average
the tuned model uses (Figure~\ref{fig:loopnoise}). Every comparison in this paper
uses the five-copy average, except on M5, where each model is a single fit, and
three seeds where a change is tried.

\begin{table}[t]
\centering\footnotesize\setlength{\tabcolsep}{3pt}
\caption{Coverage (\%) of each model's 80\% interval, from its own errors in earlier windows, item by store and at
the total, mean over the scored windows.}
\label{tab:sup-intervals}
\begin{adjustbox}{max width=\columnwidth}\begin{tabular}{lcccccc}
\toprule
 & \multicolumn{3}{c}{Item by store} & \multicolumn{3}{c}{Total}\\
\cmidrule(lr){2-4}\cmidrule(lr){5-7}
Data set & WM & LGB & Avg & WM & LGB & Avg\\
\midrule
Analgesics & 91 & 91 & 91 & 82 & 82 & 82 \\
Bottled juice & 72 & 73 & 73 & 85 & 98 & 92 \\
Cheese & 74 & 74 & 75 & 75 & 95 & 92 \\
Frozen entrees & 70 & 71 & 71 & 88 & 82 & 90 \\
Soft drinks & 75 & 75 & 76 & 68 & 68 & 68 \\
Snacks & 73 & 73 & 74 & 48 & 75 & 65 \\
Frat & 88 & 87 & 88 & 92 & 62 & 75 \\
M5 & 94 & 94 & 94 & 73 & 80 & 73 \\
Favorita & 81 & 80 & 80 & 65 & 85 & 81 \\
\bottomrule
\end{tabular}\end{adjustbox}
\end{table}

\secbarrier
\section{\ifjournal Reconciliation\else Decision costs and reconciliation\fi}
\label{sec:sup-decisions}

\ifjournal\else
% Decision costs: among the supporting analyses for arXiv, in the Results for the journal (IJPE leads with decisions).
\ifjournal\textbf{Simulated ordering loss and category-plan error are lowest with the average
on most data sets.} Accuracy and inventory cost can rank forecasts differently~\cite{kourentzes2020}, so each
model is also scored by the decisions made from it. \else\paragraph{Decision costs}\fi
The average gives the lowest category-plan cost on \supCatPlanAvgLowest{} of
\supCatPlanSets{} data sets and LightGBM on the other \supCatPlanLgbLowest{}
(frozen entrees and snacks); the world model alone is lowest on none
(Table~\ref{tab:sup-decisions}). Ordering each forecast's 0.8 quantile for every
item and store, with each unit over costing 1 and each unit short 4
\ifjournal (Section~\ref{sec:sup-method})\else (Section~\ref{sec:setup})\fi, gives a lower simulated loss with the average than with LightGBM on
\supOrderAvgBelowLgb{} data sets and less than with the world model on
\supOrderAvgBelowWm{}. Both are simulations under stated costs, not business
results: the ordering loss is for a single period, with no stock carried over, no
lead time and no shelf life, and the category-plan cost is a forecast error in
dollars, not a realised commercial outcome. They rank the forecasts by a stated
cost; they do not establish inventory savings in a deployed retailer, which would
need a multi-period inventory system with the retailer's own stock records.

\begin{table}[t]
\centering\footnotesize\setlength{\tabcolsep}{3pt}
\caption{Decision costs of the tuned models. Store orders: cost of ordering each forecast's 0.8 quantile, summed over
windows, relative to LightGBM's. Category plan: absolute error in dollars over sales in dollars at category by store
(\%), mean over windows; lowest shaded.}
\label{tab:sup-decisions}
\begin{adjustbox}{max width=\columnwidth}\begin{tabular}{lcccccc}
\toprule
 & \multicolumn{3}{c}{Store orders, ratio 0.8} & \multicolumn{3}{c}{Category plan (\%)}\\
\cmidrule(lr){2-4}\cmidrule(lr){5-7}
Data set & WM & LGB & Avg & WM & LGB & Avg\\
\midrule
Analgesics & 1.012 & 1.000 & 0.996 & 10.7 & 11.0 & \cellcolor{avgcell}\textbf{10.6} \\
Bottled juice & 0.983 & 1.000 & 0.965 & 13.9 & 14.0 & \cellcolor{avgcell}\textbf{13.2} \\
Cheese & 1.077 & 1.000 & 1.006 & 10.5 & 12.1 & \cellcolor{avgcell}\textbf{10.2} \\
Frozen entrees & 1.022 & 1.000 & 0.985 & 20.8 & \cellcolor{lgbcell}\textbf{18.6} & 18.8 \\
Soft drinks & 1.070 & 1.000 & 1.004 & 20.7 & 21.7 & \cellcolor{avgcell}\textbf{19.5} \\
Snacks & 1.040 & 1.000 & 1.000 & 26.5 & \cellcolor{lgbcell}\textbf{19.7} & 21.7 \\
Frat & 1.015 & 1.000 & 0.981 & 15.3 & 15.0 & \cellcolor{avgcell}\textbf{14.5} \\
M5 & 0.972 & 1.000 & 0.977 & 10.1 & 10.0 & \cellcolor{avgcell}\textbf{9.4} \\
Favorita & 0.952 & 1.000 & 0.960 & 9.8 & 9.7 & \cellcolor{avgcell}\textbf{8.7} \\
\bottomrule
\end{tabular}\end{adjustbox}
\end{table}

\fi

\paragraph{Reconciliation}
Summing the average upward is more accurate over all levels than the best
reconciled ETS forecast on \supAvgBeatsEts{} of \supReconSets{} data sets
(Table~\ref{tab:sup-reconcile}). The exceptions are soft drinks and snacks,
where reconciled ETS is more accurate at the coarse levels, which carry most of
the all-level score; combining either tuned model with ETS above it does not do
better than the average on most data sets. The comparison is with
reconciled ETS only, which does not read the price and promotion plan; it shows
that summing the average upward is a reasonable default here, not that it beats
reconciling the tuned models themselves, which was not tested.

\begin{table}[t]
\centering\footnotesize\setlength{\tabcolsep}{3pt}
\caption{Reconciliation against summing one forecast: WRMSSE averaged over all levels, mean over windows, of the best
reconciled ETS forecast, of each tuned model combined with ETS above it by weighted least squares, and of the average
summed upward.}
\label{tab:sup-reconcile}
\begin{adjustbox}{max width=\columnwidth}\begin{tabular}{lcccc}
\toprule
Data set & Best ETS & WM + ETS & LGB + ETS & Avg\\
\midrule
Analgesics & 0.836 & 0.903 & 0.933 & \cellcolor{avgcell}\textbf{0.687} \\
Bottled juice & 0.878 & 0.849 & 0.853 & \cellcolor{avgcell}\textbf{0.551} \\
Cheese & 0.742 & 1.430 & 1.215 & \cellcolor{avgcell}\textbf{0.477} \\
Frozen entrees & 0.549 & 0.510 & 0.474 & \cellcolor{avgcell}\textbf{0.332} \\
Soft drinks & \textbf{0.617} & 1.061 & 0.907 & 0.650 \\
Snacks & \textbf{0.698} & 0.979 & 0.805 & 0.790 \\
Frat & 0.693 & 0.499 & \textbf{0.466} & 0.469 \\
M5 & 0.945 & 0.821 & 0.813 & \cellcolor{avgcell}\textbf{0.658} \\
Favorita & 0.570 & 0.434 & 0.462 & \cellcolor{avgcell}\textbf{0.403} \\
\bottomrule
\end{tabular}\end{adjustbox}
\end{table}

% =====================================================================
\section{What was tried and dropped}
\label{sec:dropped}

Most of what was tried did not stay. The searches of Section~\ref{sec:loop} kept
a small share of the changes they ran (Table~\ref{tab:loop}); every change tried
alone in the first round is in Figure~\ref{fig:loopheat}, and the losses tried in
the \firstver{} are in Table~\ref{tab:losses}. Table~\ref{tab:dropped} lists
the larger things the \revn{} tried beyond the searches and did not keep, and
when each becomes worth trying again.

\begin{table*}[t]
\centering
\caption{\textbf{What the \revn{} tried beyond the searches and did not keep.}
Each was written down before it ran.}
\label{tab:dropped}
\footnotesize
\renewcommand{\arraystretch}{1.15}
\begin{tabular}{p{0.25\textwidth}p{0.36\textwidth}p{0.31\textwidth}}
\toprule
\textbf{Tried} & \textbf{What happened} & \textbf{Worth revisiting when} \\
\midrule
Count likelihoods that treat zero weeks on their own (negative binomial, zero-inflated, hurdle), on the world model & Kept on M5 and Favorita (hurdle) and snacks (negative binomial); nothing on the other six data sets & Lines are slower or more often out of stock than here \\
Pretraining the world model's encoder on the latent objective, then fitting the forecast at a lower rate & Kept on snacks only, on the search windows; with nothing kept on eight of nine data sets, Favorita's rerun on its final loss was stopped early & More unlabelled history is available than the forecast needs, as across many categories \\
Training on other data sets with each own data set (RQ4) & Helped neither model by more than \rqPoolGainMax{} on any data set, and hurt LightGBM most & Many more retailers' data in the same categories \\
The world model alone for series with less than 26 weeks of history & No better than the average on lines with history; worse on snacks & A longer history threshold is set from a retailer's own range data \\
Weights fitted to earlier windows in place of equal weights & Reported in Section~\ref{sec:sup-ablation} & Many more windows are available to fit them on \\
Timestral in place of TimesFM in the three-model ensemble & Reported in Section~\ref{sec:sup-ablation} & A retail-trained version is released \\
\bottomrule
\end{tabular}
\end{table*}

The \fvpaper{} also tried a stock term for the dip after a
promotion, extra loss weight on the periods after a promotion, and a graph model
sharing information between stores; none was kept, and they were not run again on
the tuned models. \ifjournal The\else Their record is in the \firstver{}; the\fi{} loss experiments among them are in
Figure~\ref{fig:sup-devlosses}.

\begin{figure*}[t]
\centering
\includegraphics[width=\textwidth]{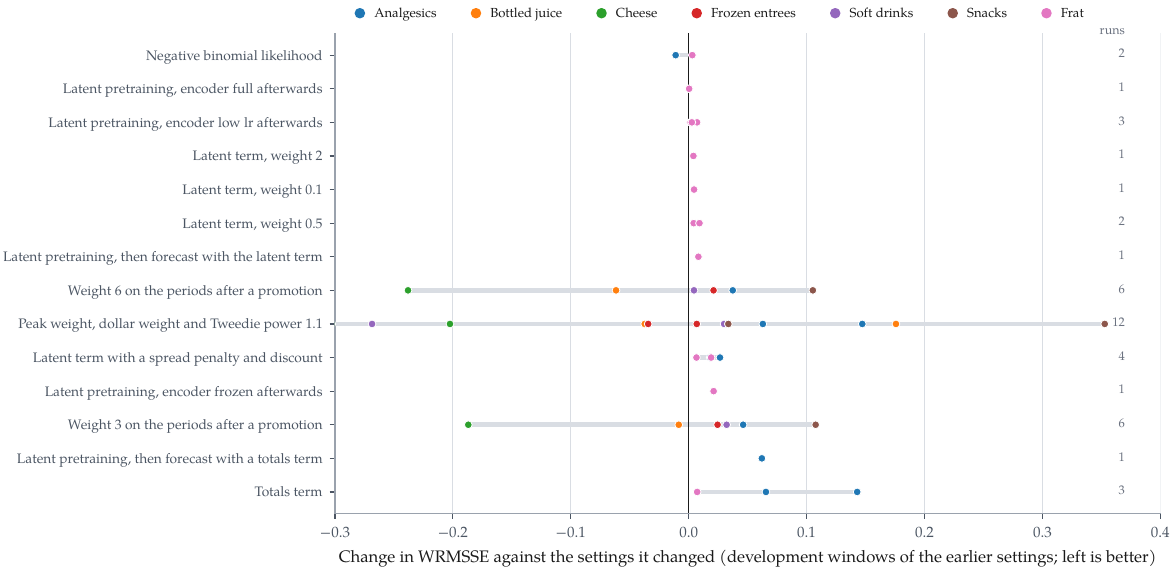}
\caption{Every loss change of the \firstver{} against the settings it changed,
on the development windows. Each dot is one run, coloured by data set; the bar
spans the runs of a change, and the right-hand column counts them. Changes not
run on the full data are not shown.}
\label{fig:sup-devlosses}
\end{figure*}

\subsection{Losses and pretraining explored}
\label{sec:sup-explore}

\textbf{Count likelihoods helped on three data sets and JEPA pretraining on one;
the plain loss stayed everywhere else.} After each search, the world model had a
round of count likelihoods and a round of JEPA pretraining
(Section~\ref{sec:loop}). The likelihood round screened \sfLossTried{}
likelihoods on each of \sfLossSets{} data sets with one seed each, \sfLossScreens{}
screens in all, of which \sfLossScreenBetter{} came in below the starting
settings' three-seed mean; a screen went on to three seeds when it beat the
starting settings on the same first seed, and a change was kept only where it beat the
start by more than the keep bar (Figure~\ref{fig:sup-explore}a and
Table~\ref{tab:sup-explore}). It was kept on \sfLossKeptSets{}: \sfLossKeptNames{}.
The JEPA round pretrained the encoder on the latent term alone
(Section~\ref{sec:losses}), for a short and a long run, then fitted the forecast
at a lower rate. Of its \sfJepaScreens{}
screens, \sfJepaScreenBetter{} came in below the starting three-seed mean (three
beat the first seed and went on to three seeds), and it was kept on
\sfJepaKeptNames{} only (Figure~\ref{fig:sup-explore}b).

The \firstver{} tried \sfDevChanges{} changes to the world model's loss in
\sfDevRuns{} runs on the development windows, \sfDevScored{} of them scored
against the settings they changed (Figure~\ref{fig:sup-devlosses}); \sfDevNotRun{}
were not run on the full data and \sfDevNoRef{} had no run of the unchanged
settings to compare with. \sfDevBetter{} of the
\sfDevScored{} lowered the error. Latent pretraining and the latent term, in
every form tried, moved the score by at most \sfDevLatentMax{} either way. Extra
weight on the periods after a promotion and the combined peak, dollar and
Tweedie weighting moved it most, in both directions: every form helped cheese
and hurt snacks and analgesics, so none was kept for every category.
The settings searches of Section~\ref{sec:loop} then tried the loss weights again
one at a time on each data set (Figure~\ref{fig:loopheat}).

\begin{table*}[t]
\centering\footnotesize\setlength{\tabcolsep}{4pt}
\caption{Every likelihood and JEPA pretraining recipe tried on the tuned world model, WRMSSE on the search windows.
Start: the settings each round began from.
A number in brackets is a one-seed screen that did not earn a three-seed try; a plain number is the three-seed mean;
bold marks the change kept, which had to beat the start by more than the keep bar on the search windows.}
\label{tab:sup-explore}
\begin{adjustbox}{max width=\textwidth}\begin{tabular}{l ccccc ccc}
\toprule
 & \multicolumn{5}{c}{Likelihood round} & \multicolumn{3}{c}{JEPA pretraining round}\\
\cmidrule(lr){2-6}\cmidrule(lr){7-9}
Data set & Start & NB & NB, spread by item & ZINB, spread by item & Hurdle NB, spread by item & Start & Short & Long\\
\midrule
Analgesics & 0.509 & (0.518) & (0.515) & (0.518) & (0.539) & 0.509 & 0.511 & (0.537) \\
Bottled juice & 0.587 & (0.592) & (0.601) & (0.614) & (0.604) & 0.587 & (0.635) & (0.615) \\
Cheese & 0.392 & (0.450) & 0.403 & 0.396 & 0.382 & 0.392 & (0.401) & (0.406) \\
Frozen entrees & 0.216 & 0.214 & 0.218 & 0.215 & 0.216 & 0.216 & (0.267) & (0.255) \\
Soft drinks & 0.572 & (0.613) & (0.604) & (0.606) & (0.598) & 0.572 & 0.612 & (0.603) \\
Snacks & 0.439 & (0.481) & \textbf{0.420} & 0.427 & 0.429 & 0.420 & (0.428) & \textbf{0.400} \\
Frat & 0.547 & (0.584) & (0.600) & (0.585) & (0.590) & 0.547 & (0.572) & (0.566) \\
M5 & 0.547 & (0.553) & (0.558) & 0.552 & \textbf{0.533} & 0.533 & (0.537) & (0.544) \\
Favorita & 0.483 & 0.488 & (0.476) & 0.469 & \textbf{0.468} & 0.483 & (0.527) & (0.523) \\
\bottomrule
\end{tabular}\end{adjustbox}
\end{table*}

\begin{figure*}[t]
\centering
\includegraphics[width=\textwidth]{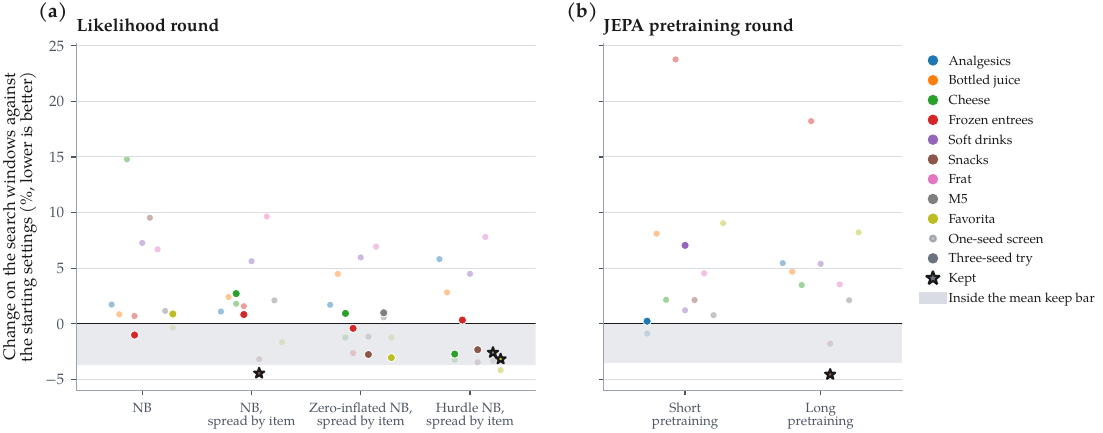}
\caption{The likelihood and JEPA pretraining rounds of the tuned world model.
Change in WRMSSE on the search windows against the settings each round started
from, for each one-seed screen (faint) and three-seed try (solid); stars mark the
change kept. A change was kept only where its three-seed mean beat the start by
more than the keep bar; the grey band is the mean keep bar over data sets.}
\label{fig:sup-explore}
\end{figure*}

% the constraints appendix as one full-width block at the top of a page: its table stays under its heading
\twocolumn[\ifsupplement\else\onecolumn\fi
\section{Constraints the models work within}\seclabel{sec:constraints}

Table~\ref{tab:constraints} lists the constraints. Each model's tuned settings are
in Table~\ref{tab:final-settings}.

\label{sec:supplement}

\begin{center}
\captionof{table}{\textbf{Every constraint the trained models and the ensemble work
within, with its value and where it lives in the code.}}
\label{tab:constraints}
\footnotesize
\renewcommand{\arraystretch}{1.15}
\begin{tabular}{p{0.29\textwidth}p{0.29\textwidth}p{0.34\textwidth}}
\toprule
\textbf{Constraint} & \textbf{Value} & \textbf{Where} \\
\midrule
Log forecast bound (world model) & $[-10, 10]$ around the item's level & \cpath{src/rwm/model/state_model.py}{model code} readout \\
Lift when nothing is planned & Exactly zero & Readout; tested in the code's tests \\
Baseline under a changed plan & Unchanged & Readout; tested in the code's tests \\
Regular price & Highest price in the last 12 periods & Plan features \\
Brand-switching share $\alpha$ & One per run, fitted on the last 8 training weeks & Dominick's world model \\
Forecasts & Not below zero & Every model \\
Average weights & One half each & \cpath{scripts/revision_analysis.py}{analysis code} (\texttt{ensemble\_2}) \\
Three-model ensemble weights & One third each, fixed before the results & \cpath{scripts/revision_analysis.py}{analysis code} \\
Forecast by history & World model alone under 26 weeks (182 days) of sales & \cpath{scripts/revision_analysis.py}{analysis code} (\texttt{by\_history}) \\
Seed copies & Five fits averaged (one on M5) & \cpath{src/rwm/seed_average.py}{model code} \\
Interval pooling & Horizons pooled below 50 earlier errors per horizon & \cpath{src/rwm/evaluation/compare.py}{evaluation code} \\
Interval ends & Rounded outwards to whole units (not Favorita) & \cpath{src/rwm/evaluation/compare.py}{evaluation code} \\
TimesFM covariate fit & Groups of 2,048 series & \cpath{src/rwm/prior_work/}{comparison-model code} \\
Timestral context and batch & 512 periods; batch 256, halved on running out of GPU memory & \cpath{src/rwm/prior_work/timestral_zero_shot.py}{comparison-model code} \\
\bottomrule
\end{tabular}
\end{center}
\ifsupplement\else\twocolumn\fi
\vspace{8pt}]
% Reproducibility and compute: in the main text for arXiv, in the online supplement for the journal.
\section{Reproducibility and compute}
\label{sec:repro}

\paragraph{Records}
All code, configuration files and tests are in the repository. A run refuses to
start on uncommitted code and writes a manifest holding the code commit, the
full configuration and its checksum, the seed, checksums of every data file
read, package versions, the machine and run times. The scores and manifests of
the final, transfer and pooled runs, every try of the searches, and the
analyses of the tuned models are committed under \nolinkurl{results/final/},
\nolinkurl{results/ftransfer/}, \nolinkurl{results/pooled/},
\nolinkurl{results/research/} and \nolinkurl{results/analysis_tuned/}; the plan
and every dated entry are in \cpath{docs/PAPER1_REVISION_PLAN.md}{the plan file}.
\texttt{make} in \texttt{paper/} rebuilds every figure, table and quoted number
from these files (\buildSeconds{} seconds from a fresh clone of the repository). Forecast files and model checkpoints are kept outside the
repository on the results volume, with their checksums recorded.

\paragraph{Compute}
The \firstver{} ran on one Google Colab machine with an NVIDIA \computeGpu{}
(Table~\ref{tab:compute}: \computeRuns{} timed runs, \computeHours{}
machine-hours). \ifjournal The tuning, final, transfer and pooled runs\else The \revn{}\fi{} ran on rented machines: the searches, final runs,
transfer tests and pooled training on GPUs from Vast, managed and analysed on
Modal. Table~\ref{tab:compute-revision} gives the run time recorded in each of
the \computeRevRuns{} run folders, \computeRevHours{} hours in all, most on RTX
4090 GPUs; because a resumed run records only its last stretch, this is a floor.
The machine time paid for on Vast was \computeVastGpuHours{} GPU-hours
(US\$\computeVastGpuUsd{}), and \ifjournal those runs\else the \revn{}\fi{} cost US\$\computeTotalUsd{} in all
up to the last billing statement: US\$\computeVastUsd{} on Vast, including
storage and data transfer, and US\$\computeModalUsd{} on Modal. The total
includes every run made, the searches' screens, repeats after machine failures
and runs lost when a machine stopped, not only the runs reported.

\begin{table}[t]
\centering
\caption{\textbf{Run times of the final and comparison runs, from their manifests.} Wall-clock minutes per
run (data loading, training, forecasting and scoring) and hours in total, on one Google Colab machine with an NVIDIA Tesla T4 (LightGBM uses its CPU). Dominick's:
six categories together; a Dominick's world-model run averages three fits. Two M5 runs resumed after the
Colab runtime dropped record only the resumed part (the shortest values in their rows). The lift model and the graph model are models of the earlier version of this study, not compared in this paper. Not counted: runs saved before
run times were recorded, the development study, and TimesFM on Dominick's and Frat.}
\label{tab:compute}
\footnotesize
\setlength{\tabcolsep}{3.5pt}
\begin{tabular}{llrrlr}
\toprule
\textbf{Data} & \textbf{Model} & \textbf{Runs} & \textbf{Median} & \textbf{Range} & \textbf{Hours} \\
\midrule
M5 & World model & 9 & 101.5 & 37.3 to 103.2 & 14.0 \\
M5 & Lift model & 9 & 88.7 & 10.3 to 94.6 & 12.2 \\
M5 & LightGBM & 9 & 78.1 & 76.5 to 79.9 & 11.7 \\
M5 & Graph model & 12 & 7.2 & 7.2 to 7.5 & 1.5 \\
M5 & TimesFM & 8 & 3.9 & 3.8 to 4.1 & 0.5 \\
M5 & Seasonal naive & 2 & 1.3 & 1.2 to 1.3 & 0.0 \\
Dominick's & World model & 18 & 9.2 & 9.0 to 10.6 & 2.8 \\
Dominick's & LightGBM & 18 & 4.6 & 3.1 to 11.5 & 1.7 \\
Frat & World model & 3 & 2.2 & 2.2 to 2.2 & 0.1 \\
Frat & Lift model & 3 & 2.0 & 1.9 to 2.0 & 0.1 \\
Frat & LightGBM & 3 & 1.5 & 1.5 to 1.6 & 0.1 \\
\midrule
\multicolumn{5}{l}{All 94 timed runs} & 44.7 \\
\bottomrule
\end{tabular}
\end{table}

\begin{table}[t]
\centering
\caption{\textbf{Run time of the \revn{}, from each run's manifest.} Runs counts every run folder, the pieces of
split runs included; hours is their summed wall-clock time; machines are the GPUs or CPU hosts the runs used, with the
number of runs on each. A run resumed from a partway save records only its last stretch, so the hours are a floor;
the machine time paid for is in the text.}
\label{tab:compute-revision}
\footnotesize
\setlength{\tabcolsep}{3pt}
\begin{tabular}{p{0.36\columnwidth}rrp{0.32\columnwidth}}
\toprule
\textbf{Study} & \textbf{Runs} & \textbf{Hours} & \textbf{Machines} \\
\midrule
Searches of both models' settings & 3,958 & 4,305 & RTX 4090 (1182), RTX 5090 (54), L4 (868), CPU (1288), other GPUs (566) \\
Final runs on the rolling windows & 245 & 41 & RTX 4090 (220), RTX 5090 (16), other GPUs (9) \\
Transfer and new-retailer tests & 368 & 156 & RTX 4090 (252), RTX 5090 (85), CPU (10), other GPUs (21) \\
Training on other retailers' data & 249 & 176 & RTX 4090 (216), RTX 5090 (17), other GPUs (16) \\
Comparison models on cleaned prices & 60 & 8 & RTX 4090 (48), other GPUs (12) \\
Forecasts 52 weeks ahead and their settings searches & 2,709 & 320 & RTX 4090 (685), RTX 5090 (1138), L4 (2), CPU (109), other GPUs (775) \\
History-length test & 654 & 136 & RTX 4090 (630), RTX 5090 (1), other GPUs (23) \\
Held-out periods & 125 & 12 & RTX 4090 (106), RTX 5090 (11), CPU (4), other GPUs (4) \\
\midrule
All & 8,368 & 5,153 & \\
\bottomrule
\end{tabular}
\end{table}

\FloatBarrier

\else\section{Introduction}
\label{sec:intro}

Retail decisions run at different grains and on different clocks. Supplier
range, space and cost reviews look at an item's volume across the chain over 26
or 52 weeks, and the terms agreed in them commit allowances for the year ahead.
Promotions are planned item by week in blocks of six to twelve weeks, and the
promotional and retail allowances that fund them follow the same blocks.
Merchants judge their category week by week. Store inventory is ordered item by
store, day by day \cite{fildes2022}. Each of these decisions reads the same
demand at a different grain and over a different span of time.

Since there is no single grain, there is no reason to expect one model to be
the most accurate at every grain and every horizon. Retailers do not want a
separate model at each grain either. Separate models disagree, and value leaks
at each hand-off from a coarse long-range plan to a finer short-range one. The
annual plan agreed with a supplier is broken into promotion blocks, weeks and
store orders. Where the finer forecasts do not add back to the plan, the gap is
paid for in stock, markdowns or lost sales. The forecasting literature treats
this as coherence across the levels of the product and store hierarchy
\cite{hyndman2011,mint} and across levels of time \cite{athanasopoulos2017},
usually by forecasting each level and reconciling the forecasts. This paper
takes the other route retailers use: one forecast made at the finest grain,
item by store by period, and summed to every coarser grain, so that the grains
add up by construction. The question this paper answers is whether one
forecast, built once at the finest grain, can serve the decisions a retailer
makes from it, across every kind of line, and which kind of model, or
combination of models, does that best.

Lines differ as much as grains do. In grocery a short head of fast lines
carries most of the units, and a long tail of slow lines sells a few units a
week or none. On the Dominick's categories studied here, \rqFastShareDomLo{}
to \rqFastShareDomHi{}\% of the items make 80\% of unit sales. Ranges also
change: items are added and delisted and stores open, so part of every forecast
is for a line or a store with little or no history. A forecast that serves the
fast lines well can fail on the tail or on a new item, and a planner deciding
for one kind of line needs to know which. Lines also differ in how they are
priced: some sell at a steady price all year, others are cut often, and a
chain's pricing, every day low prices or regular deals, sets the mix.

We test four kinds of model. Gradient-boosted trees are the proven method:
LightGBM was behind the leading entries of the M5 competition
\cite{m5accuracy}. Pretrained foundation models (TimesFM, Chronos-Bolt and
Timestral) forecast without training on the retailer's data. The Temporal
Fusion Transformer is an attention-based model trained on it. World models, the
newest of the four, learn a state for each item and carry it forward one period
at a time under the price and promotion plan; ours is shown in
Figure~\ref{fig:overview}. The world model and LightGBM were each tuned by the
common search procedure on every data set, and the average of forecasts is tested
alongside them.\ifjournal{} Because retailers order stock and plan categories from the
forecast, each model is also scored by a simulated cost of those decisions: a
single-period store order set at a quantile of the forecast, and the dollar error
of the category plan
(Section~\ref{sec:rq-decisions}).\fi{} The data are nine data sets from four retailers' public data: M5, six categories
of Dominick's, Breakfast at the Frat, and Corporaci\'on Favorita. Six of the nine
are categories of one retailer, so the evidence across retailers rests on four.

The study covers the short end of the planning cycle. Forecasts run 8 weeks
ahead on the weekly data and 28 days ahead on the daily data, and are scored at
every grain from item by store to the chain total, period by period and summed
over the whole window as a promotion block. A test 52 weeks ahead, on the six
Dominick's categories and M5, was added in \thisrev{}
(Section~\ref{sec:long-horizon}); judging year-ahead forecasts against a
retailer's own revised plans and range decisions needs data no public set
carries, and is left for work with an industry partner. Of the four Ps of retail, the study covers price and
promotion, through the plan inputs these data record; product and place are out
of scope. Every analysis was written down before it ran; the few added after
the first results were read are listed in Section~\ref{sec:limits}.

Section~\ref{sec:framework} sets out the framework, and the results follow it
part by part: grain, the nature of the line, seen and unseen data, and
transferability, with the broad view last.

\begin{figure*}[t]
\centering
\includegraphics[width=\textwidth]{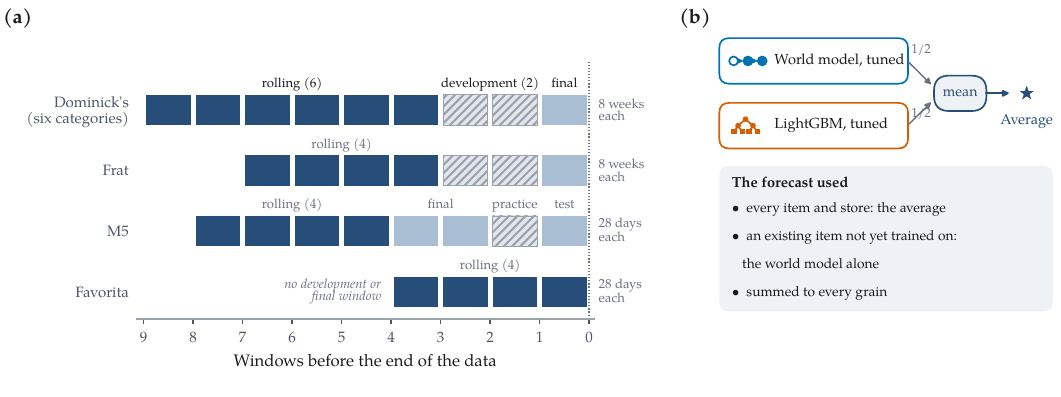}
\caption{\textbf{Every result of Section~\ref{sec:results} is scored on rolling
windows that no choice of setting saw, and the forecast recommended is the average of the two tuned
models.} \textbf{(a)} Windows by data set, in windows before the end of the data:
the rolling windows scored in this paper; and the development and final windows
\ifjournal used to develop the models\else of the \firstver{}\fi{}, which are not scored here. The searches that tuned each
model scored windows earlier than the rolling windows. \textbf{(b)} The forecast
that the results support: the equal-weight average of the tuned world model and
the tuned LightGBM for every item and store, the world model alone for an item
the models have not yet been trained on, summed to every grain.}
\label{fig:study}
\end{figure*}

\begin{figure*}[t]
\centering
\includegraphics[width=\textwidth]{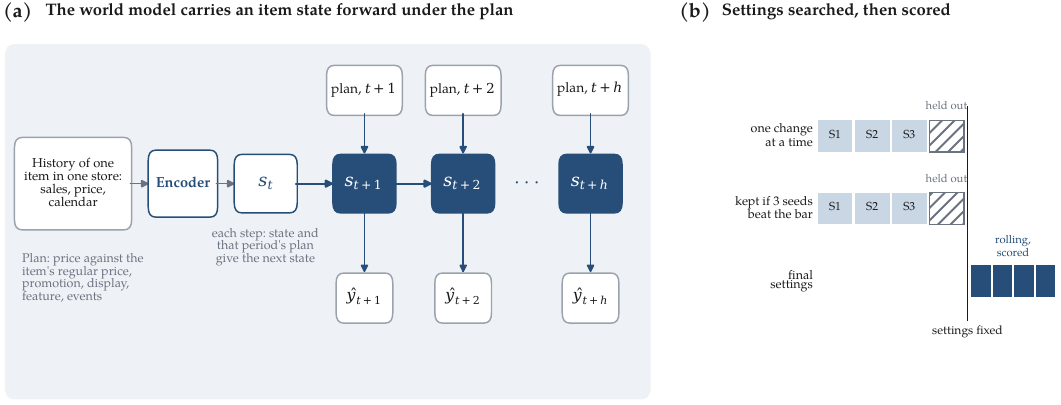}
\caption{\textbf{The world model carries each item's state forward under the
plan.} \textbf{(a)} An encoder turns the recent history of one item in one store into
a state $s_t$. A transition step takes the state and one period's plan and
gives the state one period later; each period's sales are read from the state
as it stands after the plan so far. \textbf{(b)} Each model's settings were
searched one change at a time on early windows (S1 to S3), a change kept only if
its mean over three seeds beat the best so far by the keep bar, with the latest
search window held out of every decision; the final settings were then scored
once on the later rolling windows (Section~\ref{sec:loop}).}
\label{fig:overview}
\end{figure*}

Section~\ref{sec:related} places the work among retail promotion models,
hierarchical evaluation, forecast combination, pretrained forecasters and world
models. Sections~\ref{sec:data} to \ref{sec:setup} give the data, the models and
the experimental setup. Section~\ref{sec:results} reports the results,
Section~\ref{sec:dropped} what was tried and dropped\ifjournal{} (in the online
supplement, with the ensemble ablation, the plan-input occlusion, the calibration
of the forecast ranges, every loss and pretraining recipe tried and the detail of
the searches)\fi, and
Sections~\ref{sec:discussion} to \ref{sec:conclusion} discuss what the results
mean for a retailer, their limits, and conclude.

% =====================================================================
\section{Framework}
\label{sec:framework}

The overall question of this paper is:

\begin{quote}
Can one forecast, built once at the finest grain of item by store by period and
summed to every coarser grain, serve the decisions a retailer makes from it
across every kind of line, and which kind of model, or combination of models,
does that best?
\end{quote}

\noindent It is answered through four research questions, in the order a
retailer meets them:

\begin{description}\setlength{\itemsep}{2pt}
\item[RQ1, grain.] At which grains of product, store and time does each model
  forecast best, period by period and over a whole promotion block, when every
  grain is summed from the same item-by-store forecast?
\item[RQ2, nature of the line.] How does each model do on the fast lines that
  carry most of the sales, on the long tail of slow lines, on lines with
  short history, and (added in \thisrev{}) on lines sold at a steady price
  against lines on frequent price cuts?
\item[RQ3, seen and unseen data.] How much accuracy does each model lose on
  items and stores it was not trained on?
\item[RQ4, transferability.] Do a model's settings carry to another retailer,
  and does training on other data sets (other categories of the same retailer and other retailers) make it more accurate?
\end{description}

\noindent These are the questions retailers ask of a forecast in practice\ifjournal\else{}
(Section~\ref{sec:background})\fi. Each question fixes what is measured before any
model is compared,
and the results are reported in the same order (Section~\ref{sec:results}).
The broad view then brings the four answers together.

\paragraph{RQ1, grain} Does the forecast hold up at every grain a decision is made
at, and do the grains add up? Every model forecasts each item in each store for
each period, and the forecasts are summed to item, store, category, category by
store and the chain total. Each grain is scored period by period, which is how
store orders and the category week are judged, and over the whole forecast
window as one block, which is how a promotion is judged. Because every grain is
a sum of the same item-by-store forecast, the plans built from them add up, and
any value lost between grains shows as a difference in accuracy between them.

\paragraph{RQ2, nature of the line} Does the forecast hold up on every kind of
line? Grocery sales are concentrated: a short head of fast lines carries most of
the units, and a long tail of slow lines sells a few units a week or none. Lines
also differ in how much history they have. The fast lines are defined as the
items that together make 80\% of unit sales over the year before the forecast,
the slow lines as the rest, and lines with short history as those whose first
sale falls less than 26 weeks before the forecast. Each part is scored on its
own series at every grain, so that the total of the fast lines, the slow lines
of each store, and so on, are each judged as a planner would judge them. The
share of items that makes 80\% of sales is reported for each data set.
Added in \thisrev{}, after the 52-week test: each item-by-store series is also
classed, from the year before each window, by how often it sells (in under half
the periods, or in nine of ten or more) and by how often its price is cut (never,
in under a quarter of the periods, or in a quarter or more; a cut is a price at
most 95\% of the line's most common price that year, and on Favorita, which
records no prices, the promotion flag), and the periods of each window are split
into price-cut periods and the rest. Each kind is scored on its own series, and
the world model is said to be ahead on a kind when it is clearly more accurate
than LightGBM on at least five of the nine data sets, a rule fixed before the
scoring ran.

\paragraph{RQ3, seen and unseen data} Does the forecast hold up on items and stores
the model was not trained on? Ranges change and stores open, so part of every
forecast is for lines or stores with little or no history in the model. A fifth
of the items, or of the stores, is kept out of training; once trained, each
model reads the kept-out series' recent sales and forecasts them. The cost of
not having trained on them is the accuracy lost on that fifth against a model
trained on everything.

\paragraph{RQ4, transferability} Do the settings carry to another retailer, and does
training on other data sets help? Settings tuned on M5 are carried
unchanged to Favorita and compared with settings tuned on Favorita itself. Each
model is also trained on its data set together with the other data sets, and
compared with the same model trained on its own data alone.

\paragraph{The broad view} The four parts are then brought together: which
model, or combination, gives the best balance across grains, lines and seen and
unseen data, and where each kind of model fits in the planning cycle. Findings
are reported for each grain and each kind of line, since that is what a
retailer needs to choose a model for a decision; the broad view simplifies them
and does not replace them.

% =====================================================================
\section{Related work}
\label{sec:related}

\subsection{Retail forecasting and the promotion plan}

Fildes, Ma and Kolassa~\cite{fildes2022} review retail forecasting research and
practice and note that most studies forecast and evaluate at one level of
aggregation, usually the item, although retail decisions are spread across
levels. Sales of a single item in a single store are intermittent, which led to
dedicated methods for intermittent demand~\cite{croston,syntetos2005} and to
evaluating forecasts as count distributions~\cite{kolassa2016}; Ma et al.~\cite{ma2025}
match forecasting methods to SKU demand patterns in a beverage manufacturer's data. Studies of intermittent
items score forecasting methods both by their error and by the stock they lead
to~\cite{babai2014,babai2019}, and Kourentzes, Trapero and Barrow~\cite{kourentzes2020}
propose choosing a forecasting model for its inventory cost rather than its accuracy. The Tweedie
family~\cite{jorgensen1987}, a compound Poisson--gamma distribution with a point
mass at zero, is the loss many M5 entries used, and the loss we use.

Promotions move retail sales more than anything else in the plan. PromoCast
forecast promoted-week sales for a chain from the planned event and the item's
past response~\cite{promocast}. Van Heerde, Leeflang and
Wittink~\cite{vanheerde2000} and Mac\'e and Neslin~\cite{mace2004} measured
the dips in sales before and after a promotion, when shoppers stock up and then
buy less. Ma, Fildes and Huang~\cite{ma2016} showed that promotional information
from within and across categories improves item-level forecasts. For perishable
supermarket items, van Donselaar and colleagues~\cite{vandonselaar2016} found that
promotion lifts come more from substitution within the category than from
stockpiling, and that forecasts improve when routine and non-routine categories
are modelled apart.
Gupta~\cite{gupta1988} split the promotion bump into brand switching, purchase
timing and quantity, and van Heerde, Gupta and Wittink~\cite{vanheerde2003}
showed that in units only about a third of the bump is switching from other
brands. Attraction models make an item's sales the category's sales times its
share~\cite{bell1975,cooper1988}; our brand-switching step follows that
structure. Dominick's Finer Foods ran pricing experiments across its
stores~\cite{hoch1995}, and its scanner data are behind much of this
work~\cite{dominicks}.

\subsection{Decisions across grains and lines}

Retail decisions read the same demand at different grains and over different
spans: replenishment item by store by day, promotions item by week, category
reviews by week, and range, space and supplier terms over a year or more
\cite{fildes2022}. Forecasts made separately for each grain need not agree, and
the literature restores agreement by reconciliation, across the product and store
hierarchy \cite{hyndman2011,mint}, across levels of time
\cite{athanasopoulos2017}, and across both together
\cite{kourentzes2019crosstemporal}. The alternative studied here, a single
forecast at the finest grain summed upward, is coherent by construction and is
the way the M5 competition scored its entries \cite{m5accuracy}; its cost is
that one model must serve every grain.

Lines also differ in how much they sell and how much history they have. A small
share of items carries most of the units, and the long tail sells a few units a
week or none, which led to dedicated methods for intermittent demand
\cite{croston,syntetos2005}. Lines with little or no history are forecast from
the histories of similar products: by matching a new product to the life-cycle
curves of earlier ones \cite{hu2019newproduct}, or from product attributes
\cite{vansteenbergen2020}. Global models trained across many series
\cite{montero2021} can forecast a series they were not trained on from its
recent sales alone, which is how the models in this study meet new items and
new stores.

\subsection{Hierarchies, combinations and evaluation}

Forecasts that must add up across a hierarchy are studied as hierarchical
forecasting, from bottom-up and top-down to optimal
reconciliation~\cite{athanasopoulos2009,hyndman2011,mint}. Our models forecast
item by store and their forecasts are summed, as the M5 accuracy competition
scored entries, with a weighted, scaled error averaged over twelve
levels~\cite{m5accuracy}; reconciliation is tested as a benchmark. The scaling
follows Hyndman and Koehler~\cite{hyndmankoehler}, and a companion competition
scored forecast distributions~\cite{m5uncertainty}. Tashman~\cite{tashman2000}
sets out why a comparison should use several forecast origins. Hansen, Lunde and
Nason's model confidence set~\cite{hansen2011} gives the set of models that
cannot be told apart from the best.

Combining forecasts from different models reduces error~\cite{batesgranger,clemen1989},
and an equal-weight average is hard to beat with estimated weights, the
forecast combination puzzle, which Smith and Wallis~\cite{smithwallis2009}
explain by the error in estimating the weights; Wang and
co-authors~\cite{wang2023} review the field. We fix an equal-weight ensemble in
advance and test weights fitted on earlier windows against it. Interval
forecasts by calibrating each model on its own past errors follow conformal
prediction~\cite{vovk2005}.

\subsection{Global models, trees and neural forecasters}

Global models, fitted across many series at once, are often more accurate than
one model per series~\cite{montero2021}. In M5 the leading entries were built on
gradient-boosted trees~\cite{lightgbm,januschowski2022}; the winner averaged
direct and recursive LightGBM models~\cite{inandjung}. Neural global models
include N-BEATS~\cite{nbeats}, DeepAR~\cite{deepar} and the Temporal Fusion
Transformer~\cite{tft}, the last two taking known future inputs such as price.
Graph neural networks share information between related
series~\cite{graphsage}; our item-to-item experiments use
attention~\cite{transformer} and conditional similarity
networks~\cite{veit2017}.

\subsection{Pretrained foundation models}

Pretrained time-series models forecast series they were not trained on.
TimesFM~\cite{timesfm} is a decoder-only transformer pretrained on a large
corpus of series; Chronos~\cite{chronos} tokenises values and trains a language
model on them, and Chronos-Bolt is its faster direct-quantile version;
Lag-Llama~\cite{lagllama}, Moirai~\cite{moirai} and TimeGPT~\cite{timegpt}
follow similar lines. TimesFM~2.5 adds a covariate method that fits a linear
model on future inputs and forecasts the remainder with the pretrained
model~\cite{timesfm25}. Timestral~1.0~\cite{timestral}, released by Mistral in
October 2026, reads known future inputs inside the model itself and returns
21 quantiles. We test whether any of them, given the plan, matches models
trained on it.

\subsection{World models and latent prediction}

A world model learns a state of its environment and how that state moves under
actions, so that the effect of a planned action can be rolled forward before it
is taken~\cite{worldmodels,planet,dreamerv3}. LeCun~\cite{lecun2022} proposed
joint-embedding predictive architectures (JEPA), which predict the
representation of the future rather than the future itself; I-JEPA~\cite{ijepa}
and V-JEPA~2~\cite{vjepa2} apply the idea to images and video, the latter with
an action-conditioned predictor used for planning. Such models guard against a
collapsed representation with a slowly updated target encoder~\cite{byol}, a
variance and covariance penalty~\cite{vicreg}, or the isotropic-Gaussian penalty
of LeJEPA~\cite{lejepa}, also used in LeWorldModel~\cite{leworldmodel}.
TD-MPC2~\cite{tdmpc2} rolls a latent state forward under actions; DWM~\cite{dwm}
keeps the effect of an action apart from the world's own dynamics, and
AD-WM~\cite{adwm} trains world models to tell actions apart for counterfactual
planning. Stable-worldmodel~\cite{stablewm} evaluates world models under
controlled changes, as our perturbation tests do. Closest to us,
WorldTS~\cite{worldts} learns covariate-conditioned latent dynamics for
forecasting with known future inputs. \ifjournal Earlier work, M-JEPA~\cite{mjepa},\else Our earlier work, M-JEPA~\cite{mjepa},\fi{}
compared JEPA fine-tuning recipes with leakage-resistant splits. In retail the
actions are the plan. Our world model uses a recurrent transition~\cite{gru}
with layer normalisation~\cite{layernorm} on a transformer
encoder~\cite{transformer}. It is not compared here with the same network
without the transition; the carried-state tests (Section~\ref{sec:sup-mechanism})
show what the state does, not how much accuracy it adds over a network without one.

\subsection{What this paper adds}

Two results here are not new: an equal-weight average of different models is
hard to beat \cite{smithwallis2009,wang2023}, and gradient-boosted trees are
strong on retail panels \cite{m5accuracy}. The finding here is that the average
of a tuned world model and tuned trees is robust across grains, but the choice of
model changes for items absent from training. The paper adds four things. It scores
one forecast at every grain a retailer decides at, period by period and over the
whole promotion block, so that the choice of model is read grain by grain rather
than from one score averaged over levels. It reads the same comparison by the
kind of line, separating the fast lines that carry most sales from the slow tail
and from lines with short history. It tests what each model loses on items and
stores it was not trained on, with LightGBM also run without the item label, and whether its settings and training data carry
across retailers. And it tunes the world model and gradient-boosted trees by a common search
procedure, each over its own settings, so that the comparison is between two
tuned models.

% =====================================================================
\section{Data}
\label{sec:data}

\begin{figure*}[t]
\centering
\includegraphics[width=\textwidth]{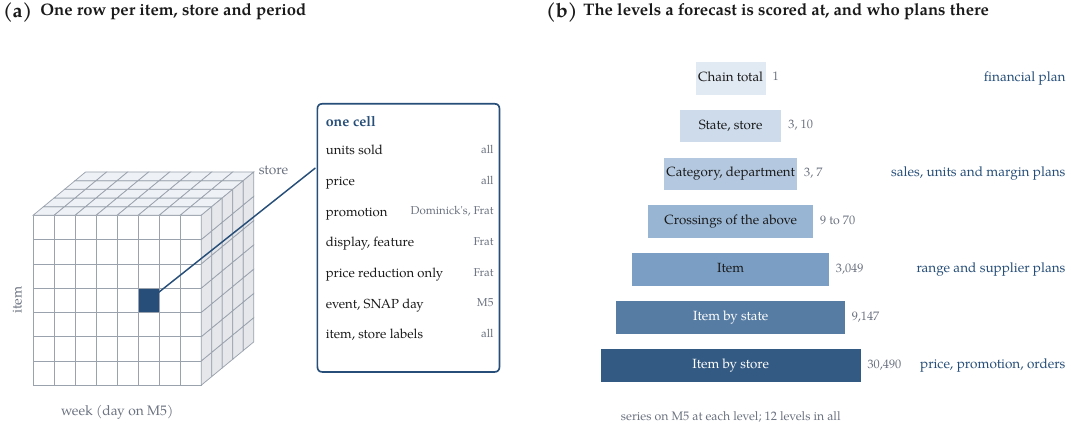}
\caption{\textbf{Every data set is one table of item, store and period, scored
at every level a retailer plans at.} \textbf{(a)} Each cell holds the units sold,
the price and whatever plan the data set records; a lever a data set does not
record is passed to the models as unknown, not as zero. \textbf{(b)} The M5
hierarchy with the number of series at each level, and the decisions usually
made at each. Dominick's and Frat have six levels: total, category, store,
category by store, item and item by store; Favorita has total, store, family,
store by family, item and item by store.}
\label{fig:cube}
\end{figure*}

All four data sets are public (Table~\ref{tab:data}) and share one table shape
(Figure~\ref{fig:cube}a). \emph{M5} holds daily Walmart sales of 3,049 items in 10
stores across three states from 2011 to 2016, with prices by week, event days
and the days on which SNAP food assistance is paid~\cite{m5accuracy}.
\emph{Dominick's Finer Foods} holds weekly scanner data by store for a Chicago
chain from 1989 to 1997, with price, a promotion code and cost~\cite{dominicks}.
We use six of its \domCatN{} categories (cheese, snack crackers, soft
drinks, bottled juice, analgesics, frozen entrees), chosen deliberately to test the world model
where it was weakest. A first run of all \domCatN{} with untuned settings, before any setting was
searched, put snacks, cheese, bottled juice and soft drinks \domCatRankSna{}, \domCatRankChe{},
\domCatRankBjc{} and \domCatRankSdr{} of \domCatN{} for how far the world model trailed LightGBM; these four
were taken with two categories where it was ahead (analgesics, on which the models were first set,
and frozen entrees). The six hold \domCatDollarShare\% of the chain's dollar sales across the
\domCatN{} categories and run from one of the least promoted (analgesics) to the most promoted
(soft drinks; Table~\ref{tab:domcats}). Every model sees cleaned
prices: a recorded price more than five times above, or below a fifth of, the
same item's median price that week across the stores that recorded one (at least
three) is treated as not recorded, and the store's last good price is used. \emph{Breakfast at the Frat}
holds weekly sales of 55 breakfast items in 77 stores from 2009 to 2012, with
the shelf price, base price, and whether the item was on display, in the weekly
circular (feature) or on a tag-only price reduction~\cite{frat}.
\emph{Corporaci\'on Favorita} holds daily sales of an Ecuadorian grocery chain
to 15 August 2017~\cite{favorita}. It was added in \thisrev{} as a data set
kept out of development: the series are the 10 stores with the most units sold
in 2016 crossed with the items sold in all ten that year, from 1 January 2015;
returns are set to zero; the plan is the promotion flag, national holidays and
events, and paydays (the 15th and the last day of each month). Favorita records
no prices, so its WRMSSE weights are unit weights, not revenue weights. No raw
data is redistributed.

\begin{table}[t]
\centering
\caption{\textbf{Four public data sets, all scored item by store and at every
level above it.} Rolling windows were added in \thisrev{} and precede every
window used for development.}
\label{tab:data}
\footnotesize
\renewcommand{\arraystretch}{1.2}
\setlength{\tabcolsep}{4pt}
\begin{tabular}{p{0.17\columnwidth}p{0.14\columnwidth}p{0.27\columnwidth}p{0.28\columnwidth}}
\toprule
\textbf{Data} & \textbf{Series} & \textbf{Plan inputs} & \textbf{Rolling windows} \\
\midrule
M5 (daily) & 30,490 & Price, event days, SNAP days & Four of 28 days, Nov 2015 to Feb 2016 \\
Dominick's (weekly) & Six categories & Price, promotion & Six of 8 weeks per category \\
Frat (weekly) & 3,909 & Price, promotion, display, feature, price reduction only & Four of 8 weeks \\
Favorita (daily) & Top 10 stores & Promotion, holidays and events, paydays & Four of 28 days, the last ending 15 Aug 2017 \\
\bottomrule
\end{tabular}
\end{table}

\paragraph{What each data set holds and what the runs use}
Every run trains on a fixed length of the most recent history before each test
window, not on all the data loaded (Table~\ref{tab:data-spans}): 156 weeks on
Dominick's, 90 on Frat and 730 days on M5 and Favorita. Each data set therefore
holds history no final run trains on: Dominick's from September 1989 to the
start of training in September 1992, M5 from January 2011 to November 2013, and
Favorita two years before 1 January 2015, the start set when Favorita was added
in \thisrev{}. The final 24 weeks (84 days on M5) of each data set but Favorita are
held out of the search and of every final run; the \fvpaper{} scored its earlier models once on
their last 8 weeks (28 days on M5). Over the same span a daily series has seven times the points of a weekly one;
over their training lengths an M5 series has 730 points against 156 for
Dominick's, though M5's training covers two years and Dominick's three.

\begin{table*}[t]
\centering
\caption{\textbf{What each data set holds and what the runs use.} The source data, the part the paper loads, the
training length of the final runs and the earliest period it reaches, the rolling test windows, and the final periods
held out of every run. Dominick's: the six categories share their dates; training counts weeks with data, so it starts earlier than the
calendar would give where a category has weeks without data, and the table gives the earliest start of the six.
M5's last 28 days are in the organisers' test file and are not loaded. Everything loaded before the training
start is not trained on by the final runs.}
\label{tab:data-spans}
\footnotesize\setlength{\tabcolsep}{4pt}
\begin{adjustbox}{max width=\textwidth}\begin{tabular}{llllll}
\toprule
Data & Source holds & Loaded & Training (final runs) & Test windows & Held out \\
\midrule
M5 & Jan 2011 to Jun 2016 & Jan 2011 to May 2016 & 730 days, from Nov 2013 & 4 of 28 days, Nov 2015 to Feb 2016 & 84 days from Feb 2016 \\
Dominick's & Sep 1989 to May 1997 & Sep 1989 to May 1997 & 156 weeks, from Sep 1992 & 6 of 8 weeks, Dec 1995 to Nov 1996 & 24 weeks from Nov 1996 \\
Frat & Jan 2009 to Jan 2012 & Jan 2009 to Jan 2012 & 90 weeks, from Mar 2009 & 4 of 8 weeks, Dec 2010 to Jul 2011 & 24 weeks from Jul 2011 \\
Favorita & Jan 2013 to Aug 2017 & Jan 2015 to Aug 2017 & 730 days, from Apr 2015 & 4 of 28 days, Apr 2017 to Aug 2017 & none \\
\bottomrule
\end{tabular}\end{adjustbox}
\end{table*}

\paragraph{Measure}
We use the weighted root mean squared scaled error (WRMSSE) of
M5~\cite{m5accuracy} at each level of the hierarchy: 12 levels on M5
(Figure~\ref{fig:cube}b) and six on the others. For a level with series $j$,
history $y_{j,1},\dots,y_{j,T}$ before the window and forecasts $\hat y_{j,t}$
for the $H$ periods after it,
\begin{gather}
\mathrm{RMSSE}_j = \sqrt{\frac{\tfrac1H\sum_{t=T+1}^{T+H}(y_{j,t}-\hat y_{j,t})^2}
{\tfrac{1}{T-t_j}\sum_{t=t_j+1}^{T}(y_{j,t}-y_{j,t-1})^2}},\\
\mathrm{WRMSSE} = \textstyle\sum_j w_j\,\mathrm{RMSSE}_j ,
\end{gather}
where $t_j$ is the series' first period with sales and $w_j$ is its share of the
level's dollar sales (units on Favorita) over the last weeks of training. We
report item by store, where orders and promotions are decided, the chain total,
which a financial plan is set against, and the mean over all levels (the M5
score). Each Dominick's data set is one category, so its category level repeats
the total and its category-by-store level repeats the store level; the results
count each distinct level once, four levels on Dominick's. Lower is better.

% =====================================================================
\section{Models}
\label{sec:models}

Table~\ref{tab:models} lists every model. The world model and LightGBM were
each tuned on every data set by a common search procedure (Section~\ref{sec:loop}); their
final settings are in Table~\ref{tab:final-settings}, and the constraints each
works within are in Appendix~\ref{sec:constraints}. Each tuned model's forecast
is the average of five fits with different seeds, except on M5, where each
model is a single fit to bound the cost of the larger data set.

\begin{table}[t]
\centering
\caption{\textbf{Every forecast compared.} Plan: sees the plan for the
forecast periods. Trained: fitted on the data set. Tuned: settings set by the
search of Section~\ref{sec:loop}.}
\label{tab:models}
\footnotesize
\renewcommand{\arraystretch}{1.15}
\setlength{\tabcolsep}{3pt}
\begin{tabular}{p{0.30\columnwidth}cccp{0.33\columnwidth}}
\toprule
\textbf{Model} & \textbf{Plan} & \textbf{Trained} & \textbf{Tuned} & \textbf{Technique} \\
\midrule
Average & yes & yes & yes & Equal-weight mean of the world model and LightGBM \\
Three-model ensemble & yes & yes & yes & Equal-weight mean of the world model, LightGBM and TimesFM with the plan \\
World model & yes & yes & yes & Item state rolled forward under the plan \\
LightGBM & yes & yes & yes & Direct gradient-boosted trees \\
Temporal Fusion Transformer & yes & yes & no & Attention over history and plan \\
TimesFM with the plan & yes & no & no & TimesFM~2.5 with its covariate method \\
Timestral with the plan & yes & no & no & Timestral-28M with known future inputs \\
TimesFM & no & no & no & TimesFM~2.5, 512 periods of units \\
Chronos-Bolt & no & no & no & Chronos-Bolt base \\
Seasonal naive & no & no & no & Sales one season earlier \\
\bottomrule
\end{tabular}
\end{table}

\subsection{World model}

We use \emph{world model} for an action-conditioned state model: a learned
state for each item, carried forward period by period under the plan, which
plays the part of the action, and read out as sales. The same model can be
rolled forward under a plan that was never made, which is the question a
merchant asks of it. It is trained end to end on sales. The objective that
trains the state to predict its own later values, the latent term of
Section~\ref{sec:losses} and JEPA pretraining, was tried on every data set; the
latent check, seven steps ahead, was kept on Favorita and JEPA pretraining on snacks, and
neither on the other seven (Section~\ref{sec:loop}). The case
for the model therefore rests on its accuracy and on how its carried state
responds to a plan, which the carried-state tests describe
(Section~\ref{sec:sup-mechanism}), not on that objective; how much accuracy the
state adds over the same network without it was not measured.

\paragraph{Encoder and state}
For item $i$ in one store, an encoder (a transformer over the last $L$
periods) reads sales and price, each scaled by the item's recent level
$\ell_i$, together with the calendar and the recorded levers, and gives a
state $s_t$; the item and store labels enter the readout. The item's level is
its average over the history window, or, on Dominick's, over its last 8 weeks.

\paragraph{Transition}
A gated recurrent transition $f$ carries the state forward under the plan
$p_{t+h}$ for each future period,
\begin{equation}
s_{t+h}(p) = \mathrm{LN}\big(f(p_{t+h},\, s_{t+h-1}(p))\big), \quad s_t(p) = s_t,
\end{equation}
where LN is layer normalisation (used on M5, Dominick's and Favorita). The plan
holds the period's price as a share of the item's regular price (its highest
price in the last 12 periods) and the recorded levers.

\paragraph{Readout and loss}
Expected sales are a baseline under the ``nothing planned'' plan $p^0$ (regular
price, no lever) times a lift,
\begin{equation}
\begin{split}
\log \mu_{i,t+h} = {} & \log \ell_i + b\big(s_{t+h}(p^0), p^0_{t+h}\big) \\
 & + g\big(s_{t+h}(p), p_{t+h}\big) - g\big(s_{t+h}(p^0), p^0_{t+h}\big),
\end{split}
\end{equation}
with $b$ and $g$ small networks that also take the item's labels and the
horizon step, and the log forecast bounded to $[-10, 10]$. The lift is exactly
zero when nothing is planned, and the baseline does not move when the plan
changes; both are checked by tests in the repository. Training minimises the
loss of Section~\ref{sec:losses} on sales relative to $\ell_i$, with batches of
item, store and date drawn at random.

\paragraph{Brand switching (Dominick's)}
In a store and week where several items are promoted, much of each item's lift
is taken from the others. Following attraction models~\cite{bell1975,cooper1988},
the world model fits on the last 8 training weeks one number per run, the share
$\alpha$ of the summed lift that the category total keeps, and rescales its item
forecasts in each store and week so that the category total is the baseline
total plus $\alpha$ times the summed lift: with baselines $b_i$, lifts
$\lambda_i$ and forecasts $\hat y_i = b_i + \lambda_i$ for the items $i$ of one
store and week,
\begin{equation}
\hat y_i' = \hat y_i\,\frac{\sum_k b_k + \alpha\sum_k \lambda_k}{\sum_k \hat y_k},
\end{equation}
so each item keeps its share of the category forecast. With $\Lambda_g$ the summed
lift and $G_g$ the actual minus baseline sales of store-week group $g$ in the
last 8 training weeks, $\alpha$ is the least-squares slope
$\sum_g \Lambda_g G_g / \sum_g \Lambda_g^2$, kept between 0 and 1.
$\alpha$ is a forecast-implied split
of lift, not an identified estimate of where promoted sales come from, because
promotions are not assigned at random.

\begin{table*}[t]
\centering\footnotesize\setlength{\tabcolsep}{4pt}
\caption{The tuned settings used for every result of Section~\ref{sec:results}, read from the final configurations.
Copies: forecasts averaged over that many fits with different seeds. Size: state width by encoder layers. Steps:
training steps (and pretraining steps on the latent objective, where used). Periods: training periods read. The
changes each search kept are listed in Table~\ref{tab:loop}.}
\label{tab:final-settings}
\begin{tabular}{l ccccc cccc}
\toprule
 & \multicolumn{5}{c}{World model} & \multicolumn{4}{c}{LightGBM}\\
\cmidrule(lr){2-6}\cmidrule(lr){7-10}
Data set & Copies & Size & Steps & Periods & Loss & Copies & Rounds & Periods & Objective\\
\midrule
M5 & 1 & 128$\times$3 & 30,000 & 730 & Hurdle negative binomial & 1 & 800 & 730 & Tweedie \\
Favorita & 5 & 128$\times$3 & 30,000 & 730 & Hurdle negative binomial & 5 & 800 & 730 & Poisson \\
Frat & 5 & 64$\times$2 & 3,000 & 90 & Tweedie & 5 & 500 & 90 & Tweedie \\
Analgesics & 5 & 64$\times$2 & 6,000 & 156 & Tweedie & 5 & 500 & 156 & Tweedie \\
Bottled juice & 5 & 64$\times$2 & 6,000 & 156 & Tweedie & 5 & 500 & 156 & Poisson \\
Cheese & 5 & 64$\times$2 & 6,000 & 156 & Tweedie & 5 & 500 & 156 & Tweedie \\
Frozen entrees & 5 & 64$\times$2 & 6,000 & 156 & Tweedie & 5 & 500 & 156 & Tweedie \\
Soft drinks & 5 & 64$\times$2 & 6,000 & 156 & Tweedie & 5 & 500 & 156 & Tweedie \\
Snacks & 5 & 64$\times$2 & 6,000, pretrained 18,000 steps & 156 & Negative binomial & 5 & 1000 & 156 & Tweedie \\
\bottomrule
\end{tabular}
\end{table*}

\subsection{Training losses}
\label{sec:losses}

\paragraph{The loss used}
The world model is trained on sales relative to the item's level,
$z = y/\ell_i$, by default with the Tweedie deviance for a mean $\mu$ and power
$p = 1.5$,
\begin{equation}
L_p(z, \mu) = -\frac{z\,\mu^{1-p}}{1-p} + \frac{\mu^{2-p}}{2-p},
\end{equation}
the negative log-likelihood, up to terms free of $\mu$, of a compound
Poisson--gamma distribution with a point mass at zero~\cite{jorgensen1987}.
Because the loss is on $z$, every item counts alike whatever its volume. The
search replaced it on three data sets (Table~\ref{tab:final-settings}): on M5
and Favorita by a hurdle negative binomial, a probability of any sale times a
negative binomial for the units sold when there is one, and on snacks by a
negative binomial; in each the spread of the negative binomial is learnt for
each item. LightGBM uses the Tweedie deviance with $p = 1.1$ on units, so large
items count for more, or the Poisson deviance where its search chose it, and the
Temporal Fusion Transformer a negative binomial likelihood on units.

\ifjournal
The changes to the loss tried in the \firstver{} and again in the search are
described and scored in Section~\ref{sec:sup-method} of the online supplement.

\else

\fi

\subsection{Average and ensemble}
The average is the equal-weight mean of the tuned world model's and the tuned
LightGBM's forecasts for each item, store and period, with no weights fitted.
The three-model ensemble adds TimesFM with the plan, also with equal weights; it
was fixed before any result of the \revn{}, and is reported
beside the average. The forecast by history of Section~\ref{sec:rq-broad} uses
the world model alone for series with less than 26 weeks of sales and the
average for the rest.

\subsection{Comparison models}
The Temporal Fusion Transformer and the pretrained models are untuned reference
models: the Temporal Fusion Transformer keeps the settings of the \firstver{},
and the pretrained models run as published. Searching each of them on nine data
sets as the two tuned models were searched was beyond the compute of this study,
so the comparisons that decide the rule of Section~\ref{sec:rq-broad} are
between the two tuned models and their average.
\emph{LightGBM}~\cite{lightgbm}: a direct model, one per store on M5 and
Favorita, reading sales at lags no shorter than the horizon, rolling averages and
spread of sales, price against the item's highest recent price, the calendar and
the plan inputs, with the item as a label; tuned as in Table~\ref{tab:final-settings}.
\emph{Temporal Fusion Transformer}~\cite{tft}: one model per data set,
fitted on the last 156 weeks (730 days with 112 days of input on M5 and
Favorita) with the plan as known future inputs and a negative binomial
likelihood, 1,500 steps; on M5 its four windows took 14 minutes on one GPU. \emph{TimesFM}~\cite{timesfm,timesfm25}:
the 200M-parameter TimesFM~2.5 checkpoint given each series' last 512 periods;
we use its mean output, chosen in the \firstver{} because its median forecast
fell well below M5 totals, as the median of intermittent daily sales does. \emph{TimesFM with the plan}: the same model
through its covariate method, a linear model on the plan inputs fitted over
groups of 2,048 series, then TimesFM on what that model leaves.
\emph{Chronos-Bolt}~\cite{chronos}: the base checkpoint, zero-shot, the mean of
its quantiles. \emph{Timestral}~\cite{timestral}: Timestral-28M (code commit
a534bdb), zero-shot, with the plan as known future inputs (price relative to
the context's highest price, other levers as given) and from history alone;
the point forecast is the mean of its 21 quantiles, each weighted by the share
of probability nearest it. \emph{Seasonal naive}: sales one season earlier.

% =====================================================================
\section{Experimental setup}
\label{sec:setup}

\paragraph{Windows}
Every result of Section~\ref{sec:results} is scored on rolling windows that no
choice of setting ever saw: six 8-week windows on each Dominick's category, four
on Frat, and four 28-day windows on M5 (November 2015 to February 2016) and on
Favorita (the last ending on its last day), each forecast from everything
earlier. The searches of Section~\ref{sec:loop} scored windows before these,
and held the latest of their own windows out of every decision. The windows used
to develop \ifjournal the models' first settings\else the \firstver{} of the models\fi{} lie after the rolling windows and are
not scored in this paper (Figure~\ref{fig:study}a).

\paragraph{Rules fixed in advance}
The \revn{} plan, with every window, model, test and pass rule, was written down
before the \revn{}'s first run, and every change to it is a dated entry,
made before the results it governs were read unless the entry says otherwise.
Each data set's world model and LightGBM run in every rolling window with the
final settings of their searches (Table~\ref{tab:final-settings}), unchanged
from window to window; each forecast is the average of five fits with different
seeds (a single fit on M5). No result on the rolling windows changed a
setting, and every model, window and test is reported.

\paragraph{Comparing two models}
For models $A$ and $B$, window $k$ and a level, let $d_k = \mathrm{WRMSSE}_k(A) -
\mathrm{WRMSSE}_k(B)$. Within a window, $d_k$ gets a 95\% interval from 1,000
resamples of items: an item drawn twice counts twice in every series and
aggregate it feeds, with the weights recomputed. Across the $K$ windows, the
mean $\bar d = \frac1K\sum_k d_k$ gets a 95\% interval from the 2.5\% and 97.5\%
quantiles of $\bar d^{*}$ over 5,000 resamples of windows with replacement. A
model is said to beat another only when that interval excludes zero;
when it does not, we say no difference was detected, which is not a finding that
the two are equal. For each data set and level we report the 90\% model
confidence set~\cite{hansen2011}: starting from all models $\mathcal M$, the range
statistic $T = \max_{a,b\in\mathcal M} |\bar d_{ab}|/\hat\sigma_{ab}$, with
$\hat\sigma_{ab}$ from the bootstrap over windows, is compared with its
bootstrap distribution under equal expected loss (5,000 resamples); while the
$p$-value is below 0.10 the model whose standardised losses against the others
are largest is removed, and the models left form the set. With four to six windows these tests have little power, and adjacent
windows are dependent: each is forecast from all data before it, so they share
most of their training history, and an error tied to a season or a stretch of
trading can recur in neighbouring windows. A bootstrap that resamples whole
windows treats them as exchangeable and so understates that dependence; we
therefore also count, over levels and windows, the comparisons in which one model
is clearly more accurate than the other by the item bootstrap
(Section~\ref{sec:rq-grain}).

\ifjournal
\paragraph{Supporting analyses} Forecast ranges come from each model's own errors in
the earlier rolling windows~\cite{vovk2005}; decision costs, reconciliation and ranking are
set out with their results in the online supplement (Section~\ref{sec:sup-method}).

\else

\fi

\paragraph{Analyses added after the first results}
Several analyses were added after the first scores of the tuned models were
read, each written down before it ran; Section~\ref{sec:limits} lists them all.
The transfer test and the pooled training of RQ4 were written down before any
tuned model was scored.

% The research loop: how both models' settings were searched (revision, November 2026).
% Numbers come from paper/generated/tab_loop.tex and the figures, all built from results/research/*/log.jsonl
% by paper/figs/make_loop_figures.py.
\section{Searching the settings of both models}
\label{sec:loop}

Both models were tuned with the same procedure on every data set. On the windows the searches
scored, tuning improved LightGBM on all \loopImprovedLgb{} data sets and the world model on \loopImprovedWm{}, and the gains
were mostly small next to the differences between data sets. The held-out window agreed with the
search windows only weakly, so the paper judges the models on later windows the searches never
saw. Table~\ref{tab:loop} gives every search from its starting settings to its final settings, and
Figure~\ref{fig:looppath} shows every run in the order it was made.

\subsection{The procedure}

The procedure is a fixed loop, written down before it ran, that changes one setting at a time. It
starts from the settings the \fvpaper{} used. Each change on a menu fixed in
advance is run once, on one seed, on top of the best settings so far. Every change that beats the
best settings on that seed is run on two more seeds. The best of those is kept if its mean over
three seeds beats the best settings' mean by at least the keep bar, which is the spread of the
three seeds of the starting settings. The loop then starts a new round from the new best and stops
when a round keeps nothing. A change cannot be kept because one seed was lucky, and the bar is set
by the model's own noise on that data set before any change is tried.

The windows the loop scores lie before every window this paper reports. The latest of them is held
out: it is scored for every three-seed run and never used to decide. The menus differ between the
models, because the models have different settings, but each covers what a practitioner would try
first: for the world model the loss weights, the roll-forward, training length, learning rate and
batch, weight averaging and the latent checks of Section~\ref{sec:models}; for LightGBM the
number of leaves, the learning rate and number of rounds, the minimum data in a leaf, the
objective, feature sampling and the lag windows.

After its loop each world model had two more rounds of one change each. The first tried four count
likelihoods that treat weeks with no sales on their own: the negative binomial, the same with a
spread for each item, the zero-inflated negative binomial and the hurdle negative binomial. The
second tried JEPA pretraining of the encoder with the forecast head fitted afterwards at a lower
rate, at two lengths of pretraining. Both used the keep rule above. LightGBM has no counterpart
to either, so it had none. Pretraining used only each data set's own history, so the two models
still saw the same data.

Each model on each data set was searched as a five-copy average, the mean forecast of five fits
with different seeds, except on M5, where both models were single fits. Section~\ref{sec:loopnoise} shows why the
average matters.

\begin{table*}[t]
\centering\footnotesize
\caption{Every search, from starting to final settings. Scores are WRMSSE (lower is better), the mean of three seeds on the search windows and one score on the held-out window, which never decided anything. The keep bar is the seed spread of the starting settings of the first round; each later round (likelihoods, pretraining) measures its own bar from its own starting settings, and a change was kept only when its three-seed mean beat the best so far by at least that bar. Tries counts every run, first-seed screens and three-seed confirmations.}
\label{tab:loop}
\begin{tabular}{llrrrrrrp{5.2cm}}
\toprule
Data & Model & Tries & Keep bar & \multicolumn{2}{c}{Starting} & \multicolumn{2}{c}{Final} & Kept, in order \\
& & & & Search & Held-out & Search & Held-out & \\
\midrule
M5 & World model & 59 & 0.013 & 0.570 & 0.550 & 0.533 & 0.534 & each period's sales fed into the next, hurdle negative binomial, spread by item \\
 & LightGBM & 51 & 0.003 & 0.603 & 0.553 & 0.556 & 0.522 & a quarter of the leaves \\
\addlinespace[2pt]
Favorita & World model & 60 & 0.028 & 0.547 & 0.600 & 0.468 & 0.550 & latent check over seven steps, hurdle negative binomial, spread by item \\
 & LightGBM & 99 & 0.009 & 0.922 & 0.576 & 0.580 & 0.573 & four times the data per leaf, Poisson objective, half the learning rate \\
\addlinespace[2pt]
Frat & World model & 53 & 0.028 & 0.588 & 0.591 & 0.546 & 0.607 & half the batch \\
 & LightGBM & 52 & 0.003 & 0.501 & 0.546 & 0.489 & 0.546 & double the learning rate \\
\addlinespace[2pt]
Analgesics & World model & 35 & 0.004 & 0.509 & 0.619 & 0.509 & 0.619 & nothing \\
 & LightGBM & 52 & 0.010 & 0.548 & 0.757 & 0.528 & 0.764 & a quarter of the leaves \\
\addlinespace[2pt]
Bottled juice & World model & 30 & 0.028 & 0.587 & 0.637 & 0.587 & 0.637 & nothing \\
 & LightGBM & 67 & 0.008 & 0.708 & 0.582 & 0.674 & 0.593 & Poisson objective, 80\% of features per tree \\
\addlinespace[2pt]
Cheese & World model & 33 & 0.034 & 0.392 & 0.714 & 0.392 & 0.714 & nothing \\
 & LightGBM & 49 & 0.002 & 0.352 & 0.833 & 0.346 & 0.839 & a longer rolling window \\
\addlinespace[2pt]
Frozen entrees & World model & 95 & 0.007 & 0.246 & 0.251 & 0.216 & 0.241 & loss on the summed total, weight 0.1, half the batch \\
 & LightGBM & 53 & 0.010 & 0.222 & 0.183 & 0.181 & 0.195 & four times the leaves \\
\addlinespace[2pt]
Soft drinks & World model & 65 & 0.020 & 0.594 & 0.579 & 0.572 & 0.522 & double the learning rate \\
 & LightGBM & 80 & 0.004 & 0.554 & 0.435 & 0.532 & 0.507 & a quarter of the leaves, 255 bins per feature \\
\addlinespace[2pt]
Snacks & World model & 42 & 0.018 & 0.439 & 0.568 & 0.400 & 0.608 & negative binomial, spread by item, long JEPA pretraining \\
 & LightGBM & 54 & 0.006 & 0.358 & 0.425 & 0.351 & 0.424 & double the rounds \\
\addlinespace[2pt]
\bottomrule
\end{tabular}
\end{table*}

\ifjournal\else\fi

% =====================================================================
\section{Results}
\label{sec:results}

% Results by research question (Framework section). Tables and every number from scripts/make_rq_tables.py;
% paired counts from results/analysis_tuned/paired.csv (item bootstrap within each window, 95% intervals).

The results follow the four research questions. Every number comes from the
tuned world model and the tuned LightGBM, each set by a common search procedure on each
data set (Section~\ref{sec:loop}), scored on the paper's windows. The average is
the equal-weight mean of the two forecasts.

\subsection{RQ1: grain}
\label{sec:rq-grain}

\textbf{Use the average at every grain.} Across the nine data sets and the
three views of Table~\ref{tab:rq-grain} (store grain and above by period, item
grain by period, and the whole window as one block), the average has the lowest
error in \rqAvgBestCells{} of \rqCells{} cases and is never the least accurate of the three. The world
model is the least accurate in \rqWmWorstCells{} cases and LightGBM in \rqLgbWorstCells{}.

The paired tests give the same answer. They compare two forecasts level by level
and window by window, and count a difference only when it holds after
resampling items. Against the world model, the average is clearly more accurate
in \rqPAvgWmBetter{} of \rqPAvgWmN{} comparisons and clearly less accurate in \rqPAvgWmWorse{}; against LightGBM, \rqPAvgLgbBetter{}
and \rqPAvgLgbWorse{}. At item grain it is clearly less accurate than either model in at most \rqAvgItemWorseMax{}
of \rqPAvgWmItemn{} comparisons. Across data sets and levels it is in the 90\% model
confidence set in \rqMcsAverage{} of \rqMcsCells{} cases, against
\rqMcsLightgbm{} for LightGBM and \rqMcsWorldModel{} for the world model. The paired
item tests are the main evidence: the window bootstrap behind the confidence set
treats adjacent windows, which share most of their training data, as
independent, and so overstates its confidence (Section~\ref{sec:setup}).

\textbf{Grain does not decide between the two models on their own.} The world
model is clearly more accurate than LightGBM in \rqPWmLgbCoarsebetter{} comparisons at store grain
and above and clearly less accurate in \rqPWmLgbCoarseworse{}; at item grain the counts are \rqPWmLgbItembetter{} and
\rqPWmLgbItemworse{}. Neither is ahead at either grain across the data sets, so a retailer that
picks one model for a grain is picking by data set, not by grain. The average
removes that choice.

\textbf{On the whole promotion block, the average is again the safe choice.} It
is lowest on \rqAvgBestBlockWord{} of \rqNineWord{} data sets and second on the other four. LightGBM is
lowest on analgesics, bottled juice and snacks, and the world model is behind it
on five of the six Dominick's categories; on M5, Frat, Favorita and soft drinks
the order turns, most at the chain total. On M5 LightGBM's block error is
\rqBlockMfiveLgb{}, against \rqBlockMfiveWm{} for the world model and \rqBlockMfiveAvg{}
for the average.

\begin{table*}[t]
\centering\footnotesize\setlength{\tabcolsep}{3pt}
\caption{RQ1, grain. WRMSSE of the tuned world model (WM), the tuned LightGBM (LGB) and their equal-weight average
(Avg), averaged over the levels at store grain and above, over the item levels, and over every level for the whole
forecast window summed as one block; mean over the paper's windows. Shading marks the lowest of the three: \colorbox{wmcell}{world model}, \colorbox{lgbcell}{LightGBM}, \colorbox{avgcell}{average}.}
\label{tab:rq-grain}
\begin{adjustbox}{max width=\textwidth}\begin{tabular}{l ccc ccc ccc}
\toprule
 & \multicolumn{3}{c}{Store and above} & \multicolumn{3}{c}{Item levels} & \multicolumn{3}{c}{Whole block}\\
\cmidrule(lr){2-4}\cmidrule(lr){5-7}\cmidrule(lr){8-10}
Data set & WM & LGB & Avg & WM & LGB & Avg & WM & LGB & Avg\\
\midrule
Analgesics & 0.688 & 0.698 & \cellcolor{avgcell}\textbf{0.678} & 0.710 & 0.705 & \cellcolor{avgcell}\textbf{0.695} & 0.552 & \cellcolor{lgbcell}\textbf{0.499} & 0.508 \\
Bottled juice & 0.569 & 0.566 & \cellcolor{avgcell}\textbf{0.539} & 0.581 & 0.596 & \cellcolor{avgcell}\textbf{0.563} & 0.540 & \cellcolor{lgbcell}\textbf{0.434} & 0.457 \\
Cheese & 0.381 & 0.471 & \cellcolor{avgcell}\textbf{0.377} & 0.623 & 0.594 & \cellcolor{avgcell}\textbf{0.578} & 0.482 & 0.444 & \cellcolor{avgcell}\textbf{0.416} \\
Frozen entrees & 0.339 & 0.327 & \cellcolor{avgcell}\textbf{0.320} & 0.375 & 0.350 & \cellcolor{avgcell}\textbf{0.344} & 0.362 & 0.337 & \cellcolor{avgcell}\textbf{0.327} \\
Soft drinks & 0.822 & 0.805 & \cellcolor{avgcell}\textbf{0.754} & 0.589 & 0.555 & \cellcolor{avgcell}\textbf{0.547} & 0.625 & 0.641 & \cellcolor{avgcell}\textbf{0.523} \\
Snacks & 1.241 & \cellcolor{lgbcell}\textbf{1.040} & 1.071 & 0.545 & 0.516 & \cellcolor{avgcell}\textbf{0.508} & 0.776 & \cellcolor{lgbcell}\textbf{0.571} & 0.612 \\
Frat & 0.471 & 0.479 & \cellcolor{avgcell}\textbf{0.449} & 0.528 & 0.524 & \cellcolor{avgcell}\textbf{0.508} & \cellcolor{wmcell}\textbf{0.345} & 0.406 & 0.364 \\
M5 & 0.660 & 0.634 & \cellcolor{avgcell}\textbf{0.591} & 0.859 & 0.886 & \cellcolor{avgcell}\textbf{0.856} & 0.598 & 0.953 & \cellcolor{avgcell}\textbf{0.589} \\
Favorita & 0.314 & 0.325 & \cellcolor{avgcell}\textbf{0.278} & 0.655 & 0.696 & \cellcolor{avgcell}\textbf{0.653} & 0.482 & 0.565 & \cellcolor{avgcell}\textbf{0.412} \\
\bottomrule
\end{tabular}\end{adjustbox}
\end{table*}

\paragraph{The held-out periods (added in \thisrev{})}
\textbf{On the periods every final run held out, the average is the safe choice
rather than the most accurate one: it is never the least accurate of the three,
but it is the most accurate on only \hoAvgBest{} of \hoSets{} data sets.} Each
final model was scored once more, after every other result, on the last 24 weeks
of each Dominick's category and of Frat and the last 84 days of M5, which no
search or setting of \thisrev{} used (Table~\ref{tab:heldout}). They are excluded from tuning, not unseen:
the \fvpaper{} scored its earlier models once on the last 8 weeks of these periods (the last 28 days
on M5), before the search of \thisrev{} began. LightGBM is the
most accurate on \hoLgbBest{} data sets and the world model on \hoWmBest{}; the
order of the three matches the rolling windows on \hoOrderSame{}. In paired
comparisons the average is clearly more accurate than LightGBM more often than
clearly less on \hoAvgBeatsLgb{} data sets, and than the world model on
\hoAvgBeatsWm. Errors are higher in the held-out periods on most Dominick's
categories than on the rolling windows. On M5's final 28 days, the competition's
evaluation period, LightGBM scores \hoMfiveFinalLGB{}, the average
\hoMfiveFinalAvg{} and the world model \hoMfiveFinalWM. A retailer choosing one
forecast for periods it has not seen gains most from the average's never losing
badly, not from its being best.

\begin{table*}[t]
\centering
\caption{\textbf{The periods every final run held out.} WRMSSE, mean over
levels and the held-out windows (three of 8 weeks on Dominick's and Frat, three of 28 days on M5), each window trained on
the history before it, beside the same models on the paper's rolling windows. Avg against LGB, Avg against WM:
comparisons (levels and windows) where the average is clearly more / less accurate (paired item bootstrap). Bold: lowest
on the held-out windows. Favorita held nothing out. Planned after the other results.}
\label{tab:heldout}
\footnotesize\setlength{\tabcolsep}{4pt}
\begin{tabular}{lccccccrr}
\toprule
 & \multicolumn{3}{c}{Held-out periods} & \multicolumn{3}{c}{Rolling windows} & & \\
\cmidrule(lr){2-4}\cmidrule(lr){5-7}
Data set & WM & LGB & Avg & WM & LGB & Avg & Avg vs LGB & Avg vs WM \\
\midrule
Analgesics & \textbf{0.714} & 0.768 & 0.726 & 0.699 & 0.702 & 0.687 & 7 / 0 & 2 / 0 \\
Bottled juice & 0.983 & \textbf{0.934} & 0.943 & 0.575 & 0.581 & 0.551 & 4 / 0 & 6 / 1 \\
Cheese & 0.840 & \textbf{0.726} & 0.761 & 0.502 & 0.532 & 0.477 & 2 / 2 & 10 / 0 \\
Frozen entrees & 0.460 & 0.460 & \textbf{0.447} & 0.357 & 0.338 & 0.332 & 6 / 2 & 7 / 0 \\
Soft drinks & \textbf{0.698} & 0.776 & 0.718 & 0.705 & 0.680 & 0.650 & 8 / 0 & 4 / 2 \\
Snack crackers & 0.713 & \textbf{0.557} & 0.593 & 0.893 & 0.778 & 0.790 & 0 / 4 & 10 / 0 \\
Frat & 0.487 & 0.460 & \textbf{0.444} & 0.490 & 0.494 & 0.469 & 3 / 0 & 10 / 0 \\
M5 & 0.595 & 0.618 & \textbf{0.587} & 0.709 & 0.697 & 0.658 & 28 / 0 & 10 / 6 \\
\bottomrule
\end{tabular}
\end{table*}

\subsection{RQ2: nature of the line}
\label{sec:rq-lines}

\textbf{Sales are concentrated, and the average holds up on both the fast lines
and the slow tail; where history is short, the world model is ahead of
LightGBM.} On the six Dominick's categories, \rqFastShareDomLo{} to \rqFastShareDomHi{}\% of the items make 80\% of
unit sales over the year before the forecast; on M5, Frat and Favorita the head
is flatter, at \rqFastShareRestLo{} to \rqFastShareRestHi{}\% (Table~\ref{tab:rq-lines}). On the fast lines the
average has the lowest error on \rqAvgBestFast{} of \rqNineWord{} data sets, and on the slow lines
on \rqAvgBestSlow{}; on snacks LightGBM is lowest on the fast lines, and on cheese the
world model is lowest on the slow lines.

On series whose first sale falls less than 26 weeks before the forecast, the
world model's error is lower than LightGBM's on \rqShortWmBeatsLgb{} of \rqNineWord{} data sets,
\rqShortExceptions{} \rqShortExceptionsVerb{}: for example \rqShortFavWm{} against \rqShortFavLgb{} on Favorita and \rqShortMfiveWm{}
against \rqShortMfiveLgb{} on M5. These series are \rqShortShareDomLo{} to \rqShortShareDomHi{}\% of the series on the Dominick's
categories, where items enter and leave the range over the years, and under \rqShortShareRestMax{}\%
on M5, Frat and Favorita.

\begin{table*}[t]
\centering\footnotesize\setlength{\tabcolsep}{3pt}
\caption{RQ2, nature of the line. Share of items that make 80\% of unit sales over the year before the forecast, and
WRMSSE averaged over every level, by period, on the fast lines (those items), the slow lines (the other items) and the
series with short history (first sale less than 26 weeks or 182 days before the forecast). WM, LGB and Avg as in
Table~\ref{tab:rq-grain}. Shading marks the lowest of the three: \colorbox{wmcell}{world model}, \colorbox{lgbcell}{LightGBM}, \colorbox{avgcell}{average}.}
\label{tab:rq-lines}
\begin{adjustbox}{max width=\textwidth}\begin{tabular}{l c ccc ccc ccc}
\toprule
 & Items making & \multicolumn{3}{c}{Fast lines} & \multicolumn{3}{c}{Slow lines} & \multicolumn{3}{c}{Short history}\\
\cmidrule(lr){3-5}\cmidrule(lr){6-8}\cmidrule(lr){9-11}
Data set & 80\% of units & WM & LGB & Avg & WM & LGB & Avg & WM & LGB & Avg\\
\midrule
Analgesics & 17\% & 0.754 & 0.748 & \cellcolor{avgcell}\textbf{0.738} & 0.756 & 0.769 & \cellcolor{avgcell}\textbf{0.748} & \cellcolor{wmcell}\textbf{1.291} & 1.459 & 1.350 \\
Bottled juice & 21\% & 0.594 & 0.606 & \cellcolor{avgcell}\textbf{0.571} & 0.513 & 0.516 & \cellcolor{avgcell}\textbf{0.494} & \cellcolor{wmcell}\textbf{1.303} & 1.393 & 1.312 \\
Cheese & 17\% & 0.505 & 0.507 & \cellcolor{avgcell}\textbf{0.461} & \cellcolor{wmcell}\textbf{0.515} & 0.584 & 0.526 & \cellcolor{wmcell}\textbf{1.222} & 1.411 & 1.284 \\
Frozen entrees & 19\% & 0.327 & 0.310 & \cellcolor{avgcell}\textbf{0.306} & 0.461 & 0.412 & \cellcolor{avgcell}\textbf{0.402} & 1.473 & 1.471 & \cellcolor{avgcell}\textbf{1.336} \\
Soft drinks & 11\% & 0.640 & 0.602 & \cellcolor{avgcell}\textbf{0.583} & 0.604 & 0.651 & \cellcolor{avgcell}\textbf{0.595} & 1.379 & 1.397 & \cellcolor{avgcell}\textbf{1.334} \\
Snacks & 17\% & 0.650 & \cellcolor{lgbcell}\textbf{0.496} & 0.524 & 1.189 & 1.154 & \cellcolor{avgcell}\textbf{1.144} & \cellcolor{wmcell}\textbf{4.204} & 4.565 & 4.328 \\
Frat & 42\% & 0.432 & 0.442 & \cellcolor{avgcell}\textbf{0.414} & 0.659 & 0.626 & \cellcolor{avgcell}\textbf{0.622} & 2.703 & 2.642 & \cellcolor{avgcell}\textbf{2.608} \\
M5 & 37\% & 0.717 & 0.700 & \cellcolor{avgcell}\textbf{0.662} & 0.735 & 0.748 & \cellcolor{avgcell}\textbf{0.702} & \cellcolor{wmcell}\textbf{0.866} & 1.024 & 0.907 \\
Favorita & 42\% & 0.417 & 0.442 & \cellcolor{avgcell}\textbf{0.395} & 0.496 & 0.522 & \cellcolor{avgcell}\textbf{0.472} & 0.539 & 0.667 & \cellcolor{avgcell}\textbf{0.525} \\
\bottomrule
\end{tabular}\end{adjustbox}
\end{table*}

\paragraph{Steady prices and frequent cuts (added in \thisrev{})}
\textbf{Within each data set, the world model's clearest edge is on lines with
short history; its weakness in price-cut weeks is a finding of Dominick's. The
average is ahead of LightGBM on every kind of line.} Inside a data set, the world
model does relatively better against LightGBM on lines with short history than
on the rest on \kShortBetterAll{} of the nine data sets, at three of the four
retailers (Frat the exception). On every Dominick's category it does relatively
worse in price-cut weeks than in other weeks (\kCutWorseDom{} of six), and worse
on lines cut often than on lines never cut on \kHeavyWorseDom{} of six. The other
retailers do not show this: on M5 its difference to LightGBM is \kDMfivePriceCutPeriods{}
in price-cut periods against \kDMfiveOtherPeriods{} in the others, on Frat
\kDFratPriceCutPeriods{} against \kDFratOtherPeriods{}, and on Favorita, which
records no prices and is split by its promotion flag, \kDFavPriceCutPeriods{}
against \kDFavOtherPeriods{} (point estimates; negative means the world model is
more accurate). Across data sets the world model is ahead of LightGBM on
\kAheadMfive{} and \kAheadFav{} of the \kKinds{} kinds on M5 and Favorita and behind
on most on Frat and five of the six Dominick's categories (Table~\ref{tab:rq-kinds});
with four retailers, that difference cannot be traced to how each is priced.
Counted by data set, the world model is clearly more accurate than LightGBM on
short-history lines on \kWmBetterShortHistory{} of nine and less on
\kWmWorseShortHistory; on lines never promoted the count is \kWmBetterNoPromotion{} against
\kWmWorseNoPromotion{}, and on slow lines \kWmBetterSlow{} against \kWmWorseSlow{}, close to even. The average is clearly more accurate than LightGBM on every kind on
at least six of the nine data sets. For a planner: the world model's own forecast
is worth most on lines with little history, the average serves every kind of line,
and on a high-low chain like Dominick's neither model's response to deep cuts
should be taken on trust (the companion paper~\cite{paper2} tests that response directly).

\begin{table*}[t]
\centering
\caption{\textbf{The world model is ahead of LightGBM on most kinds of line on M5 and Favorita and behind on most on
Frat and Dominick's; the average is ahead almost everywhere.} Each mark compares the world model with LightGBM on the item-by-store series (or periods)
of one kind, over the rolling windows of the final runs: $+$ clearly more accurate, $-$ clearly less, $\cdot$ not
clearly different (95\% interval of the difference in revenue-weighted RMSSE, 1,000 resamples of series). Every kind is
fixed from the year before each window; a price cut is a price at most 95\% of the line's most common price that year
(Favorita, which records no prices: its promotion flag). The last columns count data sets where the world model, and
the average of the two models, are clearly more or less accurate than LightGBM. Added in \thisrev{} after the 52-week
test.}
\label{tab:rq-kinds}
\footnotesize\setlength{\tabcolsep}{3.5pt}
\begin{tabular}{lccccccccccccc}
\toprule
 & \multicolumn{9}{c}{World model against LightGBM} & \multicolumn{2}{c}{WM} & \multicolumn{2}{c}{Avg} \\
\cmidrule(lr){2-10}\cmidrule(lr){11-12}\cmidrule(lr){13-14}
Kind of line or period & Ana & Bjc & Che & Fre & Sdr & Sna & Frat & M5 & Fav & ahead & behind & ahead & behind \\
\midrule
Short history & $+$ & $+$ & $+$ & $-$ & $\cdot$ & $+$ & $\cdot$ & $+$ & $+$ & 6 & 1 & 8 & 0 \\
Long history & $-$ & $+$ & $-$ & $-$ & $-$ & $-$ & $-$ & $+$ & $+$ & 3 & 6 & 7 & 0 \\
Slow lines & $-$ & $+$ & $+$ & $-$ & $+$ & $-$ & $-$ & $+$ & $+$ & 5 & 4 & 9 & 0 \\
Fast lines & $-$ & $+$ & $-$ & $-$ & $-$ & $-$ & $-$ & $+$ & $+$ & 3 & 6 & 7 & 0 \\
Sells in under half the periods & $\cdot$ & $+$ & $+$ & $-$ & $\cdot$ & $-$ & $\cdot$ & $+$ & $+$ & 4 & 2 & 9 & 0 \\
Sells in nine periods of ten or more & $-$ & $\cdot$ & $-$ & $-$ & $-$ & $-$ & $-$ & $\cdot$ & $+$ & 1 & 6 & 7 & 2 \\
Steady price (no cut) & $-$ & $+$ & $+$ & $-$ & $+$ & $-$ & $\cdot$ & $+$ & $+$ & 5 & 3 & 7 & 1 \\
Occasional cuts (under a quarter of periods) & $-$ & $+$ & $-$ & $-$ & $\cdot$ & $-$ & $\cdot$ & $+$ & $+$ & 3 & 4 & 8 & 1 \\
Frequent cuts (a quarter of periods or more) & $-$ & $\cdot$ & $-$ & $-$ & $-$ & $-$ & $-$ & $+$ & $+$ & 2 & 6 & 8 & 0 \\
Periods at regular price & $-$ & $+$ & $-$ & $+$ & $-$ & $\cdot$ & $-$ & $+$ & $+$ & 4 & 4 & 9 & 0 \\
Price-cut periods & $-$ & $\cdot$ & $-$ & $-$ & $-$ & $-$ & $\cdot$ & $\cdot$ & $+$ & 1 & 5 & 6 & 1 \\
\bottomrule
\end{tabular}
\end{table*}

\paragraph{More history (added in \thisrev{})}
\textbf{Training on more history does not close the world model's gap to
LightGBM in any consistent way; the average stays the most accurate.} Each
model was trained on one to five years of history before each window, with every
other setting kept (Table~\ref{tab:history}). The rule for reading it was fixed
before the runs: years of history help the world model if its gap to LightGBM
shrinks with length on Dominick's, and observations per series help if it shrinks
only on the daily data. On Dominick's the gap shrinks with more years on
\histDomShrinks{} of \histDomSets{} categories and grows on the rest. On the two
daily data sets it moves in opposite directions: on M5 the world model is ahead
of LightGBM with \histMfiveYearsShort{} year of history (gap \histMfiveGapShort) and
behind with \histMfiveYearsLong{} (\histMfiveGapLong), while on Favorita it is behind
with \histFavYearsShort{} (\histFavGapShort) and ahead with \histFavYearsLong{}
(\histFavGapLong). The world model is ahead with the least history on
\histAheadShort{} of \histSets{} data sets and with the most on \histAheadLong. Neither
years of history nor the number of observations per series decides which model
is ahead; both models gain from more history, and the average of the two is the
most accurate in \histAvgBest{} of the \histCells{} combinations of data set and
length.

\begin{table}[t]
\centering
\caption{\textbf{Training on more history.} WRMSSE, mean over levels and windows, of each model trained on the given
years of history before each window, settings otherwise those of the final runs (Dominick's at 3 years and M5 at 2 are
the final runs; Favorita, loaded from 2013 for this test, at every length). Gap: world model minus LightGBM (negative,
the world model is more accurate). Clear: comparisons (levels and windows) where the world model is clearly more / less
accurate than LightGBM. Bold: lowest. Planned and added in \thisrev{}.}
\label{tab:history}
\footnotesize\setlength{\tabcolsep}{3.5pt}
\begin{tabular}{lrcccrc}
\toprule
Data & Years & WM & LGB & Avg & Gap & Clear \\
\midrule
Analgesics & 1 & \textbf{0.730} & 0.801 & 0.731 & $-$0.071 & 1 / 3 \\
 & 2 & 0.702 & 0.715 & \textbf{0.693} & $-$0.013 & 0 / 3 \\
 & 3 & 0.699 & 0.702 & \textbf{0.687} & $-$0.003 & 1 / 3 \\
 & 5 & 0.701 & 0.702 & \textbf{0.682} & $-$0.001 & 3 / 3 \\
\addlinespace[2pt]
Bottled juice & 1 & 0.706 & 0.631 & \textbf{0.605} & +0.075 & 2 / 3 \\
 & 2 & 0.631 & 0.554 & \textbf{0.551} & +0.077 & 0 / 5 \\
 & 3 & 0.575 & 0.581 & \textbf{0.551} & $-$0.006 & 2 / 0 \\
 & 5 & 0.567 & 0.569 & \textbf{0.545} & $-$0.002 & 5 / 2 \\
\addlinespace[2pt]
Cheese & 1 & 0.584 & 0.564 & \textbf{0.516} & +0.020 & 4 / 5 \\
 & 2 & 0.537 & 0.541 & \textbf{0.502} & $-$0.004 & 0 / 6 \\
 & 3 & 0.502 & 0.532 & \textbf{0.477} & $-$0.030 & 1 / 5 \\
 & 5 & 0.512 & 0.528 & \textbf{0.479} & $-$0.016 & 3 / 4 \\
\addlinespace[2pt]
Frozen entrees & 1 & 0.358 & 0.365 & \textbf{0.345} & $-$0.007 & 8 / 13 \\
 & 2 & 0.361 & 0.334 & \textbf{0.330} & +0.027 & 1 / 12 \\
 & 3 & 0.357 & 0.338 & \textbf{0.332} & +0.018 & 4 / 15 \\
 & 5 & 0.377 & \textbf{0.333} & 0.338 & +0.044 & 4 / 16 \\
\addlinespace[2pt]
Soft drinks & 1 & 0.734 & 0.708 & \textbf{0.673} & +0.027 & 2 / 4 \\
 & 2 & 0.699 & 0.669 & \textbf{0.643} & +0.030 & 6 / 8 \\
 & 3 & 0.705 & 0.680 & \textbf{0.650} & +0.026 & 4 / 7 \\
 & 5 & 0.699 & 0.682 & \textbf{0.658} & +0.017 & 6 / 5 \\
\addlinespace[2pt]
Snack crackers & 1 & 0.898 & 0.844 & \textbf{0.828} & +0.053 & 3 / 7 \\
 & 2 & 0.880 & 0.795 & \textbf{0.783} & +0.086 & 3 / 5 \\
 & 3 & 0.893 & \textbf{0.778} & 0.790 & +0.115 & 2 / 8 \\
 & 5 & 0.865 & \textbf{0.770} & 0.779 & +0.094 & 2 / 6 \\
\addlinespace[2pt]
M5 & 1 & 0.815 & 0.844 & \textbf{0.690} & $-$0.029 & 16 / 10 \\
 & 2 & 0.709 & 0.697 & \textbf{0.658} & +0.012 & 27 / 11 \\
 & 4 & 0.687 & 0.667 & \textbf{0.642} & +0.019 & 27 / 12 \\
\addlinespace[2pt]
Favorita & 1 & 0.488 & 0.468 & \textbf{0.452} & +0.020 & 8 / 12 \\
 & 2 & 0.446 & 0.479 & \textbf{0.425} & $-$0.033 & 15 / 8 \\
 & 4 & 0.434 & 0.458 & \textbf{0.423} & $-$0.024 & 13 / 4 \\
\bottomrule
\end{tabular}
\end{table}

\subsection{RQ3: seen and unseen data}
\label{sec:rq-unseen}

\textbf{On items the models were not trained on, use the world model alone: it is
more accurate there than the average, and than LightGBM with or without the item
label on most data sets tested; on
stores they were not trained on, the world model and the average are level, and
the average serves.} With a fifth of the items
kept out of training, the world model's error on them rises by \rqCostWmItemLo{} to \rqCostWmItemHi{},
against \rqCostLgbItemLo{} to \rqCostLgbItemHi{} for LightGBM (Table~\ref{tab:rq-transfer}). The tuned
LightGBM reads the item label as a feature, so on a kept-out item it meets a label
it never saw; it under-forecasts these items, at \rqFaLgbItemLo{} to \rqFaLgbItemHi{}\% of actual sales, and that pulls the
average down to \rqFaAvgItemLo{} to \rqFaAvgItemHi{}\%; the world model forecasts \rqFaWmItemLo{} to \rqFaWmItemHi{}\%. On the kept-out
items the world model alone is clearly more accurate than the average on \rqItemWmBeatsAvg{} of \rqKeptSets{}
data sets tested.

\textbf{Most of LightGBM's loss on these items comes from the item label, and the
world model stays ahead without it.} As planned before it ran, LightGBM was fitted
again on the kept-out item tests without the item label, every other setting
unchanged (Table~\ref{tab:rq-noid}). Its cost of not training on the items falls to
\rqNoidCostLo{} to \rqNoidCostHi{}, and it forecasts \rqNoidFaLo{} to \rqNoidFaHi\% of actual sales on them. On the
kept-out items the world model is still clearly more accurate on \rqWmBeatsNoid{} of
\rqNoidSets{} data sets and level on \rqWmNoidLevelSets{}; LightGBM without the label is clearly more
accurate on \rqNoidBeatsWm{}. Without the label, LightGBM is less accurate on the same items
when it is trained on them, on \rqNoidSeenWorse{} of \rqNoidSets{} data sets. LightGBM therefore serves one set of items or the other: with
the label it serves the items it was trained on, without it the items it was not.
The world model also takes the item label, but it reads each item's recent sales
into its state, so a label it never saw costs it little and one model serves both.
The gap in Table~\ref{tab:rq-transfer} overstates what the world model's design adds: most of it
is LightGBM meeting a label it never saw.

With a fifth of the stores kept out, the world model is clearly more accurate
than LightGBM on \rqStoreWmBeatsLgb{} of \rqKeptSets{} data sets and level on M5. Against the average it
has the lower error on \rqStoreWmLowest{} of \rqKeptSets{} but is clearly more accurate only on Frat; the
average is clearly more accurate on cheese, and the two are level on snacks, M5
and Favorita.

\begin{table*}[t]
\centering\footnotesize\setlength{\tabcolsep}{3pt}
\caption{RQ3, seen and unseen data. A fifth of the items, or of the stores, is kept out of training. WRMSSE on that
fifth for the world model (WM), LightGBM (LGB) and their average (Avg); the cost of not training on them, which is that
WRMSSE minus the WRMSSE on the same fifth of the same forecast trained on everything; and forecast over actual on the
fifth. Means over windows. The last two columns name the forecast that is clearly lower on the kept-out fifth when a
90\% bootstrap interval over windows of the difference excludes zero. Shading marks the lowest of the three: \colorbox{wmcell}{world model}, \colorbox{lgbcell}{LightGBM}, \colorbox{avgcell}{average}.}
\label{tab:rq-transfer}
\begin{adjustbox}{max width=\textwidth}\begin{tabular}{ll ccc ccc ccc ll}
\toprule
 & & \multicolumn{3}{c}{WRMSSE, kept out} & \multicolumn{3}{c}{Cost of not training on them} & \multicolumn{3}{c}{Forecast over actual} & \multicolumn{2}{c}{Clearly lower}\\
\cmidrule(lr){3-5}\cmidrule(lr){6-8}\cmidrule(lr){9-11}\cmidrule(lr){12-13}
Data set & Kept out & WM & LGB & Avg & WM & LGB & Avg & WM & LGB & Avg & WM or LGB & WM or Avg\\
\midrule
Cheese & Items & \cellcolor{wmcell}\textbf{0.678} & 0.731 & 0.687 & 0.011 & 0.058 & 0.035 & 0.99 & 0.93 & 0.96 & \cellcolor{wmcell}World model & \cellcolor{wmcell}World model \\
Cheese & Stores & 0.635 & 0.657 & \cellcolor{avgcell}\textbf{0.612} & 0.001 & 0.047 & 0.014 & 1.05 & 0.94 & 1.00 & \cellcolor{wmcell}World model & \cellcolor{avgcell}Average \\
Snacks & Items & \cellcolor{wmcell}\textbf{0.626} & 0.753 & 0.663 & 0.012 & 0.162 & 0.075 & 0.89 & 0.68 & 0.78 & \cellcolor{wmcell}World model & \cellcolor{wmcell}World model \\
Snacks & Stores & \cellcolor{wmcell}\textbf{0.611} & 0.700 & 0.621 & 0.001 & 0.102 & 0.034 & 1.00 & 0.72 & 0.86 & \cellcolor{wmcell}World model & Neither \\
Frat & Items & \cellcolor{wmcell}\textbf{0.546} & 0.843 & 0.654 & 0.008 & 0.311 & 0.135 & 0.95 & 0.66 & 0.81 & \cellcolor{wmcell}World model & \cellcolor{wmcell}World model \\
Frat & Stores & \cellcolor{wmcell}\textbf{0.590} & 0.779 & 0.634 & 0.007 & 0.205 & 0.071 & 0.99 & 0.77 & 0.88 & \cellcolor{wmcell}World model & \cellcolor{wmcell}World model \\
M5 & Items & \cellcolor{wmcell}\textbf{0.862} & 0.918 & 0.871 & 0.040 & 0.083 & 0.053 & 1.04 & 0.84 & 0.94 & \cellcolor{wmcell}World model & \cellcolor{wmcell}World model \\
M5 & Stores & 0.874 & 0.886 & \cellcolor{avgcell}\textbf{0.869} & 0.022 & 0.020 & 0.020 & 0.99 & 0.93 & 0.96 & Neither & Neither \\
Favorita & Items & \cellcolor{wmcell}\textbf{0.709} & 0.899 & 0.768 & 0.010 & 0.171 & 0.070 & 1.01 & 0.77 & 0.89 & \cellcolor{wmcell}World model & \cellcolor{wmcell}World model \\
Favorita & Stores & \cellcolor{wmcell}\textbf{0.729} & 0.766 & 0.731 & 0.020 & 0.017 & 0.020 & 0.97 & 0.98 & 0.97 & \cellcolor{wmcell}World model & Neither \\
\bottomrule
\end{tabular}\end{adjustbox}
\end{table*}

\begin{table*}[t]
\centering\footnotesize\setlength{\tabcolsep}{3pt}
\caption{RQ3, LightGBM without the item label. The kept-out item test of Table~\ref{tab:rq-transfer} with LightGBM fitted
without the item label as a feature (LGB$-$), every other setting unchanged. WRMSSE on the kept-out fifth; WRMSSE on the
same fifth when every item is trained on; the cost of not training on them; forecast over actual on the kept-out
fifth. Means over windows. The last column names the forecast that is clearly lower on the kept-out fifth when a 90\%
bootstrap interval over windows of the difference excludes zero.}
\label{tab:rq-noid}
\begin{adjustbox}{max width=\textwidth}\begin{tabular}{l cc ccc ccc ccc l}
\toprule
 & \multicolumn{2}{c}{Kept out} & \multicolumn{3}{c}{Trained on all} & \multicolumn{3}{c}{Cost of not training on them} & \multicolumn{3}{c}{Forecast over actual} & Clearly lower\\
\cmidrule(lr){2-3}\cmidrule(lr){4-6}\cmidrule(lr){7-9}\cmidrule(lr){10-12}
Data set & WM & LGB$-$ & WM & LGB & LGB$-$ & WM & LGB & LGB$-$ & WM & LGB & LGB$-$ & WM or LGB$-$\\
\midrule
Cheese & \cellcolor{wmcell}\textbf{0.678} & 0.732 & 0.667 & 0.673 & 0.730 & 0.011 & 0.058 & 0.002 & 0.99 & 0.93 & 1.05 & \cellcolor{wmcell}World model \\
Snacks & \cellcolor{wmcell}\textbf{0.626} & 0.663 & 0.614 & 0.591 & 0.650 & 0.012 & 0.162 & 0.013 & 0.89 & 0.68 & 0.85 & \cellcolor{wmcell}World model \\
Frat & \cellcolor{wmcell}\textbf{0.546} & 0.603 & 0.538 & 0.532 & 0.542 & 0.008 & 0.311 & 0.061 & 0.95 & 0.66 & 0.90 & \cellcolor{wmcell}World model \\
M5 & 0.862 & \cellcolor{lgbcell}\textbf{0.851} & 0.822 & 0.835 & 0.841 & 0.040 & 0.083 & 0.009 & 1.04 & 0.84 & 0.97 & Neither \\
Favorita & \cellcolor{wmcell}\textbf{0.709} & 0.797 & 0.699 & 0.728 & 0.754 & 0.010 & 0.171 & 0.043 & 1.01 & 0.77 & 0.95 & \cellcolor{wmcell}World model \\
\bottomrule
\end{tabular}\end{adjustbox}
\end{table*}

\subsection{RQ4: transferability}
\label{sec:rq-transfer}

\textbf{In the one transfer between retailers tested, M5's settings were
less accurate on Favorita than Favorita's own, and training on other data sets did not help
either model.} M5's tuned settings carried unchanged to
Favorita give \rqCarriedWm{} overall for the world model, against \rqOwnWm{} with Favorita's own
settings, and \rqCarriedLgb{} for LightGBM, against \rqOwnLgb{} (Table~\ref{tab:rq-retailer});
at item by store the difference is \rqCarriedWmIxs{} for the world model and \rqCarriedLgbIxs{} for
LightGBM. Training each model on its data set together with other data sets
lowers the error by no more than \rqPoolGainMax{} overall on any data set: by \rqPoolGainWm{} for the
world model on \rqPoolGainWmSet{} and by \rqPoolGainLgb{} for LightGBM on \rqPoolGainLgbSet{}. Elsewhere
it leaves the error unchanged or raises it, by as much as \rqPoolLossWm{} for the world model on \rqPoolLossWmSet{} and
\rqPoolLossLgb{} for LightGBM on \rqPoolLossLgbSet{}.

\begin{table*}[t]
\centering\footnotesize\setlength{\tabcolsep}{3pt}
\caption{RQ4, transferability. WRMSSE overall / item by store of each tuned model trained on its own data set and
trained on it together with other data sets (each Dominick's category with the other five and Frat; Frat with the
six Dominick's categories; M5 with Favorita and Favorita with M5). The last row carries M5's tuned settings
unchanged to Favorita, to be read against Favorita's own-data row.}
\label{tab:rq-retailer}
\begin{adjustbox}{max width=\textwidth}\begin{tabular}{l cc cc}
\toprule
 & \multicolumn{2}{c}{World model} & \multicolumn{2}{c}{LightGBM}\\
\cmidrule(lr){2-3}\cmidrule(lr){4-5}
Data set & Own data & Pooled & Own data & Pooled\\
\midrule
Analgesics & 0.699 / 0.669 & 0.697 / 0.671 & 0.702 / 0.657 & 0.916 / 0.723 \\
Bottled juice & 0.575 / 0.615 & 0.607 / 0.610 & 0.581 / 0.626 & 0.597 / 0.653 \\
Cheese & 0.502 / 0.639 & 0.500 / 0.632 & 0.532 / 0.618 & 0.592 / 0.650 \\
Frozen entrees & 0.357 / 0.422 & 0.363 / 0.431 & 0.338 / 0.396 & 0.335 / 0.406 \\
Soft drinks & 0.705 / 0.600 & 0.849 / 0.747 & 0.680 / 0.570 & 0.750 / 0.594 \\
Snacks & 0.893 / 0.579 & 0.912 / 0.572 & 0.778 / 0.565 & 0.835 / 0.610 \\
Frat & 0.490 / 0.568 & 0.492 / 0.570 & 0.494 / 0.563 & 0.548 / 0.626 \\
M5 & 0.709 / 0.838 & 0.733 / 0.840 & 0.697 / 0.858 & 0.851 / 0.876 \\
Favorita & 0.428 / 0.690 & 0.448 / 0.699 & 0.448 / 0.722 & 0.459 / 0.738 \\
\midrule
Favorita, M5's settings & 0.447 / 0.692 & & 0.450 / 0.723 & \\
\bottomrule
\end{tabular}\end{adjustbox}
\end{table*}

\ifjournal
\subsection{Decision costs}
\label{sec:rq-decisions}

\fi

\subsection{Ablation, occlusion and calibration}
\label{sec:rq-checks}

\textbf{Each tuned model earns its place in the combination, both read the
plan, and the forecast ranges hold what they state item by store.} Dropping the
world model from the three-model ensemble makes the item-by-store forecast less
accurate on \sfAblNoWmWorse{} of \rqNineWord{} data sets, and dropping LightGBM
on \sfAblNoLgbWorse{} (Section~\ref{sec:sup-ablation}). Switching the plan off
makes the world model less accurate on \supPertPlanHelpsWm{} of \supPertSets{}
data sets and LightGBM on \supPertPlanHelpsLgb{}; on the Dominick's categories
and Frat, price and the promotion flag carry it, and events, SNAP days and
paydays change the error by at most \sfCalendarMax{}
(Section~\ref{sec:sup-plan}). The forecast ranges, built from each model's
errors in earlier windows, hold their stated coverage item by store on average over
the data sets and are too narrow at the store; on the continuous ranked probability score the average is
the most accurate item by store on \sfCrpsAvgBestIxs{} of \rqNineWord{} data
sets. None of the models forecasts the chance of selling nothing better than
each series' own past share of zero-sale periods, except on
\sfBrierBeatsClimNames{} (Section~\ref{sec:sup-calibration}).

\subsection{The broad view}
\label{sec:rq-broad}

Brought together, the four answers give one rule. \textbf{Forecast every item in
every store as the equal-weight average of a tuned world model and tuned
gradient-boosted trees, and sum it to every grain; for an existing item the models
have not been trained on, with recent sales, use the world model alone until they are
retrained with it.} A new product with no sales was not tested. On the data sets studied, the average is the most accurate of the two tuned
models and their average at store grain and above on \rqAvgBestCoarseWord{} of \rqNineWord{} data sets, at the item levels on all \rqAvgBestItemWord{}, and
on the fast lines and the slow tail on \rqAvgBestFast{} of \rqNineWord{}. On the periods every final run held out it is
the most accurate on only \hoAvgBest{} of \hoSets{} data sets but never the least accurate, so it is the safe choice
rather than the best one everywhere. The two models on their own
are close on items they were trained on; they part company on items they were
not, where LightGBM under-forecasts and the world model does not. The trigger
for the exception is one a merchant always knows: whether the item is new to the
range. In the one transfer between retailers tested, M5's settings were
less accurate on Favorita than Favorita's own, and training on other data sets did not help. Where the forecast is used only for store
orders, adding TimesFM given the plan as a third member lowers the item-by-store
error on \ptThreeBelowAvg{} of the nine data sets (Section~\ref{sec:sup-ablation}), at the cost of
a third model to run.

Two parts of this rule were reached differently, and the difference matters to
a reader who would use it. Averaging models was the recommendation of the \ifjournal development runs\else\firstver{}\fi{}, and the average of the two tuned models holds it up here. The
exception was tested as part of a wider rule written down before it ran, which
also gave the world model every series with less than 26 weeks of history; that
half added nothing on the data with history (it was level with the average on
\rqRuleLevel{} data sets and worse on snacks, \rqRuleSna{} against \rqSnaAvg{}), and only the new-item half held. The rule
as stated is therefore drawn from these results.

The rule is a finding on these nine data sets, not a ranking of methods for any
retailer. A retailer, or a researcher, adopting the framework should rerun the
full set of comparisons on its own data, including the models and combinations
that lost here: the pretrained models, the Temporal Fusion Transformer, the
three-model ensemble, pooling across retailers and the forecast by history. These
results do not settle their order on data with other levers, other lines and
other history; the
code to do so is released with the paper. Section~\ref{sec:limits} sets out what
the study does not cover, including year-ahead forecasts beyond the 52-week test.

% 52 weeks ahead (docs/PAPER1_REVISION_PLAN.md). A subsection of the results in both versions (Karthik, 16 Nov:
% core to the paper); the designs and models added after the planned test are in long_horizon_more.tex, in the
% results of the arXiv version and in the supplement of the journal version. Every number from
% scripts/make_long_horizon_tables.py (generated/lh_values.tex, lh_tab.tex, lh_tab_all.tex).
\subsection{Forecasting 52 weeks ahead}
\label{sec:long-horizon}

\textbf{Over 52 weeks the average of the two models, each forecasting the whole year
at once, is the most accurate forecast on \lhLowestAvgDirect{} of \lhSets{} data sets and
LightGBM on the others. Between the two models alone, the world model is more accurate on
\lhWmDirAheadNames{} and LightGBM on the other data sets; in paired comparisons
LightGBM is clearly more accurate in \lhPWmDirWorse{} of \lhPWmDirN{} and the world model in
\lhPWmDirBetter{}. A year ahead the evidence for the rule is mixed.} Grocery retailers plan the year ahead, so the test asks
whether the rule of Section~\ref{sec:rq-broad} carries from 8 weeks to 52. It
was written down before it ran: the six Dominick's categories and M5, two
windows of 52 weeks each (364 days on M5), the final tuned settings with only the
distance ahead changed, scored at every level as in the rest of the paper
(Table~\ref{tab:long-horizon}, Figure~\ref{fig:long-horizon}). The later window
ends where the final runs' data end (November 1996 on Dominick's, February 2016
on M5) and the earlier starts 26 weeks (182 days) before it; both overlap the
8-week test windows and the weeks the settings searches were scored on, so the
test asks how the models hold over a longer distance, not how they do on unseen
weeks.

\paragraph{The planned comparison, in blocks}
Each model forecasts 13 weeks, takes its forecast in as the sales of those weeks,
and forecasts the next 13, four times. Here the world model is clearly less
accurate than LightGBM in \lhPWmWorse{} of \lhPN{} paired comparisons and clearly
more accurate in \lhPWmBetter{}; against the better of LightGBM's two setups it
is behind on \lhLgbBestBeatsWm{} of \lhSets{} data sets. The average in blocks is
clearly more accurate than the world model alone in \lhPAvgWmBetter{} comparisons
(clearly less in \lhPAvgWmWorse{}), and than LightGBM in \lhPAvgLgbBetter{}
(clearly less in \lhPAvgLgbWorse{}).

\paragraph{Taking in its own forecasts costs the world model most}
Forecasting all 52 weeks at once, the world model is more accurate than in blocks
on \lhDirectBeatsBlocksWm{} of \lhDirectSetsWm{} data sets, LightGBM on
\lhDirectBeatsBlocksLgb{} of \lhDirectSetsLgb{}. In blocks, the world model's item
by store error changes by \lhGrowWmLo{} to \lhGrowWmHi{} from the first 13 weeks
to the last, LightGBM's by \lhGrowLgbLo{} to \lhGrowLgbHi{}. The world model rolls
its state forward from the sales it takes in; when those are its own forecasts,
its errors are carried into the state and grow with each block. Forecasting directly, on Dominick's
its gap to LightGBM at item by store sits in the price-cut weeks (\lhGapCutLo{} to \lhGapCutHi{} in RMSSE
across the categories) and is near zero in regular weeks (\lhGapRegLo{} to \lhGapRegHi{}).

\paragraph{The average loses to LightGBM in few comparisons}
The average of the two models, each forecasting the whole year at once, is
clearly more accurate than LightGBM forecasting directly in \lhPAvgDirBetter{} of
\lhPAvgDirN{} paired comparisons and clearly less in \lhPAvgDirWorse{}; against
the world model forecasting directly it is clearly more accurate in
\lhPAvgDirWmBetter{} and clearly less in \lhPAvgDirWmWorse{}. These two
comparisons were added after the planned ones.
\ifjournal Designs and models added after the planned test are in
Section~\ref{sec:long-horizon-more} of the supplement.\fi

\begin{table*}[t]
\centering\footnotesize\setlength{\tabcolsep}{3pt}
\caption{Forecasts 52 weeks ahead (364 days on M5). WRMSSE over the 52 weeks, mean over levels and two windows, for
the world model (WM), LightGBM (LGB) and their average (Avg). In blocks: each model forecasts four blocks of 13 weeks,
each block's forecast taken in as sales before the next. Direct: each model forecasts all 52 weeks at once (not run for
M5's world model). Paired item bootstrap over levels and windows: comparisons clearly lower / clearly higher, of all.
Last four columns: WRMSSE item by store in the first and last block of 13 weeks, in blocks. Shading marks the lowest
of the six: \colorbox{wmcell}{world model}, \colorbox{lgbcell}{LightGBM}, \colorbox{avgcell}{average}.}
\label{tab:long-horizon}
\begin{adjustbox}{max width=\textwidth}\begin{tabular}{l ccc ccc cc cc cc}
\toprule
 & \multicolumn{3}{c}{In blocks} & \multicolumn{3}{c}{Direct} & \multicolumn{2}{c}{Paired} & \multicolumn{2}{c}{WM, item by store} & \multicolumn{2}{c}{LGB, item by store}\\
\cmidrule(lr){2-4}\cmidrule(lr){5-7}\cmidrule(lr){8-9}\cmidrule(lr){10-11}\cmidrule(lr){12-13}
Data set & WM & LGB & Avg & WM & LGB & Avg & WM vs LGB, blocks & Avg vs LGB, direct & Weeks 1 to 13 & 40 to 52 & Weeks 1 to 13 & 40 to 52\\
\midrule
Analgesics & 1.062 & \cellcolor{lgbcell}\textbf{0.865} & 0.907 & 0.964 & 0.867 & 0.868 & 0 / 3 of 8 & 1 / 0 of 8 & 0.697 & 0.686 & 0.696 & 0.683 \\
Bottled juice & 0.903 & 0.797 & 0.817 & 0.764 & 0.769 & \cellcolor{avgcell}\textbf{0.726} & 0 / 2 of 8 & 4 / 0 of 8 & 0.738 & 0.742 & 0.705 & 0.689 \\
Cheese & 1.086 & 0.689 & 0.820 & 0.773 & 0.707 & \cellcolor{avgcell}\textbf{0.677} & 0 / 8 of 8 & 2 / 0 of 8 & 0.701 & 0.938 & 0.692 & 0.738 \\
Frozen entrees & 0.424 & 0.397 & 0.393 & 0.412 & 0.393 & \cellcolor{avgcell}\textbf{0.387} & 0 / 5 of 8 & 2 / 0 of 8 & 0.469 & 0.404 & 0.450 & 0.370 \\
Soft drinks & 0.872 & 0.862 & 0.819 & 0.851 & \cellcolor{lgbcell}\textbf{0.754} & 0.767 & 1 / 0 of 8 & 0 / 2 of 8 & 0.618 & 0.767 & 0.631 & 0.781 \\
Snacks & 1.013 & 0.877 & 0.920 & 0.866 & 0.882 & \cellcolor{avgcell}\textbf{0.843} & 0 / 6 of 8 & 6 / 0 of 8 & 0.691 & 0.890 & 0.622 & 0.683 \\
M5 & 1.016 & 0.931 & 0.874 & 0.922 & 0.839 & \cellcolor{avgcell}\textbf{0.796} & 4 / 14 of 24 & 15 / 4 of 24 & 0.823 & 1.000 & 0.849 & 0.908 \\
\bottomrule
\end{tabular}\end{adjustbox}
\end{table*}

\begin{figure*}[t]
\centering
\includegraphics[width=\textwidth]{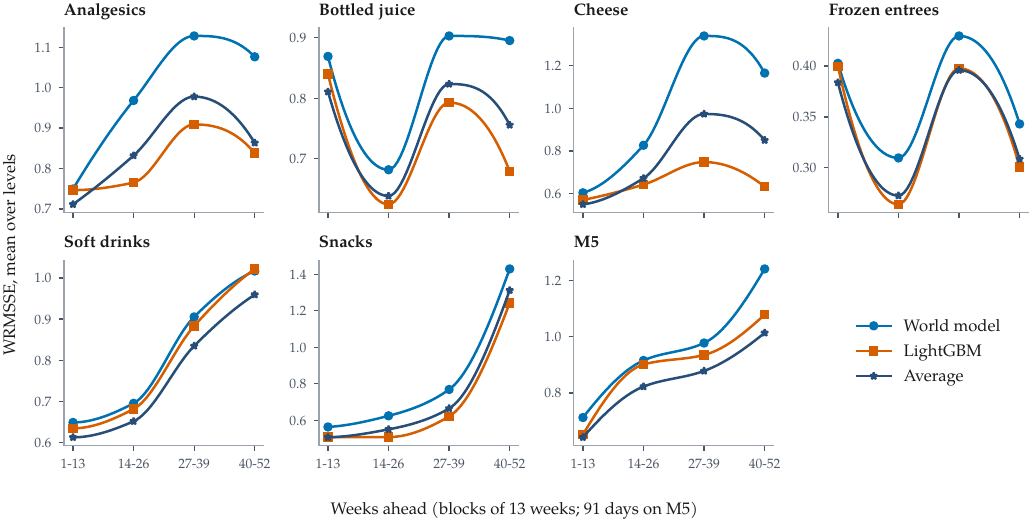}
\caption{\textbf{Taking in its own forecasts 13 weeks at a time, the world model is
the least accurate of the three in most blocks.} WRMSSE, mean over levels and two windows, in each block of 13
weeks (91 days on M5), for the world model, LightGBM and their average
forecasting in blocks.}
\label{fig:long-horizon}
\end{figure*}

\ifjournal\else\fi

\ifjournal\else

\fi

% =====================================================================
\section{Discussion}
\label{sec:discussion}

Table~\ref{tab:advice} maps the rule of Section~\ref{sec:rq-broad} to the
decisions a retailer makes from the forecast. Each row is what the tuned models
did on the nine data sets studied, within the scope of
Section~\ref{sec:limits}.

\begin{table*}[t]
\centering
\caption{The decisions a retailer makes from one item-by-store forecast, and what
the tuned models did at each on the nine data sets studied.}
\label{tab:advice}
\footnotesize
\renewcommand{\arraystretch}{1.2}
\begin{tabular}{p{0.21\textwidth}p{0.17\textwidth}p{0.18\textwidth}p{0.36\textwidth}}
\toprule
\textbf{Decision} & \textbf{Grain} & \textbf{Forecast to use} & \textbf{What the tuned models did} \\
\midrule
Store orders and inventory & Item by store, by day or week & The average & Lowest error at the item levels on all \rqAvgBestItemWord{} data sets; clearly less accurate than either model in at most \rqAvgItemWorseMax{} of \rqPAvgWmItemn{} paired comparisons; simulated order cost below LightGBM's on \supOrderAvgBelowLgb{} of \supOrderSets{} \\
Promotion and supplier funding & Item across stores, by week and over the promotion block & The average & Lowest at the item levels on all \rqAvgBestItemWord{}; over the whole block lowest on \rqAvgBestBlockWord{} and second on the other four; LightGBM alone is less accurate over the block on Frat, M5, Favorita and soft drinks \\
Category performance & Category by store and category, by week & The average & Lowest at store grain and above on \rqAvgBestCoarseWord{} of \rqNineWord{}; LightGBM lowest on snacks; lowest category-plan cost on \supCatPlanAvgLowest{} of \supCatPlanSets{} \\
Financial plan & Chain total & The average & As for category performance; the total is the sum of the same item-by-store forecast, so it agrees with every plan below it \\
Existing items not yet in training, with recent sales (range changes; not new launches) & Item by store, existing items held out of training & The world model alone & Clearly more accurate than the average on all \rqKeptSets{} data sets tested, and than LightGBM without the item label on \rqWmBeatsNoid{} of \rqNoidSets{}; LightGBM with the label forecast \rqFaLgbItemLo{} to \rqFaLgbItemHi{}\% of actual sales on these items \\
Stores not yet in training & Item by store, existing stores held out of training & The average or the world model (level) & The average and the world model each clearly ahead on \rqStoreAvgBeatsWm{} data set and level on \rqStoreLevel{}; LightGBM clearly behind the world model on \rqStoreWmBeatsLgb{} of \rqKeptSets{} \\
Choosing for another retailer & Any & Rerun the comparison & The order of models is a finding on these data; every comparison, including those that lost here, should be rerun on the retailer's own data \\
\bottomrule
\end{tabular}
\end{table*}

\paragraph{Why the average is more accurate}
The two tuned models are close in accuracy and wrong in different places. Across
the nine data sets, each is clearly more accurate than the other in a similar
number of paired comparisons (\rqPWmLgbBetter{} for the world model, \rqPWmLgbWorse{} for LightGBM), and the
data sets on which each leads differ. An equal-weight average of two forecasts
whose errors differ is more accurate than either on most series, and no weights
have to be estimated, which the forecast-combination literature finds is hard to
beat \cite{smithwallis2009,wang2023}. These findings by grain are in line with
the literature and with practice: gradient-boosted trees are strong on retail
panels~\cite{m5accuracy}, deployed promotion planning forecasts
demand with them~\cite{pepsico2026}, and the method that suits an item depends on its demand
pattern~\cite{syntetos2005,ma2025}.

\paragraph{What the world model brings}
The world model's contribution is clearest where the forecast has least to go
on. On items it was not trained on it lost at most \rqCostWmItemHi{} in scaled error, against
up to \rqCostLgbItemHi{} for LightGBM, and forecast within \rqWmFaItemOff\% of actual sales; LightGBM
forecast as little as \rqLgbFaItemLoFrac{}. It reads an item's recent sales into a state
and carries that state forward under the plan, so it does not need the item in
its training data. LightGBM leans on the item label: without it, LightGBM loses
little on these items but is less accurate on the items it was trained on, and is
still behind the world model on the kept-out items on \rqWmBeatsNoid{} of \rqNoidSets{} data sets. The same carried state keeps its totals over the
promotion block closer to actual sales than LightGBM's on M5, Favorita and soft
drinks.

\paragraph{What LightGBM adds to the average}
LightGBM is the more accurate of the two on frozen entrees and snacks at every
grain, and its block totals are closer than the world model's on every
Dominick's category but soft drinks.
It needs no pretraining and trains in minutes, and on lines with history it is
as accurate as the world model on most data sets. On snacks, where the world
model alone is weakest (\rqSnaWm{} overall against \rqSnaLgb{} for LightGBM), the average
stays close to LightGBM at \rqSnaAvg{}.

\paragraph{Why run both}
Running both models means a second pipeline, a GPU-trained network beside trees
that train in minutes, and on lines with history the average gains little over
the better single model. The case for it rests on totals: over the promotion block LightGBM alone is less
accurate than the world model on M5, Frat, Favorita and soft drinks, most at the
chain total, while the world model alone is less accurate than LightGBM on the
other five Dominick's categories.
On M5 its block error is \rqBlockMfiveLgb{}, against \rqBlockMfiveWm{} for the world
model and \rqBlockMfiveAvg{} for the average (Section~\ref{sec:rq-grain}). A retailer
whose plans are summed to the category and the chain pays for that error at every
level above the item.
The searches behind the tuned settings took \computeSearchHours{} hours of run time
across the nine data sets (Section~\ref{sec:repro}), a one-off cost. In the one transfer
between retailers tested, settings tuned on M5 were less accurate on Favorita than Favorita's own
(Section~\ref{sec:rq-transfer}), so a retailer should run the search on its own data.

% =====================================================================
\section{Limitations and threats to validity}
\label{sec:limits}
\label{sec:limitations}

\paragraph{Scope: two of the four Ps, on public data}
The paper addresses price and promotion. It forecasts items newly added to a
range but does not decide product (which items to range) or place (space,
distribution), and it does not solve the four Ps together; a unified forecast across them does not exist in
this work, and the study does not show which model would work best at each grain
under one. The nine data sets come from four retailers, and six of them are categories of
one retailer (Dominick's), so the evidence across retailers rests on four. The six were chosen from a first,
untuned run of all \domCatN{} categories, four of them because the world model did worst there, so results on
Dominick's are not an estimate for its average category (Table~\ref{tab:domcats}). The other \domCatRest{}
categories were not tuned, so whether tuning would change which categories are hardest for the world model was
not tested; on the six, tuning closed most of the first run's gap. The ordering of models is not established for a single retailer's
data, with its own levers, stock and space records, and the grain comparison
should be repeated on it before a choice is made.

\paragraph{Adapting the findings}
The rule of Section~\ref{sec:rq-broad} and the ordering of models behind it hold
on the nine data sets studied. They are not a ranking of methods for retail in
general. A retailer or researcher adopting the framework should rerun every
comparison on its own data, including the models and combinations that lost
here, since data with other levers, lines and lengths of history can change the
order. The code and settings for every comparison are released for that purpose.

\paragraph{Forecast horizon}
The main comparisons run 8 weeks ahead on the weekly data and 28 days ahead on
the daily data. The 52-week test (Section~\ref{sec:long-horizon}) covers six
Dominick's categories and M5 with two windows each, with the settings tuned for
the shorter horizon; Frat and Favorita were not tested a year ahead. The test's windows overlap earlier evaluation and search
periods, so it checks a longer horizon rather than unseen periods. Whether the order holds
on a retailer's own year-ahead plans, which are revised as the year runs, was not
tested.

\paragraph{New items}
The kept-out items are existing items held out of training, not new products. Each
model read their recent sales before forecasting them. An item with no sales at all, on its
first day in range, was not tested; the new-item part of the rule covers items
the models have not been trained on, not new products with no history; a new
product launch, with no sales and often no like item, needs its own test. That part of
the rule was drawn from the results, after a wider rule written down in advance
held only in part (Section~\ref{sec:rq-broad}).

\paragraph{Windows and seeds}
Four to six windows per data set is a thin basis for a bootstrap over windows
and for a model confidence set: the intervals are approximate, the set has
little power, and a model kept in it has not been shown to be as good as the
best. Adjacent windows share training history and season, so they carry less
information than their number suggests; the item bootstrap within a window
treats the items as the sample and does not cover variation between windows.
The kept-out comparisons of RQ3 rest on the bootstrap over windows, so their
counts of clear differences carry this caveat.
Averaging five fits with different seeds removes much of the noise between runs
that the searches measured (Section~\ref{sec:loopnoise}), but both models on M5
are single fits.

\paragraph{Analyses added after the first results}
These were added after the first scores of the tuned models were read; each was
written down before it ran, but which to run was chosen knowing earlier results:
the analyses by line type, the whole-block view, the forecast by history and the
supporting reruns; the new-item part of the recommended rule, drawn from the
results (Section~\ref{sec:rq-broad}); in the 52-week test, everything beyond the
planned comparison in blocks and direct (the world model rolled on, LightGBM by
block, the other models, the settings searched again for 52 weeks, and the
price-cut and regular weeks); the scoring by kind of line and week, with its
within-data-set contrasts; the comparison of the average with other second
models (Section~\ref{sec:sup-ablation}); and the history-length test.

\paragraph{Comparison models}
Only the world model and LightGBM were tuned; the Temporal Fusion Transformer
runs with the settings of the \firstver{}, and the pretrained models as
published. The comparison between the two tuned models is therefore the fair one,
and the others are reference points: a tuned Temporal Fusion Transformer could do
better than the one reported. Hierarchical reconciliation was tested only against
exponential-smoothing base forecasts; reconciling the tuned models' own forecasts
(for example with MinT) needs their forecasts at every level, which the study does
not make. The tests on items and stores kept out of training (RQ3) ran on five of
the nine data sets (cheese, snack crackers, Frat, M5 and Favorita), fixed when the
test was written down, before it ran; the other four Dominick's categories were
not tested. Every model was run on every data set.
TimesFM with the
plan fits its linear plan model within groups of 2,048 series. The pretrained
models' point forecast was fixed in advance as the mean of their output, which
for Timestral let a few extreme upper quantiles dominate on weekly data.

\paragraph{What lift means}
Merchants promote in weeks they expect to sell well, so part of a measured lift
is good timing. The lift is what the forecast attributes to the plan, not what
the plan caused. \ifjournal An audit of these models~\cite{paper2}\else Our companion audit~\cite{paper2}\fi{} measures the gap in generated store
data where the true lifts are known. Plan responses in this paper, including the
perturbation tests, are forecasts, not causal estimates.

\paragraph{What the data do not record}
None of the data sets records stock, so a sell-out looks like low demand; or
shelf space. Favorita records no prices. The data were collected between 1989
and 2017 and their plan inputs differ, so the results compare models within each
data set. The decision costs are single-period simulations under stated costs, not a
dynamic inventory system, and they do not establish savings in a deployed retailer.

% =====================================================================
\section{Conclusion}
\label{sec:conclusion}

A retailer makes decisions from the same demand at different grains, and on these data no single
model was the most accurate everywhere. The findings fall into four parts.

\textbf{When to use the average.} One forecast made item by store and summed upward, as the
equal-weight average of a tuned world model and tuned gradient-boosted trees, was the most accurate
of the two models and their average at store grain and above on \rqAvgBestCoarseWord{} of
\rqNineWord{} data sets and at the item levels on all \rqAvgBestItemWord{}. On the periods every
final run held out it was the most accurate on only \hoAvgBest{} of \hoSets{} data sets but never the
least accurate. It is the safe default, not the best choice everywhere.

\textbf{When to use the world model.} For existing items absent from training, with recent sales, the world model alone was more
accurate than the average on every data set tested, because LightGBM with the item label
under-forecast them; without the label LightGBM came closer but was still less accurate on
\rqWmBeatsNoid{} of \rqNoidSets{}. On lines with short history the world model was clearly more
accurate than LightGBM on \kWmBetterShortHistory{} of \rqNineWord{} data sets.

\textbf{When to use LightGBM.} On heavily promoted lines, fast sellers and lines with long history
LightGBM was clearly more accurate than the world model on most data sets, and on Dominick's in
price-cut weeks, where the world model's gap also sits a year ahead.

\textbf{Where the evidence is inconclusive.} Slow lines and lines never promoted were close to
even. More training history helped the world model on one daily data set and hurt it on the
other. Settings carried between retailers and training on other data sets did not help. A year
ahead, the average was the most accurate on \lhLowestAvgDirect{} of \lhSets{} data sets and LightGBM
on the rest.

These are findings on these data, for forecasts of 8 weeks and 28 days and a test of 52 weeks,
under price and promotion plans. The comparison behind them, run over several windows with paired
tests and every losing model reported, is what we propose a retailer reruns on its own data, and
over its own planning horizons, before adopting them.

\ifjournal\else

\fi

\ifjournal
\section{Data availability}
\label{sec:licence}
M5 is available from the competition's Kaggle page~\cite{m5accuracy};
Dominick's from the Kilts Center for Marketing, University of Chicago Booth
School of Business~\cite{dominicks}; Breakfast at the Frat from
dunnhumby~\cite{frat}; Corporaci\'on Favorita from its Kaggle
competition~\cite{favorita}. The code, configuration files and every result file
behind the tables and figures are archived in a public repository, linked on the
title page and withheld here for anonymised review; how they rebuild the paper is
set out in Section~\ref{sec:repro} of the online supplement.

\else
\section{Data, code and licence}
\label{sec:licence}

\paragraph{Data}
M5 is available from the competition's Kaggle page~\cite{m5accuracy};
Dominick's from the Kilts Center for Marketing, University of Chicago Booth
School of Business~\cite{dominicks}; Breakfast at the Frat from
dunnhumby~\cite{frat}; Corporaci\'on Favorita from its Kaggle
competition~\cite{favorita}. Each is obtained from its provider under the
provider's own terms. No data is redistributed with the code.

\paragraph{Code}
The code is at
\url{https://github.com/karthik-iyer-au/retail-world-models}; the version used
for this paper is archived as release v1.0.0 on Zenodo,
\href{https://doi.org/10.5281/zenodo.24271672}{doi:10.5281/zenodo.24271672}.

\paragraph{Copyright and licence}
Copyright \copyright{} 2026 the authors. The code is licensed under the
PolyForm Noncommercial License~1.0.0 and the manuscript text, figures and tables
under Creative Commons Attribution-NonCommercial~4.0 International
(CC~BY-NC~4.0). Industry and commercial use of the code, the manuscript or any
part of them, including use inside a company, is prohibited unless approved in
writing by the authors.

\section{Author contributions}
\label{sec:contributions}
Karthik Iyer conceived and scoped the work, designed the study, wrote the code,
ran the experiments and analyses, and wrote the paper, with AI assistance as
declared in \S\ref{sec:ai}. Madan Dabbeeru widened the range of methods
considered for each research question, and reviewed the work and the
manuscript.

\section{Practice background and independence}
\label{sec:background}
The first author has worked in retail for fifteen years, across continents, in
grocery, department stores, fashion, electronics and online marketplaces. No
data, method, code or other intellectual property from those roles or
engagements is used in this work, and no employer or client data are used: the
study is built from the ground up on public data sets. What the authors carry
from practice is the set of questions retailers ask of a forecast, which shaped
the research questions in Section~\ref{sec:framework}. The first author's
research focus is world models; the second author widened the scientific
options considered for answering each question.
\fi

\section{Declaration of generative AI and AI-assisted technologies}
\label{sec:ai}
The research questions, scope, study design, choice of data and methods, and
interpretation of results are the authors' own. Claude (Anthropic) was used as an
assistant to write and test code, run analyses, produce figures and tables, and
draft and edit text. The authors directed this work throughout, reviewed every
result, and corrected and revised the code, analyses and text over many
iterations, and take full responsibility for the content of the publication.

\FloatBarrier

\clearpage
\appendix
\ifjournal\else\fi

\fi

\end{document}